\documentclass{article} 
\usepackage{arxiv}
\setcitestyle{authoryear,open={(},close={)}}

\usepackage{fullpage}

\usepackage{amsmath,amsfonts,bm}

\def\eqref#1{equation~\ref{#1}}

\def\1{\bm{1}}

\DeclareMathAlphabet{\mathsfit}{\encodingdefault}{\sfdefault}{m}{sl}
\SetMathAlphabet{\mathsfit}{bold}{\encodingdefault}{\sfdefault}{bx}{n}

\usepackage{url}
\usepackage{graphicx}
\usepackage{wrapfig}
\usepackage{bbm}
\usepackage{amsmath}
\usepackage{amssymb}
\usepackage{mathtools}
\usepackage{amsthm}
\usepackage{enumitem}
\usepackage{natbib}
\usepackage{xcolor}
\usepackage{multirow}
\usepackage{setspace}
\usepackage{enumitem}
\usepackage{color, colortbl}
\usepackage{mathtools}
\usepackage{cleveref}

\theoremstyle{plain}
\newtheorem{theorem}{Theorem}[section]

\usepackage{hyperref}
\usepackage{listings}
\usepackage{url}
\usepackage[linesnumbered,ruled,vlined]{algorithm2e}
\usepackage{color,graphicx}
\usepackage{siunitx}
\usepackage{soul}
\usepackage{graphics} 
\usepackage{bookmark}
\usepackage[dvipsnames, svgnames, x11names]{xcolor}
\usepackage{hyperref} 
\usepackage{textcomp}
\usepackage{multirow}
\usepackage{nth}
\usepackage{indentfirst}
\usepackage{siunitx}
\usepackage{changepage}
\usepackage{bm}
\usepackage{setspace}
\usepackage{natbib}
\usepackage{multirow}
\usepackage{multicol}
\usepackage{colortbl}
\usepackage{booktabs}
\usepackage{amsmath}
\usepackage{xcolor}
\usepackage{amsthm}
\theoremstyle{remark}

\usepackage{graphicx}
\usepackage{subfigure}
\usepackage{wrapfig}
\usepackage{enumitem}

\usepackage[most]{tcolorbox}

\definecolor{violetBg}{RGB}{245,241,250}
\definecolor{violetBorder}{RGB}{136,117,201}
\definecolor{violetAccent}{RGB}{106,90,205}
\definecolor{lightblue}{rgb}{0.933,0.968,0.988}

\newtcolorbox{questionbox}[1][]{%
enhanced,
breakable,
colback=violetBg,
colframe=violetBorder,
boxrule=0.6pt,
arc=4pt,
left=8pt,right=8pt,top=8pt,bottom=8pt,
halign=center,
#1
}

\usepackage{geometry}
\usepackage{titling} 
\usepackage{setspace}

\newcommand{\gray}[1]{\textcolor{gray}{#1}}

\newcommand{\ours} {{ROVER }}

\title{\textbf{Random Policy Valuation is Enough for LLM Reasoning with Verifiable Rewards}}
\author{
{{\textbf{Haoran He$^{1}$\thanks{Equal contribution.} ~ Yuxiao Ye$^{1*}$ ~ Qingpeng Cai$^{2}$ ~ Chen Hu$^{3}$ ~ Binxing Jiao$^{3}$}}}\\
{\textbf{Daxin Jiang$^{3}$ ~ Ling Pan$^{1}$}} \\
{\normalsize{$^{1}$Hong Kong University of Science and Technology $^{2}$Kuaishou Technology $^{3}$StepFun}} \\
\normalsize{\texttt{haoran.he@connect.ust.hk} ~ \texttt{lingpan@ust.hk}}
}
\date{}

\makeatletter
\renewenvironment{abstract}{
\begin{center}{\Large\bfseries Abstract}\end{center}
\vspace{-.8em}
\begingroup
\setlength{\parskip}{0.35em}
\setstretch{1.}
\normalsize
}{\par\endgroup}
\makeatother

\begin{document}

\newgeometry{left=1in,right=1in,top=0.6in,bottom=0.6in}

\maketitle

\vspace{-.5in}
\begin{abstract}
RL with Verifiable Rewards (RLVR) has emerged as a
promising paradigm for improving the reasoning abilities of large language models (LLMs). Current methods rely primarily on policy optimization frameworks like PPO and GRPO, which follow generalized policy iteration that alternates between evaluating the current policy's value and improving the policy based on evaluation. While effective, they often suffer from training instability and diversity collapse, requiring complex heuristic tricks and careful tuning. 
We observe that standard RLVR in math reasoning can be formalized as a specialized finite-horizon Markov Decision Process with deterministic state transitions, tree-structured dynamics, and binary terminal rewards. 
Though large in scale, the underlying structure is simpler than general-purpose control settings for which popular RL algorithms (e.g., PPO) were developed,
suggesting that several sophisticated techniques in existing methods may be reduced or even omitted.
Based on this insight, we prove a surprising result: the optimal action can be recovered from the Q-function of a fixed uniformly random policy,
thereby bypassing the generalized policy iteration loop and its associated heuristics. We introduce \underline{\textbf{R}}andom P\underline{\textbf{o}}licy \underline{\textbf{V}}aluation for Div\underline{\textbf{e}}rse \underline{\textbf{R}}easoning (ROVER) to translate this principle into a practical and scalable algorithm for LLM math reasoning, a minimalist yet highly effective RL method that samples actions from a softmax over these uniform-policy Q-values.
ROVER preserves diversity throughout training, allowing sustained exploration of multiple valid pathways.
Across multiple base models and standard math reasoning benchmarks, 
ROVER demonstrates superior performance in both \textbf{quality} (\textbf{+8.2} on pass@1, \textbf{+16.8} on pass@256) and \textbf{diversity} (\textbf{+17.6\%}), despite its radical simplification compared to strong, complicated existing methods.

\noindent \textit{``Simplicity is the ultimate sophistication." - Leonardo da Vinci} 

\noindent \textbf{Github:} \url{https://github.com/tinnerhrhe/ROVER/}


\begin{figure}[H]
    \centering
    \subfigure[]{\includegraphics[width=0.38\linewidth]{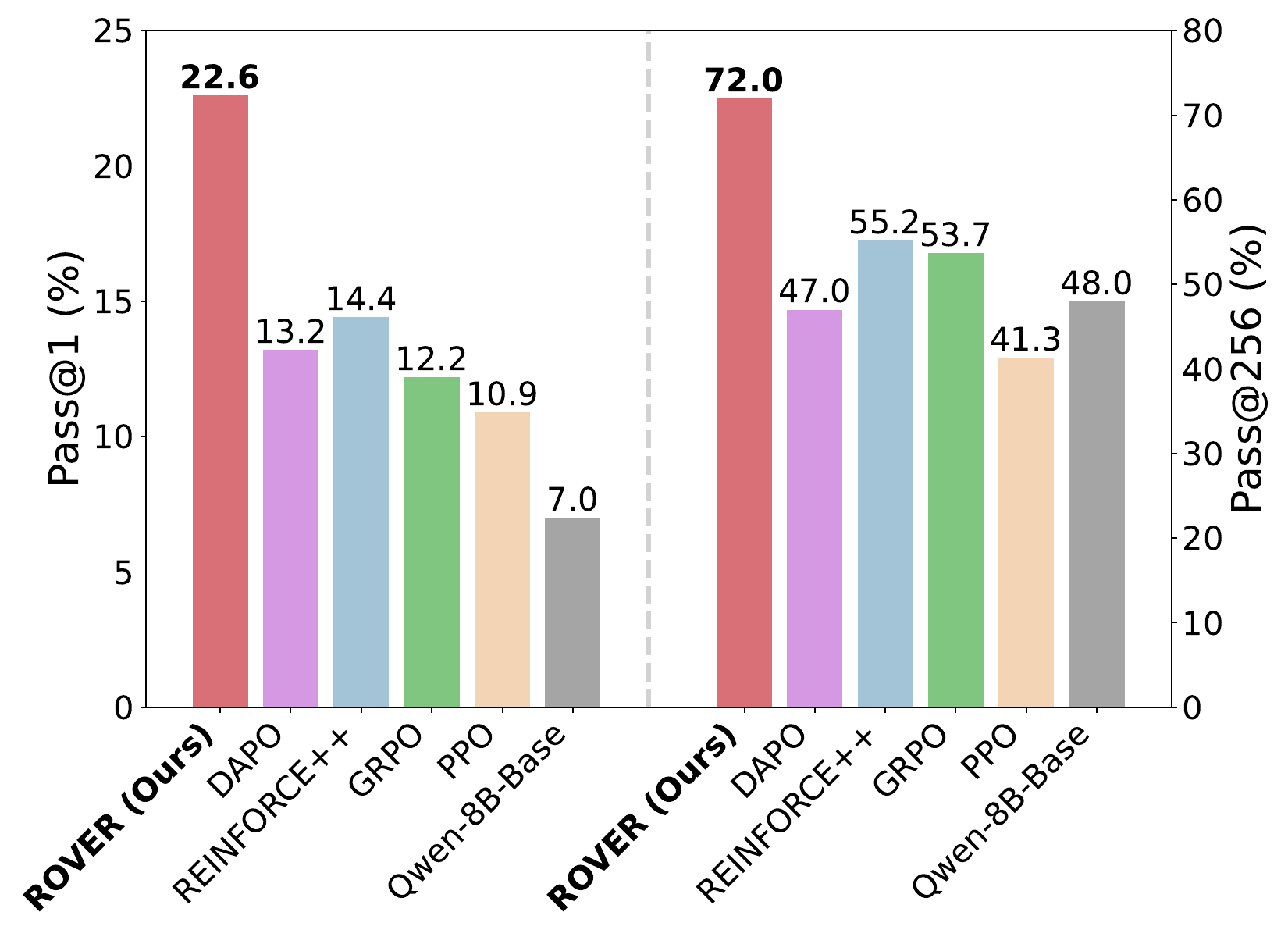}}
    \subfigure[]{\raisebox{1em}{\includegraphics[width=0.29\linewidth]{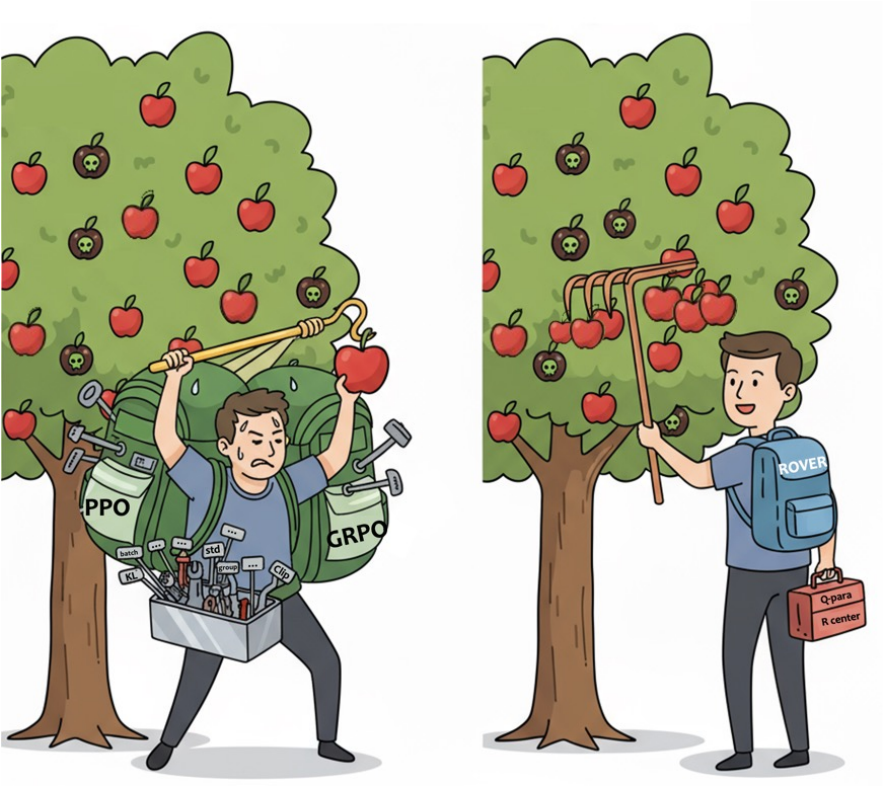}}}
    \subfigure[]{\includegraphics[width=0.31\linewidth]{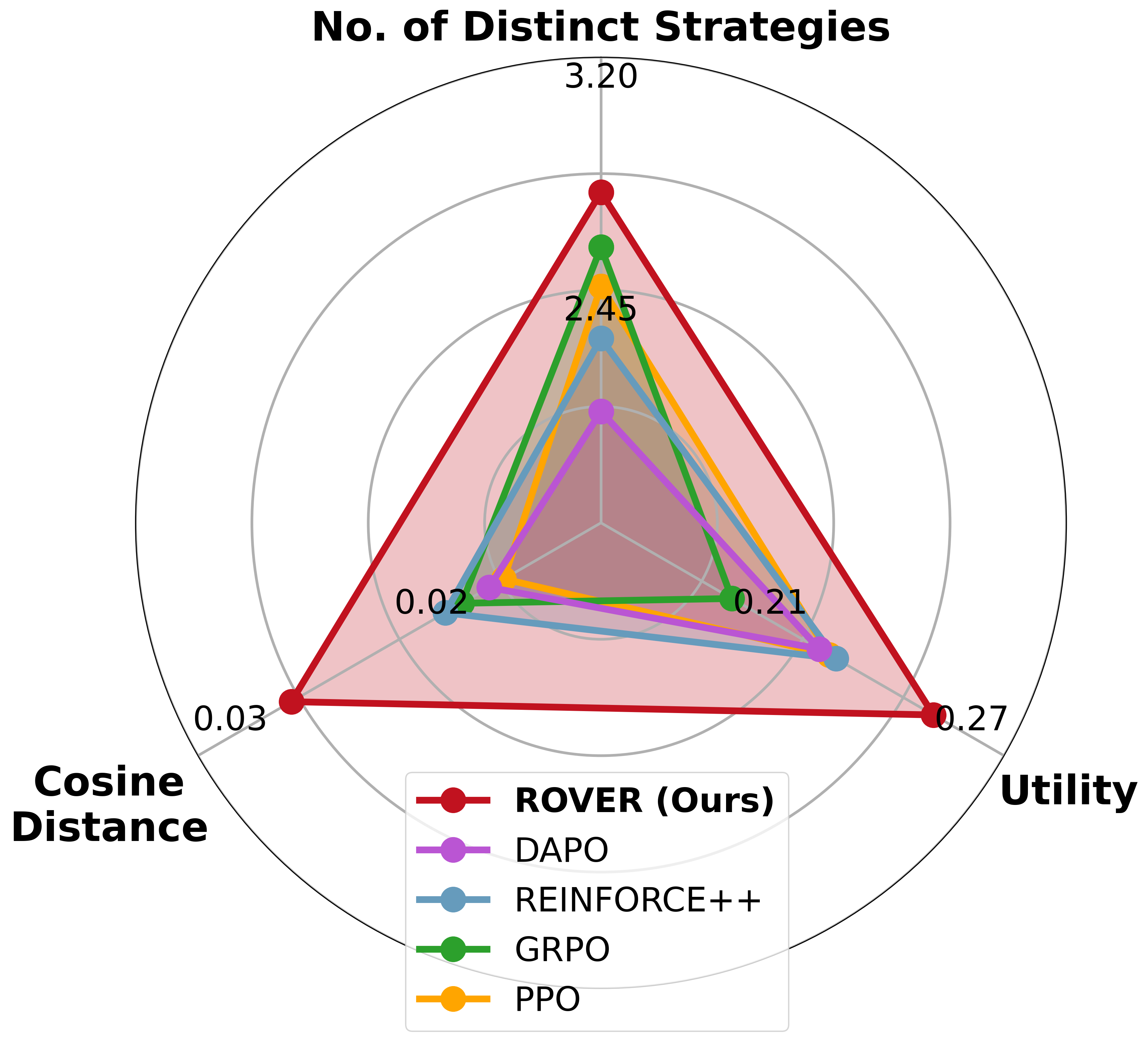}}
    \caption{(a) Pass@1 \& Pass@256 results on {Qwen3-8B-Base} averaged over AIME24, AIME25, and HMMT25 tasks. (b) Illustrative example demonstrating that \ours achieves high-quality solutions with a lightweight procedure while maintaining diversity. (c) Comparison on multiple diversity metrics. Higher value denotes better diversity.}
    \label{fig:main-figure}
\end{figure}
\end{abstract}

\clearpage
\restoregeometry

\section{Introduction}
RLVR has emerged as a promising paradigm for post-training LLMs and enhancing reasoning capabilities~\citep{jaech2024openai,guo2025deepseek}. 
The field has primarily relied on Proximal Policy Optimization (PPO)~\citep{schulman2017proximal}, a powerful algorithm originally designed for standard deep RL benchmarks 
such as computer games and robotic control. This general-purpose algorithm and its specialized derivatives like Group-Relative Policy Optimization (GRPO)~\citep{shao2024deepseekmath}
have achieved notable successes in improving LLM reasoning performance. Fundamentally, current methods follow the generalized policy iteration (GPI)~\citep{sutton1998reinforcement} paradigm, which iteratively alternates between \textit{evaluating} the current policy and \textit{improving} it based on the evaluation.

Despite its success, 
they suffer from unstable learning dynamics~\citep{yang2025qwen3} and entropy collapse~\citep{huang2024self,yang2025alignment} induced by the reward-maximizing nature within the \textit{iterative policy evaluation-improvement cycle}. As the policy continuously evolves, the evaluation target becomes non-stationary, leading to training instability and narrowed exploration spaces.
Recent variants mitigate this through an intricate ballet of heuristic techniques such as clipping~\citep{yu2025dapo}, KL regularization~\citep{liu2025prorl}, and data selection~\citep{liang2025beyond}.
While incorporating these tricks offers partial improvements, they add layers of implementation complexity and typically require careful, case-specific tuning~\citep{liu2025part}.

We take a fundamentally different approach by examining the underlying structure of LLM math reasoning tasks with verifiable rewards. Unlike standard RL environments that sophisticated RL algorithms like PPO were originally designed for and evaluated (e.g., discrete computer games with cyclic state transitions that forms a graph instead of a tree~\citep{bengio2021flow}, robotics with continuous spaces, possibly with stochastic transitions and intermediate rewards), standard RLVR for math reasoning corresponds to a specialized finite-horizon Markov Decision Process (MDP) with deterministic, tree-structured transitions, and binary terminal reward. In this structurally simplified MDP, each action induces a deterministic and new branch, and each partial sequence has exactly one parent state.
This critical observation leads us to a central question that whether we are applying unnecessarily complex tools to a structurally simpler (albeit larger) problem:
\begin{center}
\begin{questionbox}
\emph{Is there a minimalist yet highly effective RLVR algorithm that \\ maintains both quality and diversity under this specialized MDP structure?}
\end{questionbox}
\end{center}
Our theoretical analysis reveals a surprising result under this scenario: the optimal actions can be derived by simply evaluating a fixed uniformly random policy and then selecting actions greedily based on its Q-values. 
This surprising finding means that we can bypass the standard GPI cycle to identify optimal policies, which requires only policy evaluation of the simplest possible policy (uniformly random), without iterative evaluation of the updated policy and without the many heuristic tricks that plague current methods.
Although it was widely believed that this kind of uniform policy is trivial and cannot provide meaningful guidance for control~\citep{asadi2017alternative}, 
the value of uniform policies~\citep{he2025random} has been observed empirically in specific discrete environments~\citep{laidlaw2023bridging} recently, and we provide a first theoretical analysis to account for LLM math reasoning and leverage it as the foundation of our approach. 

However, as in standard reward-maximizing RL, while a naive greedy selection guarantees optimality, it sacrifices diversity critical for reasoning tasks~\citep{si2024can}. To balance quality and diversity, we leverage a key insight based on our analysis: uniform-policy Q-values capture the probability of successful continuations that lead to positive rewards. As this creates a natural value map of the reasoning landscape,
we sample actions via
softmax over the uniform-policy Q-values, which maintains performance guarantees while aligning with modern LLM practices~\citep{sheng2024hybridflow,kwon2023efficient}.
To translate our theoretical insights into a practical and scalable algorithm for LLM reasoning, which involves vast state and action spaces as well as long horizons (a wide and deep tree), we present \underline{\textbf{R}}andom P\underline{\textbf{o}}licy \underline{\textbf{V}}aluation for Di\underline{\textbf{v}}erse \underline{\textbf{R}}easoning (ROVER).
ROVER efficiently parameterizes the Q-function intrinsically based on the LLM's parameters, which eliminates the need for a separate value network and also leverages the LLM's strong priors for efficient navigation in the vast token space and stabilizing training through relative improvements. 
To mitigate the high variance caused by the reward signals, we leverage group reward centering inspired by \citet{naik2024reward}, and broadcast the reward to improve training efficiency.

Our contributions are as follows: (\romannumeral1) We prove a surprising result: in the deterministic tree-structured MDPs with binary terminal rewards that characterize math reasoning, the optimal action can be derived directly from Q-values evaluated under a uniformly random policy, a finding that fundamentally simplifies RL for this domain. (\romannumeral2) We introduce ROVER, a practical and minimalist RL algorithm that is scalable to LLM reasoning tasks
through a simplified framework compared to the current complicated methods. (\romannumeral3) Despite ROVER's radical simplification, extensive experiments across diverse tasks and various model scales demonstrate that it consistently achieves superior performance, yielding \textbf{+8.2} improvement on pass@1 and \textbf{+16.8} improvement on pass@256 on the competition-level AIME24, AIME25, and HMMT25 tasks. Interestingly, we observe \ours can find novel reasoning strategies absent from the base model and models trained through standard RL approaches (GRPO), thereby evidencing its potential to push the reasoning boundary. 

\section{Preliminaries}
\label{sec:preliminary}
\noindent \textbf{RL with Verifiable Rewards in LLMs.}
We investigate reinforcement learning (RL) for post-training LLMs with verifiable rewards, such as mathematical reasoning tasks. We formulate the problem as a Markov Decision Process (MDP), defined by a tuple $(\mathcal{S},\mathcal{V},\mathcal{R},\mathcal{P},\gamma,\mathcal{X})$. Here, the state space $\mathcal{S}$ denotes all finite-length strings formed by the concatenation
of elements in $\mathcal{V}$. The action space $\mathcal{V}$ is the vocabulary set. We set the discount factor $\gamma=1$ in practice. $\mathcal{R}:\mathcal{S}\times\mathcal{V}\to \mathbb{R}$ is the binary reward function, and $\mathcal{P}:\mathcal{S}\times\mathcal{V}\to\mathcal{S}$ is a deterministic transition function. At the beginning of each episode, a prompt $x$ is sampled from the initial state distribution $\mathcal{X}$. At each step $t$, the LLM selects an action $a_t\in\mathcal{V}$ according to $\pi_{\theta}(\cdot|s_t)$, and then transits to the next state $s_{t+1}=\{x,a_0,\cdots,a_t\}$ by concatenation. This autoregressive generation continues until forming an entire response $y=\{a_0,a_1,\cdots,a_{|y|-1}\}$, and finally receives a verifiable reward $r(x,y)\in\{0,1\}$.
The goal is to learn a policy $\pi^*=\arg\max_\pi\mathbb{E}_{x\sim\mathcal{X},y \sim \pi(x)}\big[r(x,y)]$ 
by maximizing the expected cumulative reward $r$.
The prevailing works leverage policy gradient~\citep{williams1992simple} and a surrogate objective introduced by PPO~\citep{schulman2017proximal} to optimize $\pi_{\theta}$:
\begin{equation}
    \resizebox{.93\hsize}{!}{$J(\theta)=\mathbb{E}_{x\sim\mathcal{X},y \sim \pi_{\theta_{\rm old}}(x)}\big[\frac{1}{|y|}\sum\nolimits_{t=0}^{{|y|-1}}\big(\min\big({\rm IS}_t A_t, {\rm clip}({\rm IS}_t,1-\epsilon_{\rm low},1+\epsilon_{\rm high})A_t\big)-\beta D_{KL}(\pi_\theta|\pi_{\rm ref})\big)\big],$}
    \vspace{-1mm}
\end{equation}
where ${\rm IS}_t={\pi_{\theta}(a_t|s_t)}/{\pi_{\theta_{\rm old}}(a_t|s_t)}$ is the importance sampling ratio, $\pi_{\theta_{\rm old}}$ is the behavior policy to sample data, $s_t=\{x,a_{<t}\}$ is current state, $\epsilon_{\rm low}$ and $\epsilon_{\rm high}$ is the clipping range of importance sampling ratios, $D_{KL}$ denotes the KL regularization term, and $A_t$ is the advantage of current action. $A_t$ is implemented differently across RL algorithms, such as REINFORCE++~\citep{hu2025reinforce++} and GRPO~\citep{guo2025deepseek}. For example, GRPO~\citep{guo2025deepseek} samples $G>1$ responses for each
prompt and estimates the advantage $A_t=\frac{r(x,y_i)-{\rm mean}(\{r(x,y_i)\}^G_{i=1})}{{\rm std}(\{r(x,y_i)\}^G_{i=1})}$ within each group to reduce variance. Notably, while existing policy optimization methods rely on a KL-divergence penalty ($D_{KL}$) to prevent catastrophic forgetting and maintain exploration during continual learning~\citep{liu2025prorl}, our approach achieves these desiderata without such an explicit regularization term.

\begin{wrapfigure}{r}{0.38\textwidth} \vspace{-.15in}
    \centering
    \includegraphics[width=0.375\textwidth]{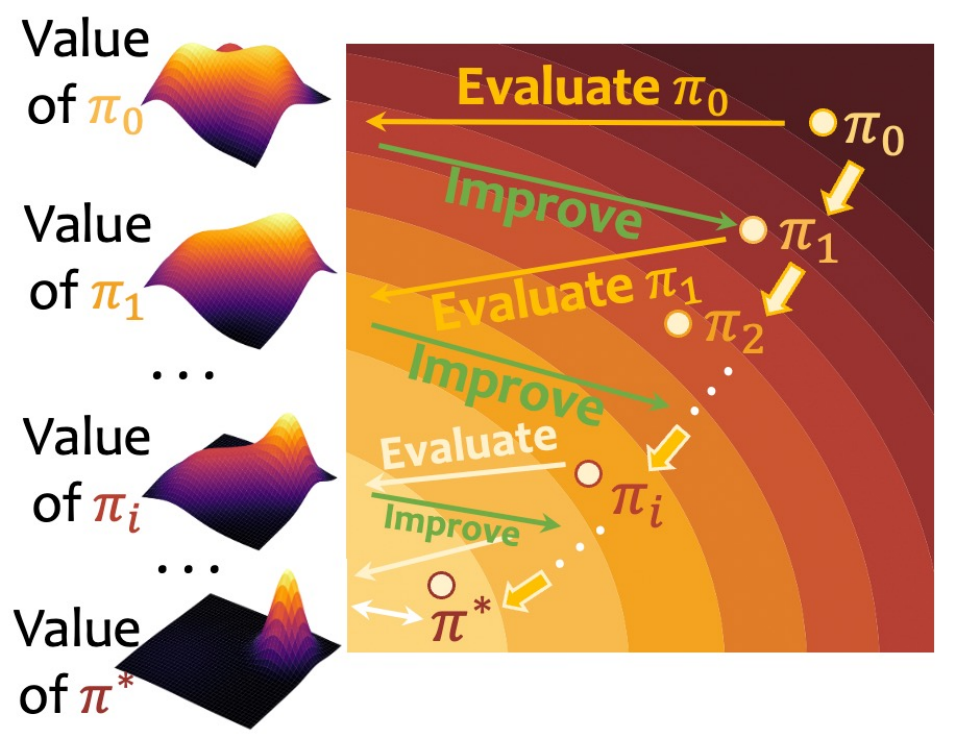}
    \vspace{-.3in}
    \caption{Illustration of GPI.}
    \label{fig:gpi}
    \vspace{-.2in}
\end{wrapfigure}
\noindent \textbf{Generalized Policy Iteration (GPI).} 
GPI~\citep{sutton1998reinforcement} is a unifying view that describes many RL algorithms (e.g., PPO) as illustrated in Fig.~\ref{fig:gpi}. GPI consists of two interacting processes, which are \emph{policy evaluation} that estimates how good a policy is, (e.g., via $Q^{\pi}(s_t, a_t) = r(s_t, a_t) + \gamma  \mathbb{E}\left[ Q^{\pi}(s_{t+1}, a_{t+1}) \right]$, value function, or advantage function), and \emph{policy improvement} that updates the policy to prefer actions scored better by the current estimates (e.g., $\pi(s) \leftarrow \arg\max_{a} Q^\pi(s, a)$ or other methods).
\citet{littman1996generalized} introduced generalized Bellman update which update the Q-function by $\hat{Q}(s_t, a_t) \leftarrow r(s_t, a_t) + \gamma \sum_{s_{t+1} \in \mathcal{S}} \gamma \mathcal{P}(s_t, a_t, s_{t+1}) \bigotimes_{a_{t+1}} \hat{Q}(s_{t+1}, a_{t+1})$ with any arbitrary operator $\bigotimes$ that replaces the max operator typically used in Q-learning~\citep{sutton1998reinforcement}. There have also been recent works studying improved operators for value estimation based on the softmax operator~\citep{soft-ql,asadi2017alternative,pan2020softmax}, while the mean operator was traditionally dismissed as unsuitable for optimization in general control tasks.
GPI-based methods require an alternative learning over these two processes until finding the fix point, where the learning target remains non-stationary throughout training~\citep{mnih2015human}. In contrast, our proposed method relies solely on \emph{policy evaluation} to derive the Q-values of a fixed, uniform random policy, which is much simpler for training and implementation (a high-level illustration is shown in Fig.~\ref{fig:rover_greedy}).

\section{ROVER: \textbf{R}andom P\textbf{o}licy \textbf{V}aluation for Div\textbf{e}rse \textbf{R}easoning}

\begin{wrapfigure}{r}{0.35\textwidth}
\vspace{-.18in}
    \centering
    \includegraphics[width=0.35\textwidth]{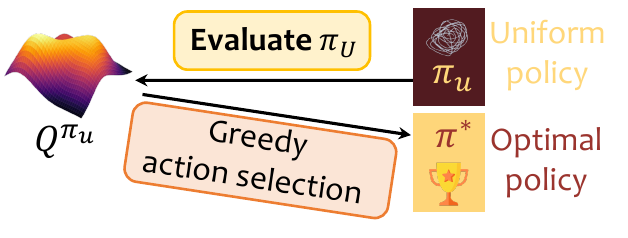}
    \vspace{-.15in}
    \caption{Illustration of ROVER (greedy).}
    \label{fig:rover_greedy}
    \vspace{-.25in}
\end{wrapfigure}
RLVR for math reasoning can be cast as a decision-making problem in a specialized finite-horizon MDP $\mathcal{M}$ with deterministic transitions and binary terminal rewards (correct or incorrect) in a tree-structured space (each state has a unique parent and actions lead to disjoint subtrees). This contrasts with general-purpose RL settings that often feature general control problems with stochastic dynamics, complex reward structures, and  discrete (or continuous) graph-based state spaces where states can have multiple parents or even cycles. 
Although the PPO family achieves promising results in LLM reasoning, it was designed for general control and can encounter entropy and diversity collapse in RLVR, which 
also introduces unnecessary computational overhead and complexity.
A comparison between traditional discrete RL tasks and RLVR tasks in LLM reasoning is summarized in Table~\ref{tab:rl-task-comparison}.

\begin{table}[!h]
    \centering
    \caption{Summarization of MDP structures between different tasks, considering the discrete Atari task from traditional RL and the countdown task from RLVR. While traditional RL tasks have smaller-scale spaces and shorter horizons (where RL agents typically train from scratch), the underlying MDP structure can be much more complex than RLVR tasks, which feature deterministic, episodic, tree-structured MDPs (which have larger spaces and longer horizons and leverage a powerful pre-trained model that can navigate in the large space).}
    \resizebox{\linewidth}{!}{
    \begin{tabular}{ccc}
    \toprule
    \textbf{Feature} & \textbf{Traditional RL Tasks (e.g., Atari)} & \textbf{RLVR Tasks (e.g., Countdown)} \\
    \midrule
    Transition dynamics & Stochastic/Deterministic & Deterministic \\
    Reward function & Stochastic/Deterministic & Deterministic \\
     & Intermediate/Episodic & Episodic \\
    State space structure & Graph-like (often cyclic) & Tree-like (no cycles) \\
    Action space & Smaller & Larger \\
    Horizon & Shorter & Longer \\
    Training & From Scratch & From powerful pre-trained model \\
    \bottomrule
    \end{tabular}
    }
    \label{tab:rl-task-comparison}
\end{table}

Motivated by this structural mismatch, 
we consider an important question overlooked in the literature: \emph{can there exist a minimalist and simple RL approach that exploits these properties of RLVR MDP to achieve both high quality and diversity?} In contrast to adding various implementation-level tricks to PPO/GRPO, we present ROVER,
which is built upon a surprising discovery: simply evaluating a uniformly random policy and selecting actions greedily based on its Q-values is sufficient for optimal behavior in this context (Fig.~\ref{fig:rover_greedy}), 
avoiding the complexities of modern deep RL algorithms~\citep{schulman2017proximal} and can bypass the traditional GPI loop in Fig.~\ref{fig:gpi}.

We first establish the theoretical basis of this unexpectedly simple yet optimal approach in \S~\ref{sec:theory}, extend it to achieve diversity while maintaining performance guarantees in \S~\ref{sec:softmax}, and present a practical algorithm that scales to large spaces and long horizons for math reasoning in \S~\ref{sec:implementation}.

\subsection{The Random Policy Valuation Framework}
\label{sec:theory}

We start from the simplest possible policy, the uniform random policy $\pi_{u}(a|s)=\frac{1}{|A|}$, where $A$ denotes the set of available actions. The corresponding Q-value for $\pi_u$ can be estimated using the generalized Bellman update~\citep{littman1996generalized,sutton1998reinforcement} 
with the mean operator~\citep{asadi2017alternative}. The mean operator corresponds to evaluating a uniform policy, and the update is simplified to Eq.~\ref{eq:rpe} for deterministic transitions and $\gamma=1$~\citep{hu2025openreasonerzeroopensourceapproach} that we consider as discussed in \S~\ref{sec:preliminary}.
\begin{equation}
\hat{Q}^{\pi_u}(s,a) \leftarrow r(s,a) + \frac{1}{|A|} \sum_{a' \in \mathcal{A}} \hat{Q}^{\pi_u}(s',a'). 
\label{eq:rpe}
\end{equation}
The literature of classical RL suggests that this mean operator is insufficient for optimal control in general MDPs~\citep{asadi2017alternative}, as it averages across all actions without preference for optimal ones, providing little guidance.
While a few recent studies have empirically noted the potential utility of uniform-policy values in certain discrete games~\citep{laidlaw2023bridging,he2025random}, 
these observations have remained primarily empirical, with limited theoretical justification. 

In our context, LLM math reasoning induces finite-horizon, deterministic, tree-structured MDPs with binary terminal rewards (correct/incorrect). For a root state $s_0=x$ (i.e., prompt), the reachable transition graph is a rooted tree, where each state has a unique path from $s_0$ and distinct actions from a state lead to disjoint subtrees.
Under this context, we prove that simply evaluating the fixed uniform policy and acting greedily with respect to its Q-values already achieves optimality in Theorem~\ref{theo:theorem1}.
The proof can be found in Appendix~\ref{app:theo_1}.

\begin{theorem}
\label{theo:theorem1}
Consider a finite-horizon episodic MDP with deterministic transitions, tree-structured state space, and binary terminal rewards $\mathcal{R}(s) \in \{0, R\}$ where $R > 0$ ($R$ for a correct solution, $0$ otherwise). Let $\pi_u$ be the uniform policy, and $Q^{\pi_u}$ its corresponding Q-function. Define the greedy policy with respect to $Q^{\pi_u}$ by $\pi_{\rm greedy}(s) = \arg\max_a Q^{\pi_u}(s,a)$, then $\pi_{\rm greedy}$ is optimal.
\end{theorem}

From Theorem~\ref{theo:theorem1}, we discover that for the specific MDP structure of LLM math reasoning, the optimal control problem reduces to a much simpler form than previously recognized. This suggests two significant implications: First, despite the perceived complexity of LLM math reasoning tasks, their underlying decision structure exhibits a more tractable structure than commonly assumed. Second, the mean operator, although generally insufficient for optimal control, proves to be surprisingly powerful when paired with a greedy action selection strategy in this context. 
 
\begin{wrapfigure}{r}{0.3\textwidth}
    \centering
    \vspace{-.4em}
    \includegraphics[width=0.3\textwidth]{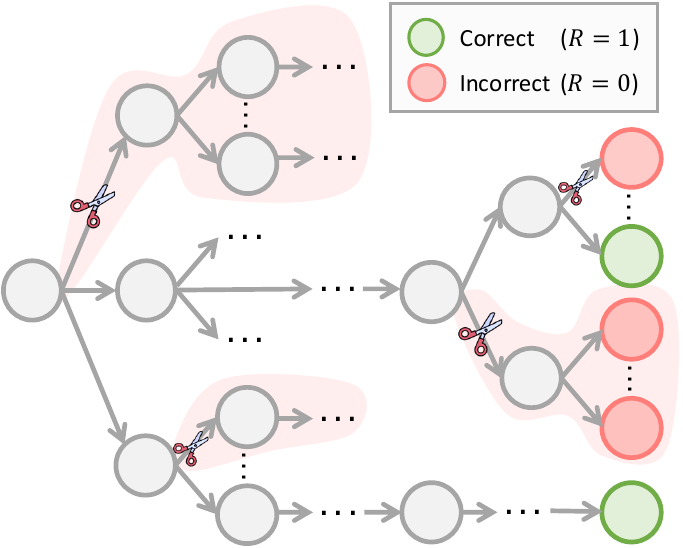}
    \vspace{-.15in}
    \caption{Intuition of ROVER (greedy) with $\pi_{\text{greedy}}$.} 
    \vspace{-.2in}
    \label{fig:theorem1}
\end{wrapfigure}

Surprisingly, although the uniformly random policy itself is far from optimal behavior, its Q-values have a meaningful interpretation here, which equals the probability that, after taking $a$ at $s$ and then acting uniformly at random until termination, we obtain a correct outcome.
As illustrated in Fig.~\ref{fig:theorem1}, when $Q^{\pi_u}(s,a)=0$, it indicates that no possible continuation from $(s,a)$ can lead to a correct solution. Conversely, higher values indicate more promising directions. By acting greedily with respect to these values, we effectively eliminate branches that cannot lead to valid solutions while prioritizing the most promising paths. This property enables optimality through a remarkably computationally simple mechanism: we need only estimate $Q^{\pi_u}(s,a)$ by policy evaluation for a fixed uniform policy $\pi_u$, without off-policy corrections or the implementation complexity of popular methods like PPO and GRPO. Additionally, since our approach evaluates a fixed uniform policy rather than iteratively improving a learned policy, it mitigates the non-stationarity issues that plague many modern deep RL methods~\citep{van2016deep}, which can also be advantageous for the high-dimensional, complex LLM math reasoning tasks.

\begin{figure}[!h]
    \centering
    \vspace{-1.2em}
    \subfigure[Toy MDP]{\includegraphics[width=0.17\linewidth]{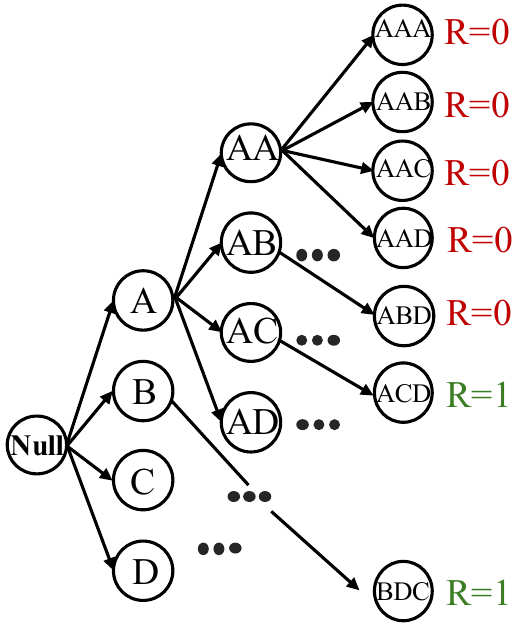}\label{fig:toy_mdp}}
    \subfigure[Q-learning]{\includegraphics[width=0.2\linewidth]{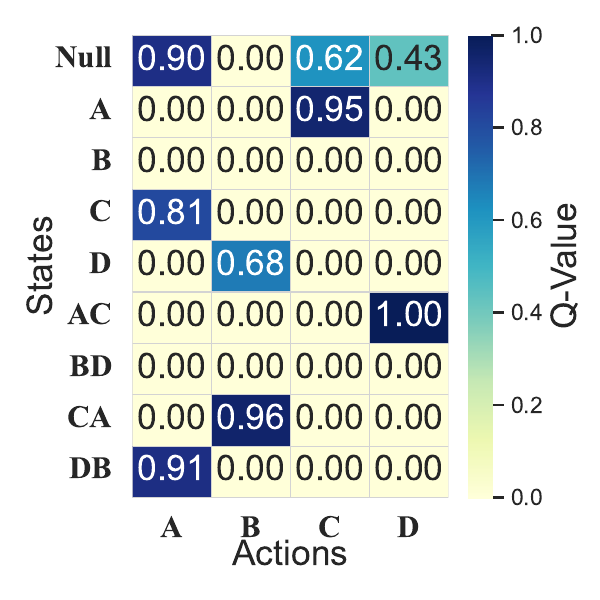}}
    \subfigure[ROVER (greedy)]{\includegraphics[width=0.2\linewidth]{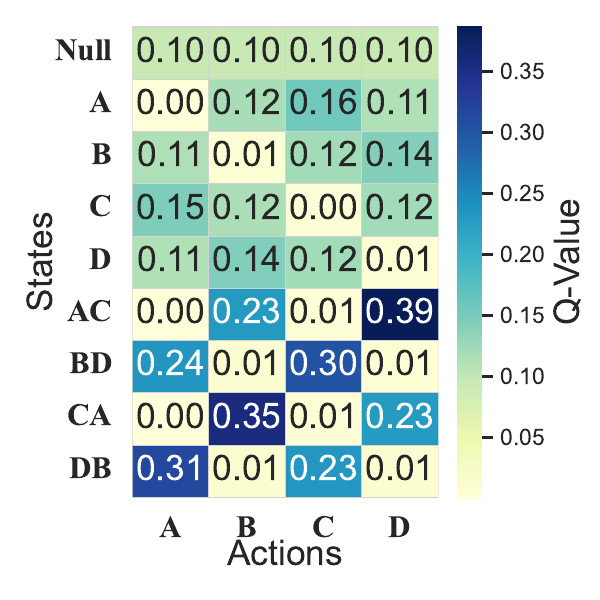}\label{fig:gorp}}
    \subfigure[ROVER]{\includegraphics[width=0.2\linewidth]{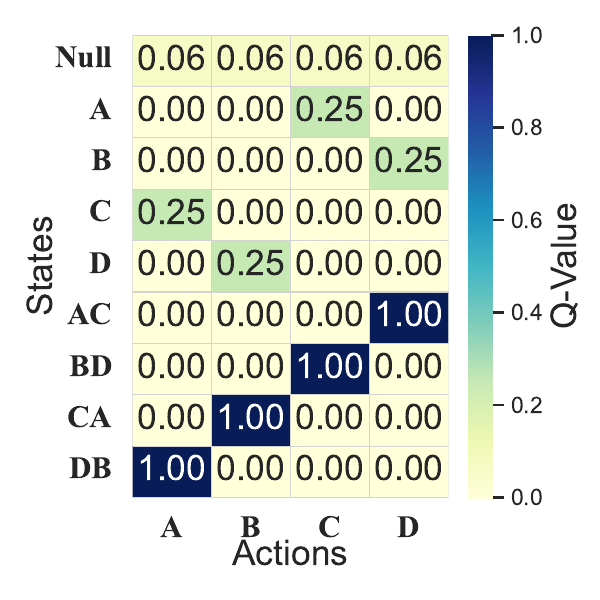}\label{fig:toy_rover}}
    \subfigure[Mode Coverage]{\includegraphics[width=0.2\linewidth]{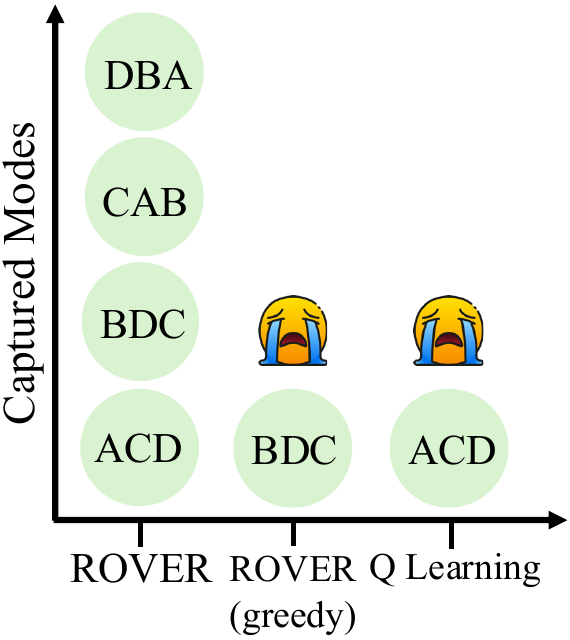}\label{fig:toy_mode}}
    \caption{(a) Illustration of the tabular MDP. (b)-(d) Comparison of learned Q-value maps. According to the Q-values, standard Q-learning with $\epsilon$-greedy exploration converges to the mode ACD.
    ROVER (greedy) assigns the highest Q-values to optimal actions, but still converges to a single mode BDC due to its greedy behavior. ROVER is able to assign equally high Q-values to all optimal actions. (e) Q-learning and ROVER (greedy) converge to a single mode despite both being optimal, whereas ROVER successfully covers all 4 optimal modes.}
    \label{fig:toy-case}
    \vspace{-.8em}
\end{figure}

\noindent \textbf{A Didactic Example.} To empirically validate the optimality of the greedy policy derived from the Q-function of a uniformly random policy, we design a tabular environment as illustrated in Fig.~\ref{fig:toy_mdp}. The environment is a deterministic, tree-structured MDP capturing the essential properties of LLM math reasoning tasks
while remaining transparent for analysis (and we will introduce how to scale up the method in \S~\ref{sec:implementation}). Starting from an initial null state, a policy executes an action $a\sim\mathcal{A}=\{A,B,C,D\}$ by appending it to the current state sequence. We consider an episodic setup with binary terminal rewards, with 4 specific terminal states (ACD, BDC, CAB, DBA) yielding a reward of 1 and all others yielding 0. 
From Fig.~\ref{fig:gorp}, we observe that the simple mechanism of acting greedily with respect to a random policy's Q-function also learns to generate the sequence with the highest reward, achieving the same optimal behavior as Q-learning (with $\epsilon$-greedy exploration).

\subsubsection{Beyond Greedy Selection: Balancing Quality and Diversity}
\label{sec:softmax}
While our theoretical analysis shows that the simple scheme of greedy selection over the Q-values of a uniform policy is already enough for achieving optimality, this deterministic approach often leads to mode collapse and sacrifices diversity (Fig.~\ref{fig:toy_mode}).
For LLM math reasoning tasks, as a given prompt can elicit multiple viable responses that yield correct solutions, diversity is critical for robust problem-solving~\citep{darling}, which is also important for improving pass@$k$ performance and generalization to novel problems. 

Our analysis reveals a key insight: the $Q^{\pi_u}(s,a)$ characterizes the probability of successful continuations following the action $a$, where higher Q-values indicate action branches with denser successful pathways.
To improve the diversity of policy generation, based on this insight, we transition from deterministic to stochastic action selection by converting $Q^{\pi_u}$ into a soft sampler, i.e., $\pi_s(a|s)=\frac{\exp(Q^{\pi_u}(s,a)/\rho)}{\sum_{a'} \exp(Q^{\pi_u}(s,a')/\rho)}$, where $\rho$ is a temperature parameter. 
This strategy selects actions proportional to their estimated success probability, which is able to explore multiple reasoning pathways for improving diversity, rather than committing to a single path.
Additionally, it aligns with contemporary LLM decoding strategies~\citep{kwon2023efficient}, making it readily integrable into existing training frameworks~\citep{sheng2024hybridflow}.
The following result shows that our softmaxing $Q^{\pi_u}$ approach maintains a guaranteed level of performance relative to the optimal policy, with the bound tightening as temperature decreases. The proof can be found in Appendix~\ref{app:theo_2}.

\begin{theorem}
\label{theo:pi_bound}
Consider the same MDP $\mathcal{M}$, and let $Q^{\pi_u}(s,a)$ denote the Q-function under the uniform random policy $\pi_u$ from state-action pair $(s,a)$, $N(s)=|\{a: Q^{\pi_u}(s,a)=0\}|$ be the number of zero-valued actions at state $s$, $A(s)$ be the number of available actions at state $s$, and $P$ denotes the set of key states where both optimal and suboptimal actions exist, i.e., $P=\{s: 1\leq N(s)\leq A(s)-1\}$.
Given the softmax policy $\pi_s(a|s)=\frac{\exp( Q^{\pi_u}(s,a)/\rho)}{\sum_{a'}\exp(Q^{\pi_u}(s,a')/\rho)}$ with temperature $\rho>0$, and $Pr^{\pi_s}(s|s_0)$ is the probability of reaching $s$ from $s_0$ with the policy $\pi_s$,  the value function of the induced policy $\pi_s$ satisfies: $V^{\pi_s} (s_0)\geq R\left( 1-\sum_{s \in P} Pr^{\pi_s}(s|s_0) \frac{N(s)}{N(s) + \exp(\max_a Q^{\pi_u}(s,a)/\rho)} \right)$.
\end{theorem}

Theorem~\ref{theo:pi_bound} characterizes that the temperature $\rho$ trades off between diversity and quality. As $\rho$ increases, the policy samples more diverse actions while still favoring higher-value paths. When $\rho$ approaches zero, the performance gap between the softmax policy and the optimal policy vanishes, showing that our diversity-promoting approach maintains performance guarantees.

\noindent \textbf{Justification.}
In our didactic example (Fig.~\ref{fig:toy_rover}\&\ref{fig:toy_mode}), we empirically demonstrate that it achieves an effective tradeoff. While both greedy approaches (Q-learning and ROVER (greedy)) achieve optimal reward but collapse to a single solution mode, ROVER (with $\rho=1$) successfully identifies all four optimal modes while maintaining 100\% success rate. Our diversity-seeking RL approach stands in contrast to typical RL diversity methods that often rely on complex and task-related reward engineering~\citep{he2025rewarding,cheng2025reasoning,darling} or post-hoc sampling techniques~\citep{shur2024growing,chen2025acereason} without guarantees, while remaining simple.

\subsection{Practical Implementation}
\label{sec:implementation}
We now adapt our method to LLMs, where the induced MDP still remains deterministic and tree-structured, but presents computational challenges due to long horizons (deep trees) and large vocabularies (wide branching). To address these challenges for making training practical, we introduce practical techniques for approximation, which stabilize the training process and improve sample efficiency as summarized in Alg.~\ref{alg:ours}, while preserving the core idea of random policy evaluation. We also provide gradient analysis and connections to policy-gradient methods in Appendix~\ref{app:gradient}.

\begin{algorithm}[t]
\label{alg:ours}
\caption{Random Policy Valuation for Diverse Reasoning (ROVER)}
\KwIn{pre-trained LLM $\pi_{\theta}$, epochs M, prompt dataset $\mathcal{D}$, group size $n$, lr $\eta$, temperature $\rho$}
\For{epoch $m=\{1,\cdots, M\}$}{
    Set $\pi_{\theta_{\rm old}}\gets \pi_\theta$; Sample a batch of prompts $\mathcal{B} \sim \mathcal{D}$ via $\pi_{\theta_{\rm old}}$ \\
    \For{each prompt $x \in \mathcal{B}$}{
        Rollout responses and compute rewards: $\{y_i\}_{i=1}^n \sim \pi_{\theta_{\rm old}}(\cdot|x)$; $\tilde{r} = r_i - \frac{1}{n}\sum_{i=1}^n r_i$
    }
    \For{each prompt-response pair $\{x,y_i\}$ in batch}{
        \For{each state $s_t \in \{x,y_i\}$}{
            Compute Q-value ${Q(a_{t+1}|s_{t+1})}= \rho\big(\log \pi_{\theta}(a_{t+1}|s_{t+1})-\log \pi_{\theta_{\rm old}}(a_{t+1}|s_{t+1})\big)$ \\
            Obtain $\hat{Q}(a_t|s_t) \leftarrow \tilde{r} + \frac{1}{|\mathcal{V}|}\sum_{a_{t+1}\in\mathcal{V}}{Q(a_{t+1}|{s_{t+1})}}$ \textcolor{gray}{// $\mathcal{V}$: the vocabulary set.}
        }
        $\mathcal{L}_{\text{ROVER}}=\frac{1}{\sum_{i=1}^n |y_i|}\sum_{i=1}^n\sum_{t=0}^{|y_i|-1}\Vert Q(a_{t}|s_{t}),{\text{sg}}[\hat{Q}(a_{t}| s_{t})]\Vert ^2$  \textcolor{gray}{// sg: stop gradient.} \\
        $\theta \leftarrow \theta - \eta \nabla_\theta \mathcal{L}_{\text{ROVER}}$ by an AdamW optimizer
    }
}
\end{algorithm}

\noindent $\bullet$ \textbf{Q Parameterization.}
While we begin with a reference LLM, we lack a pre-trained Q-function.
Training a Q-model from scratch presents substantial costs due to the large scale of action and state spaces. 
A compelling approach is to represent the Q-function directly through the LLM's intrinsic parameters $\theta$~\citep{li2025generalist}, thereby eliminating the need for a separate value network. 
Fortunately, as indicated in Theorem~\ref{theo:pi_bound} and following the mean operator for evaluating the value of the uniform policy in \S~\ref{sec:theory}, the Q-values $Q(s_t,a_t)$ and policy $\pi_\theta$ can be intrinsically linked through $\rho\log \pi_{\theta}(a_t|s_t)$ with $\rho$ the temperature, which captures the relative preference over actions within each state, though it omits a state-dependent constant term.
However, this direct formulation is unstable in practice since the learning target drifts as the policy changes and 
the Q-value updates are prone to divergence. 
To mitigate this instability, we introduce a relative Q-function that measures the improvement over a fixed baseline:
\begin{equation}
    Q(s_t,a_t)=\rho\big(\log \pi_{\theta}(a_t|s_t)-\log\pi_{\theta_{\rm old}}(a_t|s_t)\big),
\end{equation}
where $\pi_{\theta_{\rm old}}$ is the behavior policy used to sample data in each epoch, serving as a stable anchor that reduces fluctuations. This parameterization centers the initial Q-values around zero and ensures the model learns the change relative to the previous policy instead of absolute values.

\noindent $\bullet$ \textbf{Low-Variance Reward.} 
To create a stable and dense reward signal for learning uniform-policy Q-values, we sample $n$ responses for each prompt to reduce estimation variance and enrich our approximation of the value landscape. Inspired by \citet{naik2024reward}, we subtract the empirical average reward of the $n$ responses from the raw rewards to obtain mean-centered rewards. Specifically, the centered reward is given by 
\begin{equation}
    \tilde{r}(x,y_i)={r(x,y_i)-\frac{1}{n}\sum_{i=1}^n r(x,y_i)},
\end{equation}
where $r(x,y_i)$ reflects the correctness of the corresponding response $y_i$ given the prompt $x$. This is also related to GRPO's style of estimating the advantage function, but without the standard deviation normalization term~\citep{liu2025understanding}. Additionally, to ensure efficient credit assignment, especially for long reasoning chains, we broadcast this centered reward $\tilde{r}(x,y_i)$ to every token in the generation following \citet{hu2025openreasonerzeroopensourceapproach}. 

\section{Experiments}\label{sec:exp}
Although simple, our method substantially enhances both the quality and diversity of LLM generations, leading to improved reasoning capabilities on complex tasks. We evaluate our approach on two verifiable tasks that require sophisticated reasoning: countdown tasks, which have multiple valid answers, and math competitions, which possess single, unambiguous answers.
\subsection{Countdown Tasks}
We begin evaluating our method on the countdown task. Given an array of numbers and a target, the LLM must find the correct sequence using the four basic arithmetic operations ($+,-,\times,\div$) to reach the target number. We selected Countdown since it offers a restricted search space and multiple valid answers for a question that enables tractable analysis of both the reasoning behavior and diversity.

\noindent \textbf{Setup.} We evaluate on the TinyZero~\citep{tinyzero} dataset with 1,024 test problems. We employ Qwen2.5-3B~\citep{qwen2.5} as our base model, which demonstrates near-zero accuracy on this specific task that establishes a clear baseline for improvement. We benchmark our method against the well-recognized GRPO~\citep{shao2024deepseekmath} and two GRPO variants designed for policy entropy preservation: one with varying KL coefficients and another incorporating the clip-higher technique~\citep{yu2025dapo}. Detailed task descriptions and the training details are in Appendix~\ref{app:countdown}.

\begin{figure}[htbp]
\begin{minipage}[c]{0.68\textwidth}
\vspace{0pt}
    \centering
    \subfigure[Test Score]{\includegraphics[width=0.33\linewidth]{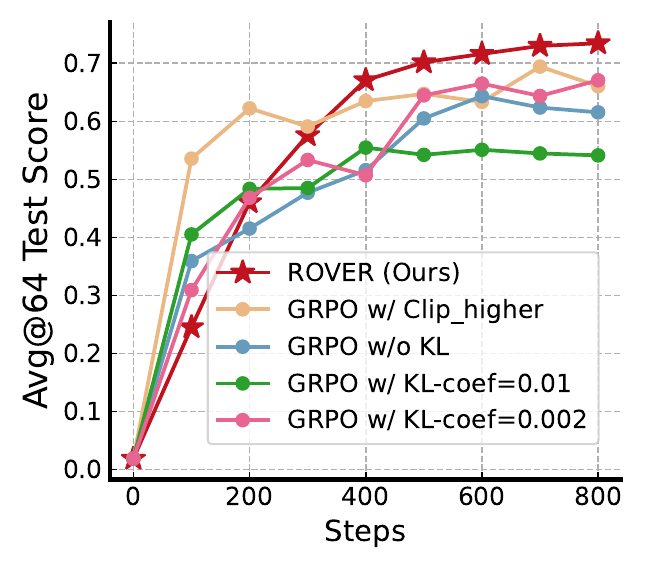}\label{fig:countdown-a}}
    \subfigure[Entropy]{\includegraphics[width=0.32\linewidth]{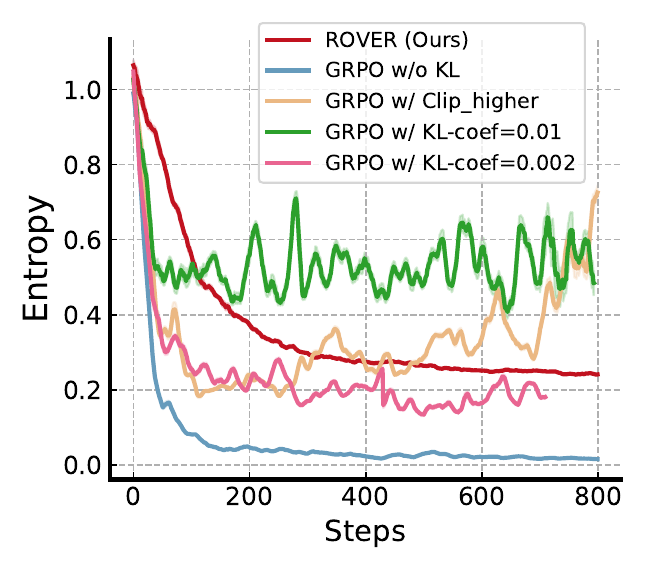}\label{fig:countdown-b}}
    \subfigure[Diversity\&Quality]{\includegraphics[width=0.33\linewidth]{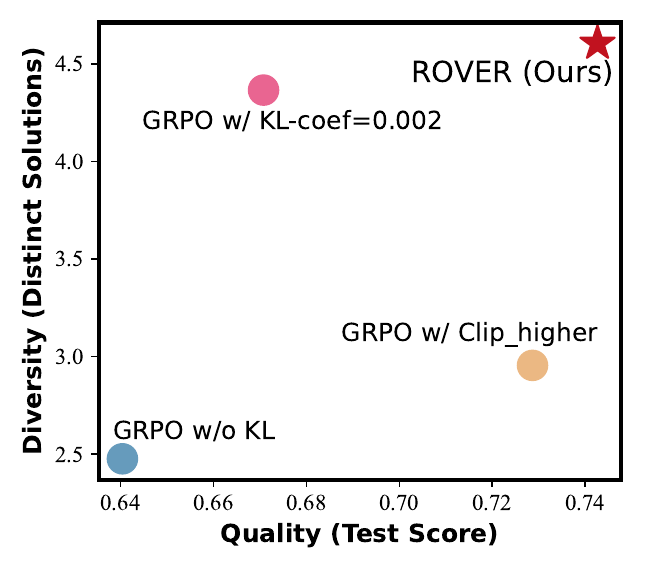}\label{fig:countdown-c}}
    \vspace{-1.5em}
    \caption{Performance of our method and baselines over training on countdown tasks. The y-axis of (c) denotes the number of found distinct correct solution equations, averaged over 1024 questions.}
    \label{fig:countdown}
\end{minipage}
\begin{minipage}[c]{0.31\textwidth}
\vspace{0pt}
    \centering
    \includegraphics[width=1\linewidth]{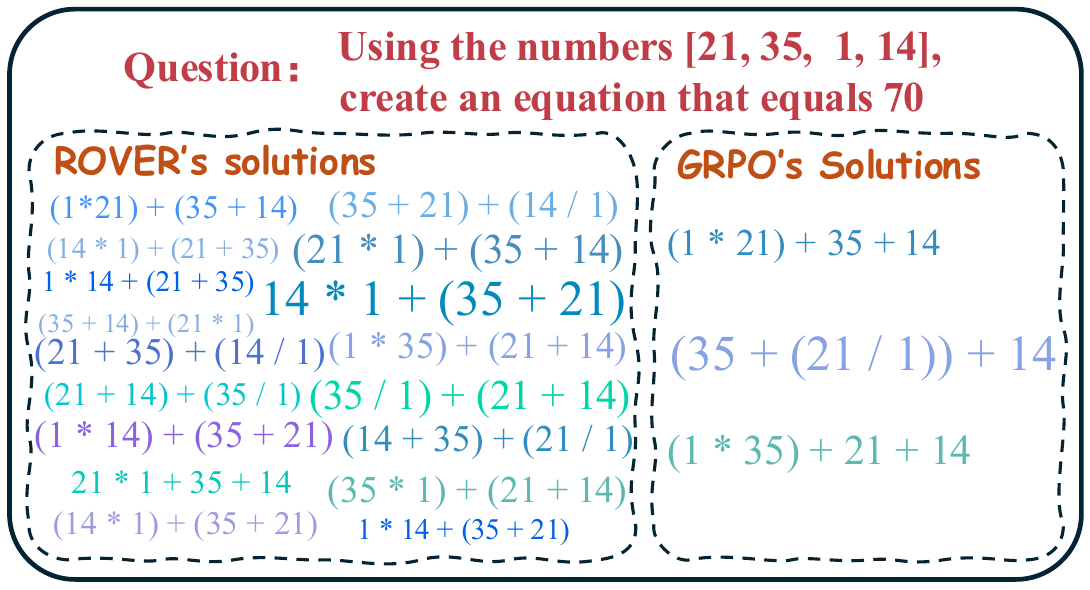}
    \vspace{-1.5em}
    \caption{ROVER successfully finds 17 diverse solution equations, while only 3 different equations are given by GRPO.}
    \label{fig:solution_diversity}
\end{minipage}
\end{figure}

\noindent \textbf{Results Analysis.} From the results shown in Fig.~\ref{fig:countdown}, we have the following observations: (\romannumeral1) In terms of test scores shown in Fig.~\ref{fig:countdown-a}, our method surpasses all baselines after 400 training steps, ultimately reaching the highest ceiling performance. Conversely, the GRPO with a KL coefficient of 0.01 performs distinctly worse, indicating that its performance is hampered by excessive regularization. We attribute the efficacy of our method to the preservation of high policy entropy throughout training. As shown in Fig.~\ref{fig:countdown-b}, our method's entropy decays gracefully while remaining significantly higher than that of the baselines, which either collapse (GRPO w/o KL) or fluctuate erratically (GRPO w/ Clip\_higher). A stable high entropy encourages sustained exploration, which is the primary driver of our model's performance, enabling it to achieve the highest scores on both quality and diversity metrics, as validated in Fig.~\ref{fig:countdown-c}, where our method finds more diverse solutions to address a question. Fig.~\ref{fig:solution_diversity} further provides a visualization example to demonstrate the solution diversity of \ours.

\begin{wrapfigure}{r}{0.425\textwidth}\vspace{-.2in}
    \centering
    \vspace{-.5em}
    \subfigure[Entropy]{\includegraphics[width=0.49\linewidth]{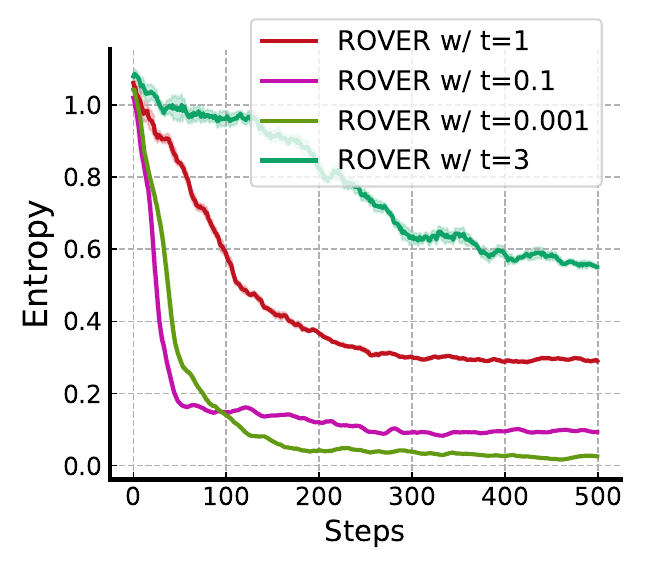}}
    \subfigure[Test Score]{\includegraphics[width=0.49\linewidth]{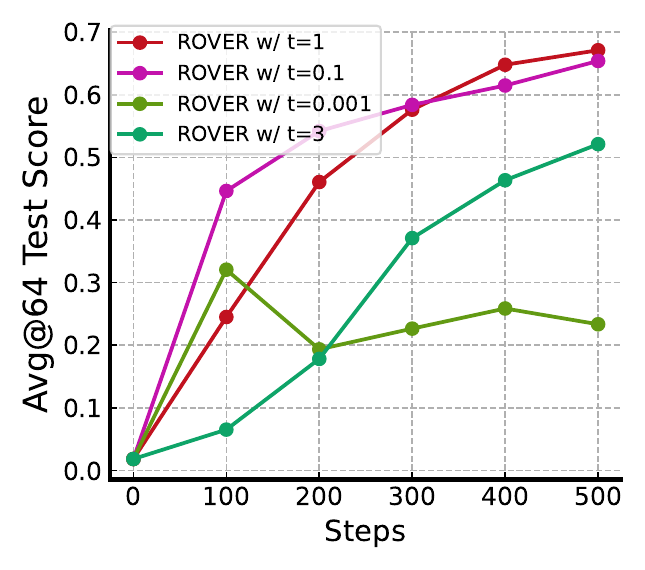}}
    \vspace{-1.3em}
    \caption{Performance under different $\rho$.}
    \label{fig:ablation_coundown_rho}
    \vspace{-1.5em}
\end{wrapfigure}

\noindent \textbf{Ablation on temperature $\rho$.} Consistent with standard LLM sampling practices~\citep{sheng2024hybridflow}, we set temperature $\rho=1$ for softmax sampling for all experiments without any task-specific tuning. This parameter balances the exploration-exploitation trade-off: $\rho\to0$ encourages greedy, deterministic behavior, while higher values promote diverse sampling. Our ablation study on $\rho$ in Fig.~\ref{fig:ablation_coundown_rho} confirms that $\rho=1$ achieves a robust and desirable performance. A higher temperature causes under-exploitation and slower convergence, while a lower value triggers premature exploitation, causing an accelerated collapse in policy entropy and constrained exploration space. In the extreme case where $\rho=0.001$, the near-deterministic policy sampling leads to severe training instability (evidenced in test score), highlighting the importance of a balanced temperature for effective exploration.
We further investigate the effect of $\rho$ on math reasoning tasks, where similar conclusions are validated. Results are provided in Appendix~\ref{app:abaltion_rho_math}.

\subsection{Reasoning on Math Tasks}
\noindent \textbf{Training Setup.} We employ models of various sizes for validating the efficacy of our proposed method, including {Qwen3-8B-Base}, {Qwen3-4B-Base}, and {DeepSeek-R1-Distill-Qwen-1.5B}, where the results of DeepSeek-1.5B are provided and analyzed in Appendix~\ref{app:deepseek_1.5b} due to space limitations. All models are trained on the open-source DeepScaler dataset~\citep{deepscaler2025}. A binary reward is assigned by the open-source verification tool \texttt{math\_verify}~\citep{mathverify2025} upon the completion of LLM generation. We employ standard RLVR methods as baselines, including PPO~\citep{schulman2017proximal}, GRPO~\citep{shao2024deepseekmath}, REINFORCE++~\citep{hu2025reinforce++}, and DAPO~\citep{yu2025dapo}.

\noindent \textbf{Evaluation.} We select various widely-acknowledged math reasoning benchmarks: AIME24~\citep{aime24}, AIME25~\citep{aime25}, HMMT25~\citep{balunovic2025matharena}, OlympiadBench~\citep{he2024olympiadbench}, AMC23~\citep{amc}, and MATH500~\citep{hendrycks2021measuring}, along with the O.O.D benchmark GPQA-diamond~\citep{rein2024gpqa}. We report pass@1 and pass@$k$ for comprehensive analysis, where pass@$k$ measures diversity and the reasoning boundary~\citep{yue2025does}. With increased diversity, the model has a higher probability of discovering a correct reasoning path within $k$ attempts. More details about the experimental setup can be found in the Appendix~\ref{app:train-eval-details}. 

\definecolor{red1}{RGB}{192,0,0}
\definecolor{green1}{RGB}{58,128,37}

\begin{table}[]
\caption{Pass@1 results across different methods on mathematical and O.O.D benchmarks. The highest and the second-best scores are shown in bold and underlined, respectively.}
\vspace{-.05em}
\resizebox{\linewidth}{!}{
\begin{tabular}{@{}clcccccccc@{}}
\toprule
\multicolumn{2}{c|}{\multirow{2}{*}[-1.5ex]{Pass@1}}                                              & \multicolumn{6}{c|}{Mathematical}                                                                                                                                                                                                                                                              & \multicolumn{1}{c|}{O.O.D}                                                  & \multirow{2}{*}[-1.5ex]{Avg.} \\ \cmidrule(lr){3-9}
\multicolumn{2}{c|}{}                                                                     & \begin{tabular}[c]{@{}c@{}}AIME\\ 2024\end{tabular} & \begin{tabular}[c]{@{}c@{}}AIME\\ 2025\end{tabular} & \begin{tabular}[c]{@{}c@{}}HMMT\\ 2025\end{tabular} & \multicolumn{1}{c}{\begin{tabular}[c]{@{}c@{}}Olympiad\\Bench\end{tabular}} & \begin{tabular}[c]{@{}c@{}}AMC\\ 2023\end{tabular} & \multicolumn{1}{c|}{\begin{tabular}[c]{@{}c@{}}MATH\\ 500\end{tabular}} & \multicolumn{1}{c|}{\begin{tabular}[c]{@{}c@{}}GPQA\\ diamond\end{tabular}} &                       \\ \midrule
\multicolumn{10}{c}{\emph{Qwen3-4B-Base}} \\
                                                                            & \gray{Base Model}  & \gray{8.8}                                                 & \gray{4.9}                                                 & \gray{0.8}                                                 & \gray{27.3}                                                                           & \gray{35.2}                                               & \gray{55.6}                                                                    & \gray{9.7}                                                                         & \gray{20.3}                     \\
                                                                            & PPO        & 15.7                                                & 8.3                                                & 1.6                                                 & 40.8                                                                           & 47.5                                               & 77.0                                                                    & 39.4                                                                        & 32.9                     \\
                                                                            & GRPO        & 16.4                                                & 9.4                                                 & 2.4                                                 & 43.6                                                                           & 57.0                                               & 79.9                                                                    & 38.7                                                                        & 35.3                     \\
                                                                            & DAPO        & 17.1                                                & 10.9                                                & 0.7                                                 & 41.7                                                                           & 56.6                                               & 78.4                                                                    & 38.5                                                                        & 34.8                     \\
                                                                            & REINFORCE++ & 14.8                                                & 7.8                                                 & 2.8                                                 & 42.3                                                                           & \textbf{57.9}                                      & 76.8                                                                    & 31.8                                                                        & 33.5                     \\
                                                                            & \cellcolor{blue!10}\textbf{\ours (Ours)}         & \cellcolor{blue!10}\textbf{17.6} \textcolor{green1}{\small $\uparrow$ $+$8.8}                                       & \cellcolor{blue!10}\textbf{12.6} \textcolor{green1}{\small $\uparrow$ $+$7.7}                                       & \cellcolor{blue!10}\textbf{3.1} \textcolor{green1}{\small $\uparrow$ $+$2.3}                                        & \cellcolor{blue!10}\textbf{45.4} \textcolor{green1}{\small $\uparrow$ $+$18.1}                                                              & \cellcolor{blue!10}\underline{57.1} \textcolor{green1}{\small $\uparrow$ $+$21.9}                                               & \cellcolor{blue!10}\textbf{80.5} \textcolor{green1}{\small $\uparrow$ $+$24.9}                                                           & \cellcolor{blue!10}\textbf{39.5} \textcolor{green1}{\small $\uparrow$ $+$29.8}                                                               & \cellcolor{blue!10}\textbf{36.5} \textcolor{green1}{\small $\uparrow$ $+$16.2}                     \\ \midrule
\multicolumn{10}{c}{\emph{Qwen3-8B-Base}} \\
                                                                            & \gray{Base Model}  & \gray{11.5}                                                & \gray{8.8}                                                 & \gray{0.8}                                                 & \gray{34.7}                                                                           & \gray{48.1}                                               & \gray{68.8}                                                                    & \gray{29.1}                                                                        & \gray{28.8}                     \\
                                                                            & PPO        & 15.3                                                & 12.4                                                & 5.0                                                 & 46.7                                                                           & 60.9                                               & 82.0                                                                    & 45.1                                                                        & 38.2                     \\
                                                                            & GRPO        & 16.8                                                & 15.1                                                & 4.8                                                 & 48.6                                                                           & 66.9                                               & 81.9                                                                    & 43.8                                                                        & 39.7                     \\
                                                                            & DAPO        & 20.8                                                & 15.2                                                & 3.6                                                 & 49.0                                                                           & 67.9                                               & 84.3                                                                    & 46.6                                                                        & 41.1                     \\
                                                                            & REINFORCE++ & 19.4                                                & 16.7                                                & 7.1                                                 & 47.6                                                                           & 63.5                                               & 83.6                                                                    & 46.3                                                                        & 40.6                     \\
                                                                            & \cellcolor{blue!10}\textbf{\ours (Ours)}         & \cellcolor{blue!10}\textbf{30.6} \textcolor{green1}{\small $\uparrow$ $+$19.1}                                      & \cellcolor{blue!10}\textbf{22.7} \textcolor{green1}{\small $\uparrow$ $+$13.9}                                       & \cellcolor{blue!10}\textbf{14.6} \textcolor{green1}{\small $\uparrow$ $+$13.8}                                       & \cellcolor{blue!10}\textbf{56.4}  \textcolor{green1}{\small $\uparrow$ $+$21.7}                                                             & \cellcolor{blue!10}\textbf{74.8} \textcolor{green1}{\small $\uparrow$ $+$26.7}                                      & \cellcolor{blue!10}\textbf{89.6} \textcolor{green1}{\small $\uparrow$ $+$20.8}                                                           & \cellcolor{blue!10}\textbf{50.2} \textcolor{green1}{\small $\uparrow$ $+$21.1}                                                               & \cellcolor{blue!10}\textbf{48.4} \textcolor{green1}{\small $\uparrow$ $+$19.6}                     
     \\ \bottomrule
\end{tabular}
}
\label{table:pass1}
\vspace{-1.2em}
\end{table}

\subsubsection{Performance Analysis}

\noindent \textbf{ROVER consistently outperforms all RL baselines in terms of average pass@1.} As detailed in Table~\ref{table:pass1}, ROVER consistently outperforms standard RL methods across all model sizes. For the {Qwen3-8B-Base} model, ROVER achieves pass@1 improvements of \textbf{+7.3} and \textbf{+8.2} over the strongest baseline, averaged on all benchmarks and on the subset of AIME24, AIME25 and HMMT25, respectively.
The superiority of our method over baseline methods becomes more pronounced on increasingly challenging tasks. Notably, for {Qwen3-8B-Base}, ROVER delivers substantial relative improvements of \textbf{+47.1\%} on AIME24 and \textbf{+35.9\%} on AIME25 over the best-performing baseline. On HMMT25, \ours nearly doubles the performance of the strongest baseline, REINFORCE++. 

\begin{figure}[htbp]
    \centering
    \subfigure[AIME 2024]{\includegraphics[width=0.32\linewidth]{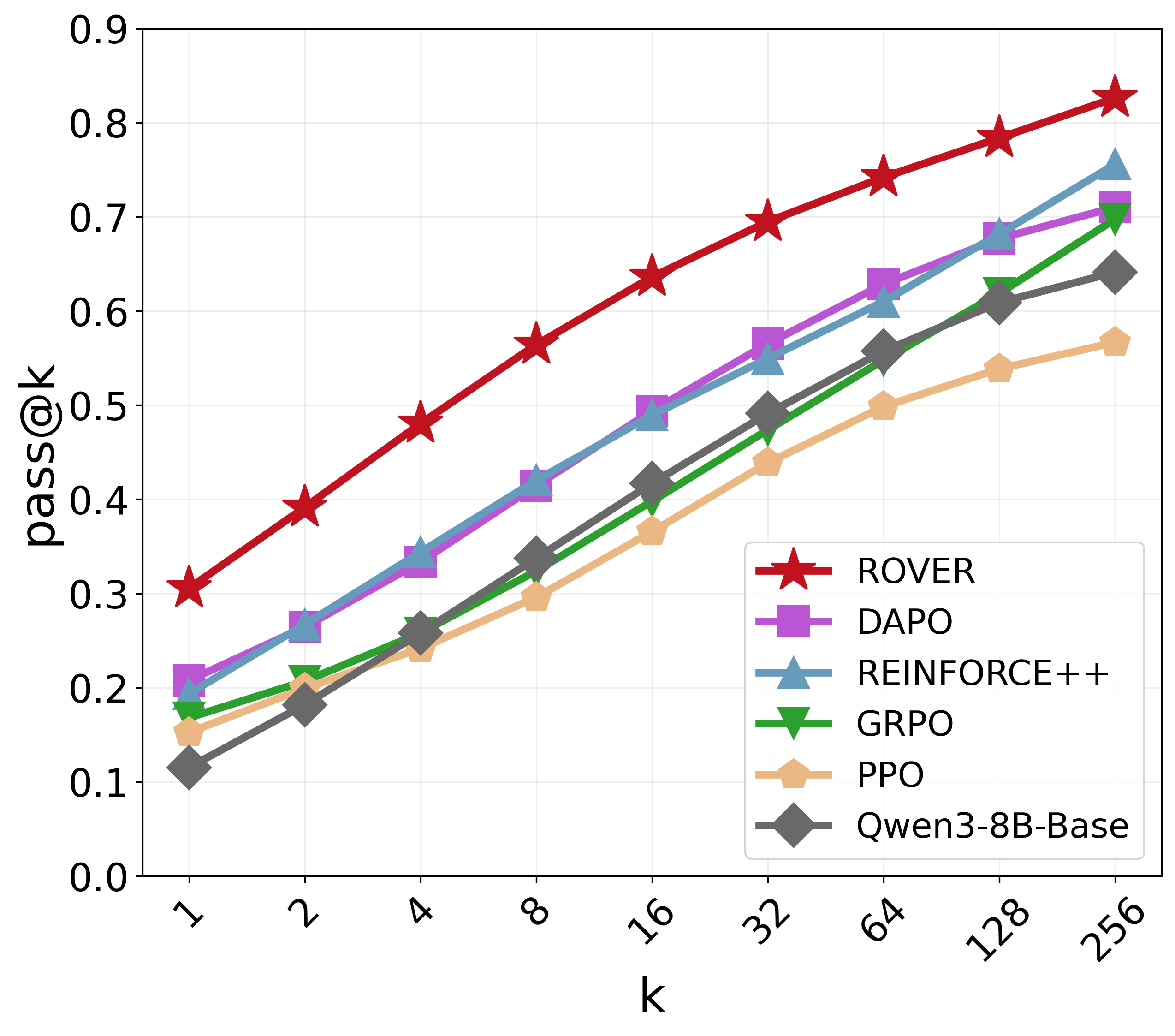}}
    \subfigure[AIME 2025]{\includegraphics[width=0.32\linewidth]{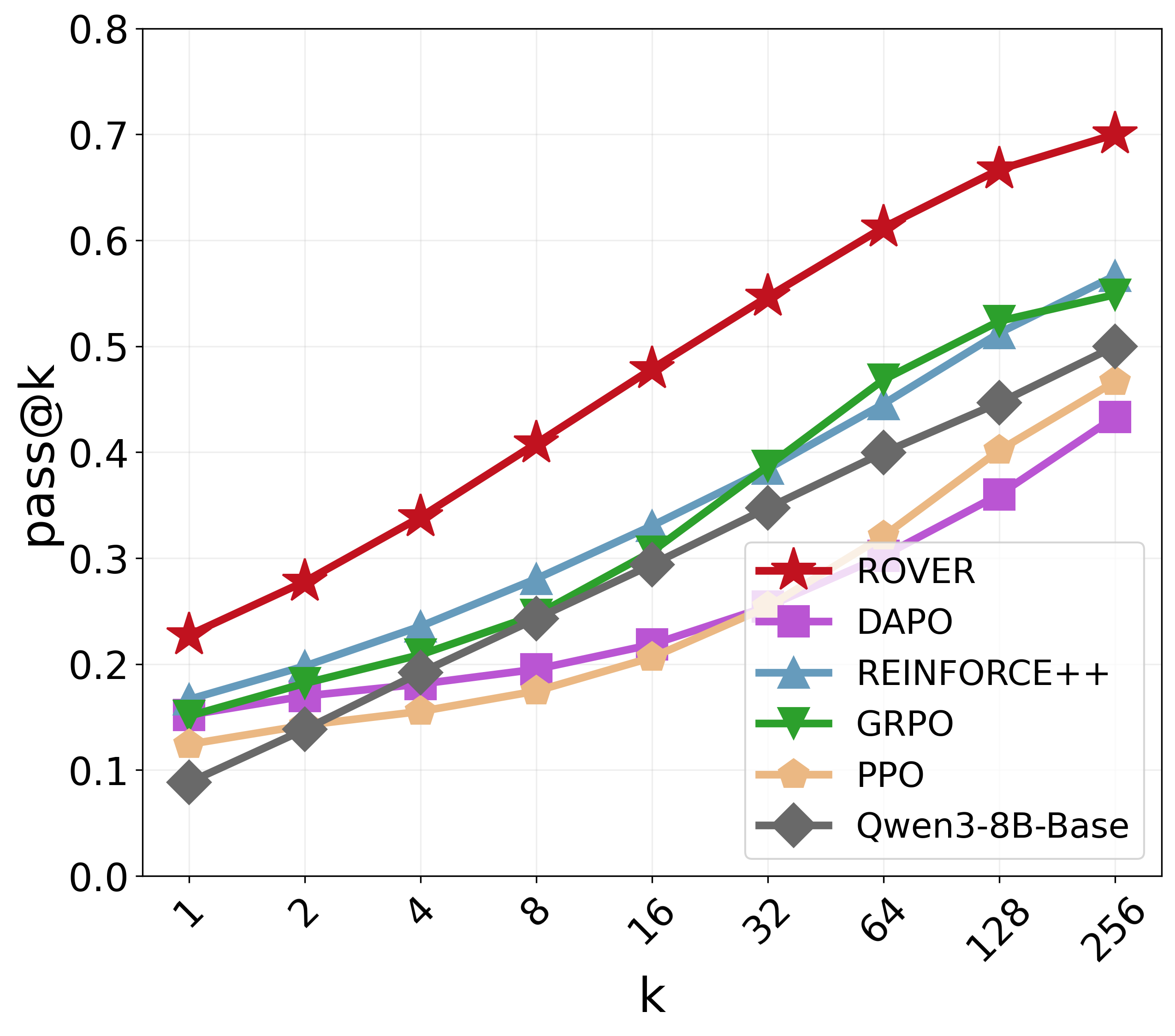}}
    \subfigure[HMMT 2025]{\includegraphics[width=0.32\linewidth]{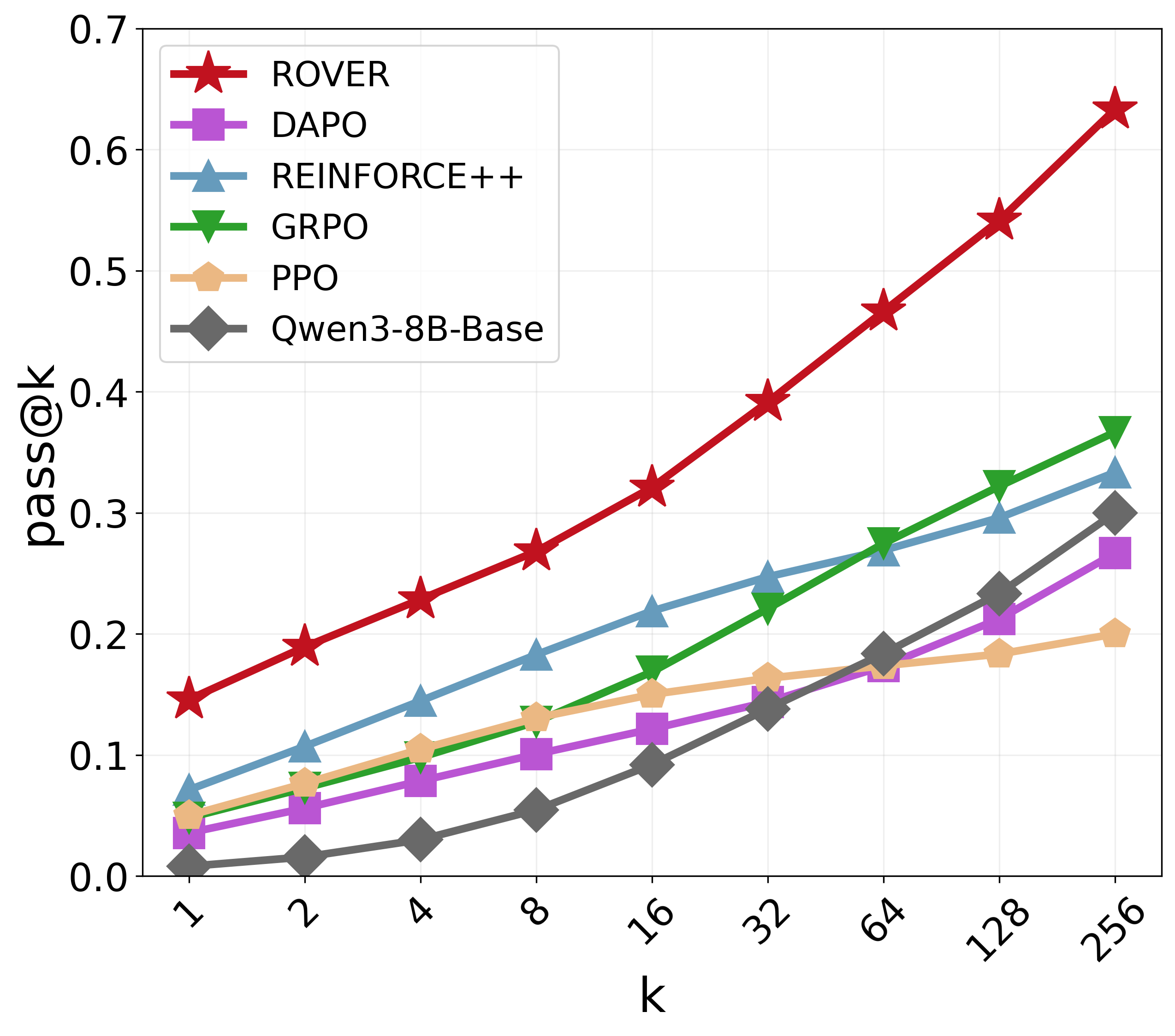}}
    \caption{pass@$k$ of \ours and baselines on {Qwen3-8B-Base}.}
    \label{fig:passk-8b}
\end{figure}

\noindent \textbf{ROVER significantly improves pass@$k$}. The average pass@$k$ over a dataset reflects the proportion of problems a model can potentially solve within $k$ trials, serving as a robust evaluation metric of the model's reasoning breadth and diversity. 
To demonstrate the effectiveness of our method in incentivizing reasoning diversity, we compare ROVER with baselines by scaling pass@$k$ from 1 to 256. Consistent with previous observations~\citep{yue2025does,darling}, the results in Fig.~\ref{fig:passk-8b} reveal that while standard RL baseline methods enhance pass@1, their performance quickly saturates and plateaus, ultimately underperforming the base model at large $k$ values.
For example, DAPO even shows worse performance on AIME25 after $k>4$, a trend that is also observed on HMMT25 for $k>32$. In contrast, our method demonstrates sustained and significant performance gains as $k$ increases, consistently surpassing all the baselines and the base model (\textbf{+16.8} over the best baseline on pass@256 averaged on AIME24, AIME25, and HMMT25). This advantage is particularly pronounced on the most challenging HMMT25 task, where our method's pass@$k$ score continues to accelerate while all baselines have saturated. We attribute the improved pass@$k$ to ROVER's ability to maintain a relatively higher entropy during training (see Fig.~\ref{fig:qwen-entropy}), which ensures sustained exploration of different reasoning strategies and enhances reasoning diversity.

\noindent \textbf{ROVER shows remarkable generalization on O.O.D tasks}. To further evaluate the generalization capability of ROVER, we incorporate the GPQA-diamond benchmark, a challenging math-unrelated task containing 198 graduate-level questions in biology, physics, and chemistry. The results in Table~\ref{table:pass1} demonstrate ROVER's stronger generalization beyond the training distribution, achieving the best performance on the unseen GPQA-diamond benchmark.

\subsubsection{Diversity Analysis}
\vspace{-.5em}
\begin{wrapfigure}{r}{0.3\textwidth}
\vspace{-1.5em}
\centering
    \centering
    {\includegraphics[width=1\linewidth]{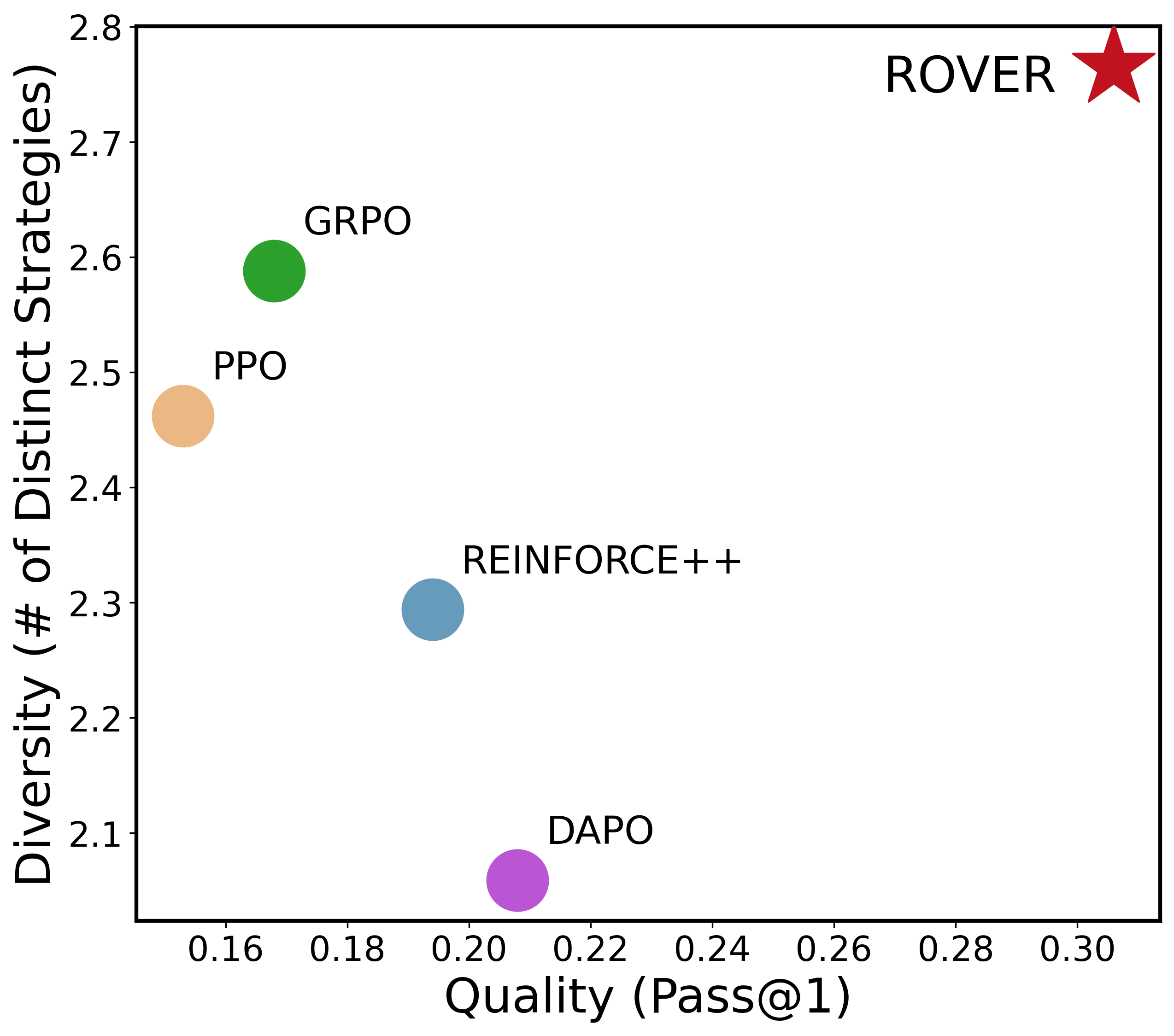}}
    \vspace{-.2in}
    \caption{Quality-Diversity tradeoff.}
    \label{fig:quality-diversity-trade-off}
    \vspace{-1.5em}
\end{wrapfigure}
\noindent \textbf{ROVER possesses the highest diversity across different metrics}. To quantify reasoning diversity, we employ the ``number of distinct strategies" metric from NoveltyBench~\citep{zhang2025noveltybench}. Specifically, we sample up to 32 correct responses for each problem from the AIME24 datasets, and leverage Claude-3.5-Sonnet as the LLM judger to determine strategic equivalence between these response pairs (template in Fig.~\ref{fig:prompt-llm-judge}). A higher number of distinct strategies (classes) indicates greater reasoning diversity. 
We report the results in Fig.~\ref{fig:quality-diversity-trade-off} (with a 0.6 decoding temperature) and the results across different decoding temperatures in Fig.~\ref{fig:quality-diversity-trade-off-temperature}.
From Fig.~\ref{fig:quality-diversity-trade-off}, we observe that ROVER demonstrates relative diversity improvements of \textbf{+6.8\%} and \textbf{+17.6\%} when compared with GRPO and the average of all baselines, respectively.
Conventional RL approaches struggle to improve diversity merely through increasing sampling temperature during inference, while \ours consistently improves the Pareto front between quality and diversity.
For a more comprehensive quantitative analysis of generation diversity, we refer to Appendix~\ref{app:diversity}, which includes results for additional metrics such as utility~\citep{zhang2025noveltybench} and cosine distance (Fig.~\ref{fig:radar-all-temperature}).

\subsubsection{Behavioral Analysis}
\vspace{-.5em}
\begin{wrapfigure}{r}{0.52\textwidth}
\vspace{-1.5em}
\centering
    \centering
    \subfigure[AIME 2024]{\includegraphics[width=0.49\linewidth]{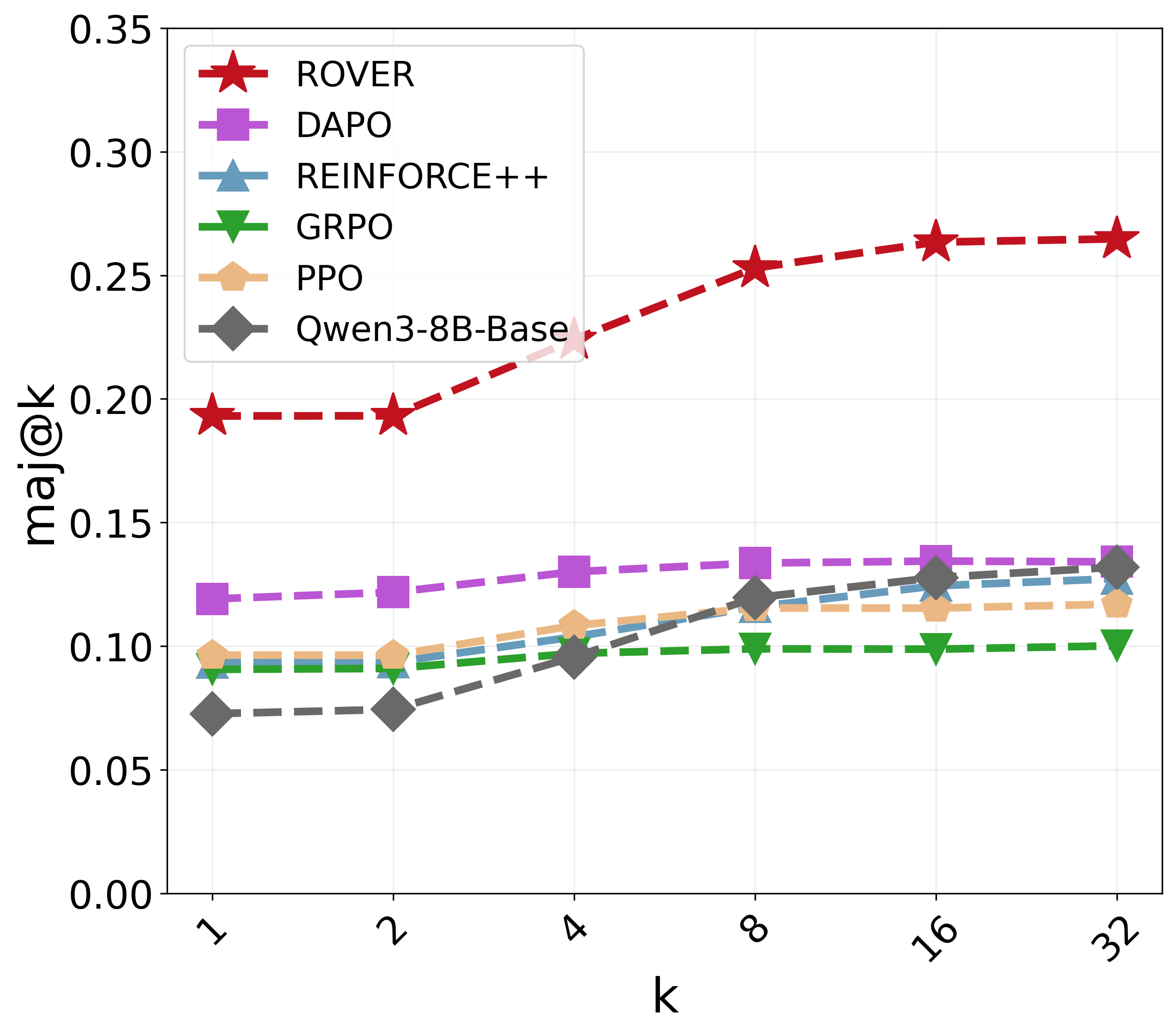}}
    \subfigure[HMMT 2025]{\includegraphics[width=0.49\linewidth]{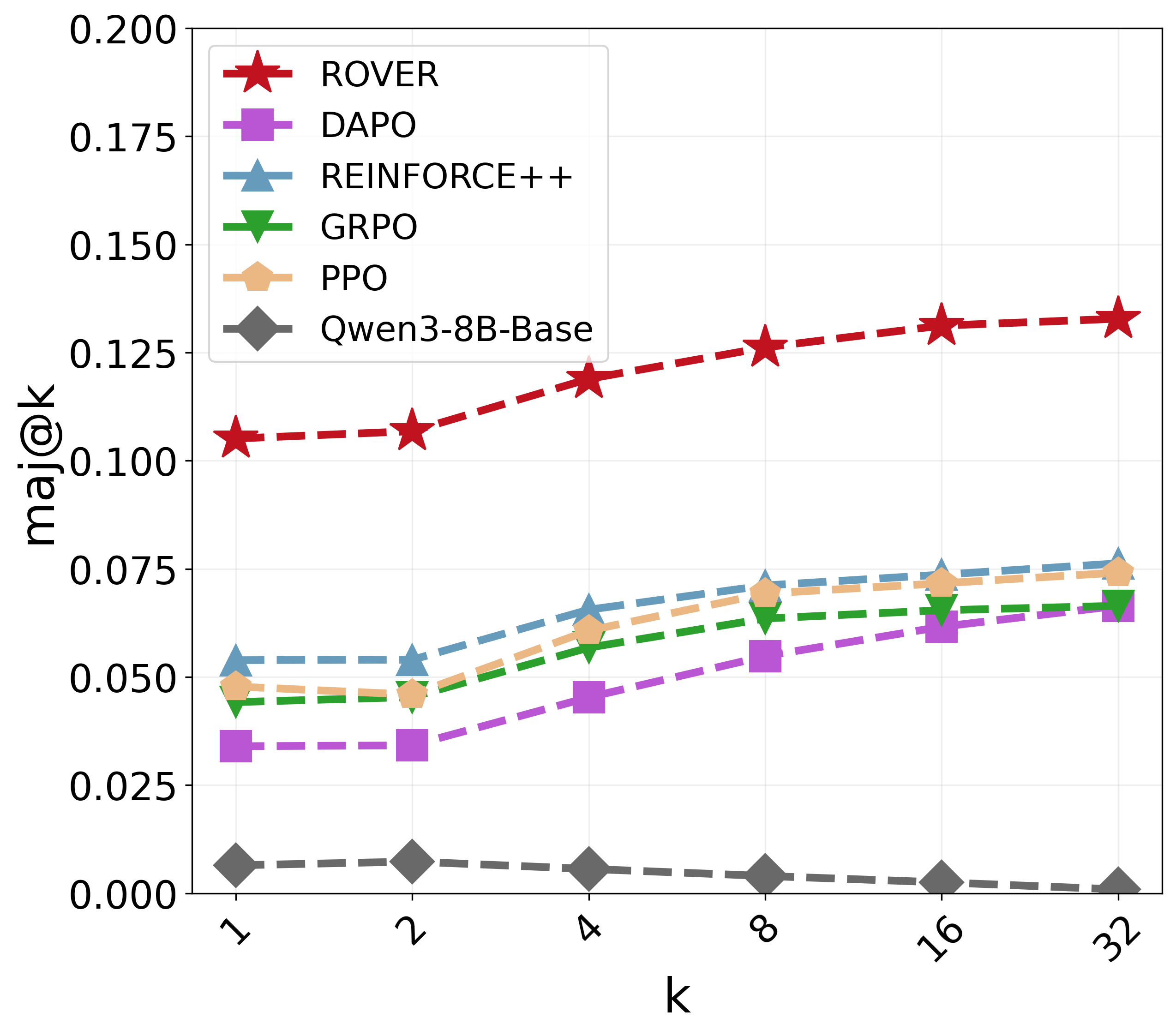}}
    \vspace{-1.em}
    \caption{Maj@$k$ performance of \ours and baselines on {Qwen3-8B-Base}.}
    \label{fig:majk-8b}
    \vspace{-1.em}
\end{wrapfigure}
\noindent \textbf{ROVER scales best at test-time due to maintained diversity}. Test-time scaling has received significant attention due to its potential to enhance reasoning performance, where majority voting is a fundamental baseline for evaluating LLM scalability at test-time~\citep{liu2025can}. Fig.~\ref{fig:majk-8b} confirms that ROVER's maj@$k$ performance scales robustly, consistently improving upon the base model across all $k$ values, even on the most challenging HMMT25 task. This superior scalability stems from ROVER's ability to maintain a diverse distribution over valid reasoning paths, while baseline methods suffer from mode collapse, causing them to confidently converge on similar incorrect solutions and preventing performance gains from additional samples.

\begin{wrapfigure}{r}{0.35\textwidth}
    \centering
    \vspace{-.5em}
    \includegraphics[width=\linewidth]{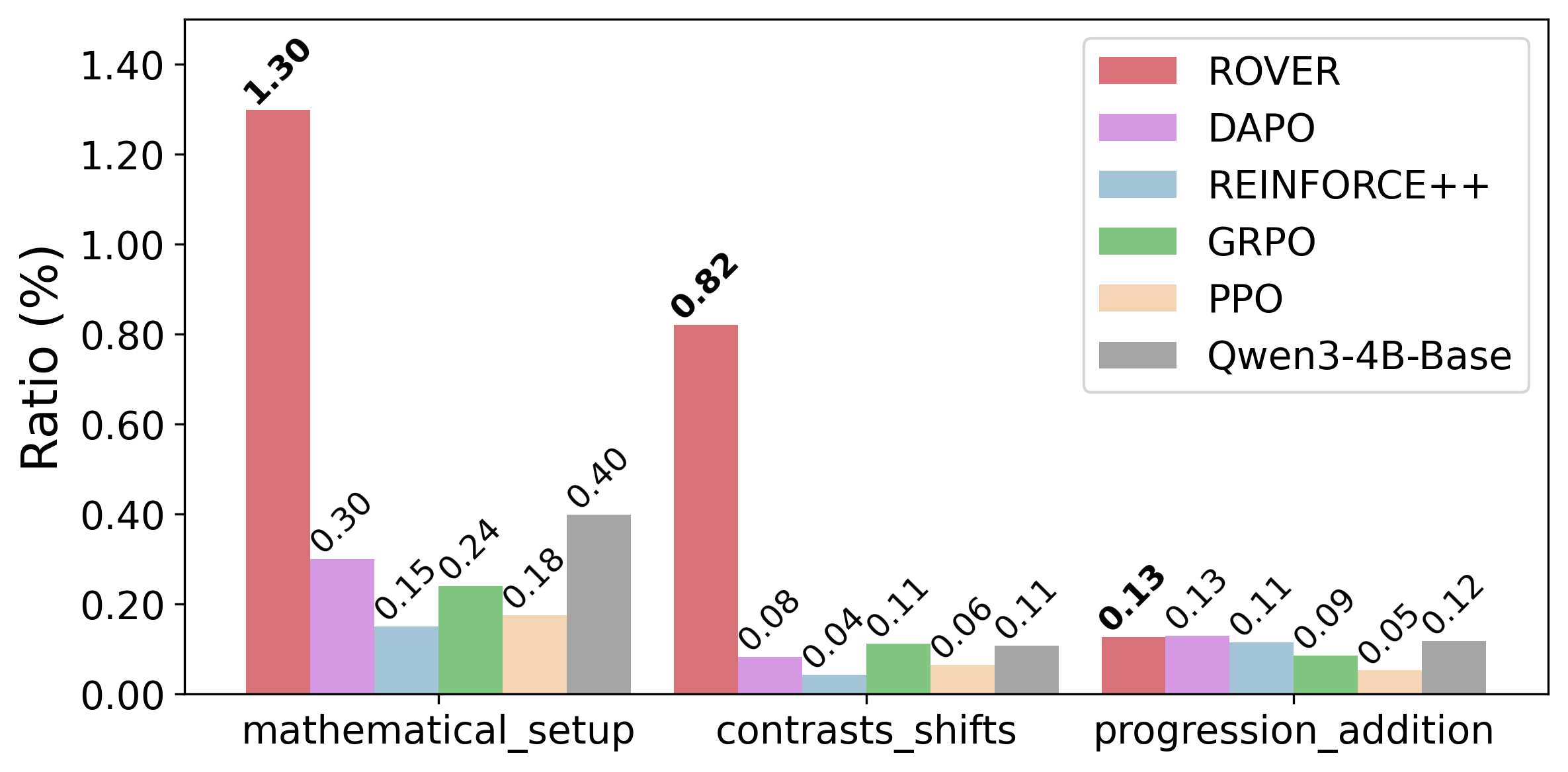}
    \vspace{-1em}
    \caption{Comparison of reflection frequency. ROVER outputs more reasoning-related tokens.}
    \label{fig:reflection_analysis}
    \vspace{-1.5em}
\end{wrapfigure}

\begin{figure}[tb]
    \centering
    \includegraphics[width=1.0\linewidth]{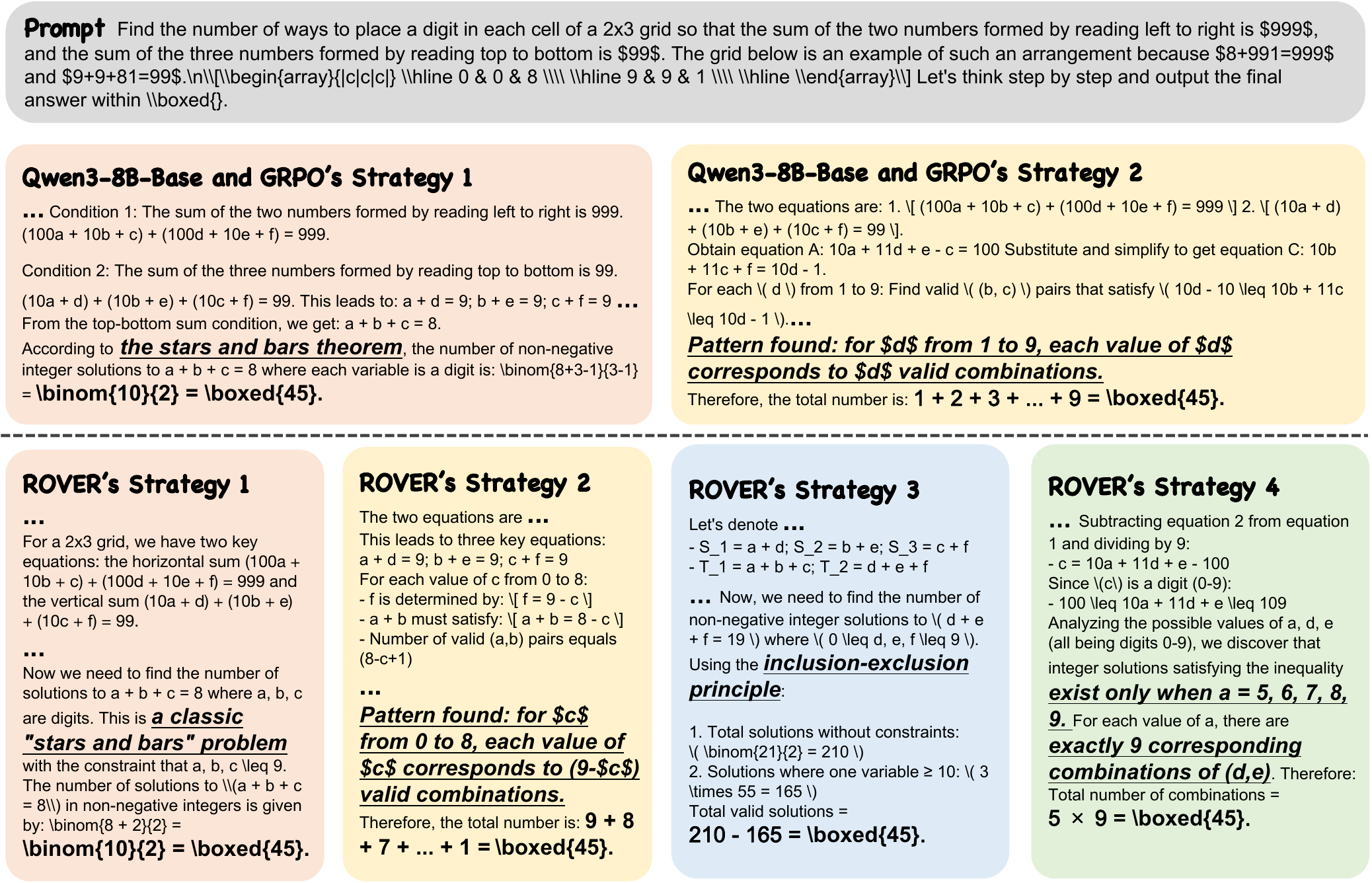}
    \vspace{-1.5em}
    \caption{Illustration of strategies discovered by Qwen3-8B-Base, GRPO and \ours. For each model, 32 samples are generated and subsequently clustered based on strategic equivalence using an LLM judge (the prompt of the LLM judge is given by Fig.~\ref{fig:prompt-llm-judge}). Responses sharing the same color represent strategically identical approaches. While {Qwen3-8B-Base} and GRPO find two distinct strategies 1\&2, \ours not only discovers the same two strategies but also uncovers two additional strategies 3\&4. For example, beyond discovering the \emph{Stars and Bars theorem} (strategy 1), \ours also discovered a solution based on the \emph{inclusion-exclusion principle} (strategy 3), which demonstrates \ours's capability in pushing reasoning boundaries.}
    \vspace{-.15in}
    \label{fig:demo}
    \vspace{-1.em}
\end{figure}

\noindent \textbf{Enhanced reflection behaviors}. To analyze the reasoning patterns learned via ROVER, we adopt the \emph{forking tokens} defined in \citet{wang2025beyond} (see Table~\ref{tab:forking_tokens}) and quantify the normalized frequency of these tokens in the generated outputs (256 rollouts per prompt on AIME24, AIME25, and HMMT25).
Fig.~\ref{fig:reflection_analysis} shows models trained with ROVER generate a significantly higher proportion of these \emph{forking tokens}, particularly those associated with rethinking and self-correction (e.g., `wait' and `however'). As detailed in Fig.~\ref{fig:case_study_token_prob_diff}, ROVER encourages the model to actively reflect upon, verify, and pivot between different reasoning strategies, rather than committing to a single reasoning path.

\noindent \textbf{Discovered strategies comparison.} To intuitively show the enhanced diversity of \ours, we present a representative prompt from AIME24 that holds multiple potentially feasible strategies. Representative CoT examples for each cluster are illustrated in Fig.~\ref{fig:demo}, where \ours discovers two additional novel strategies compared to the base and GRPO-trained models.

\subsubsection{Ablation Studies}\label{sec:ablation}
\begin{figure*}[htbp]
    \centering
    \vspace{-.10in}
    \subfigure[]{\includegraphics[width=0.26\linewidth]{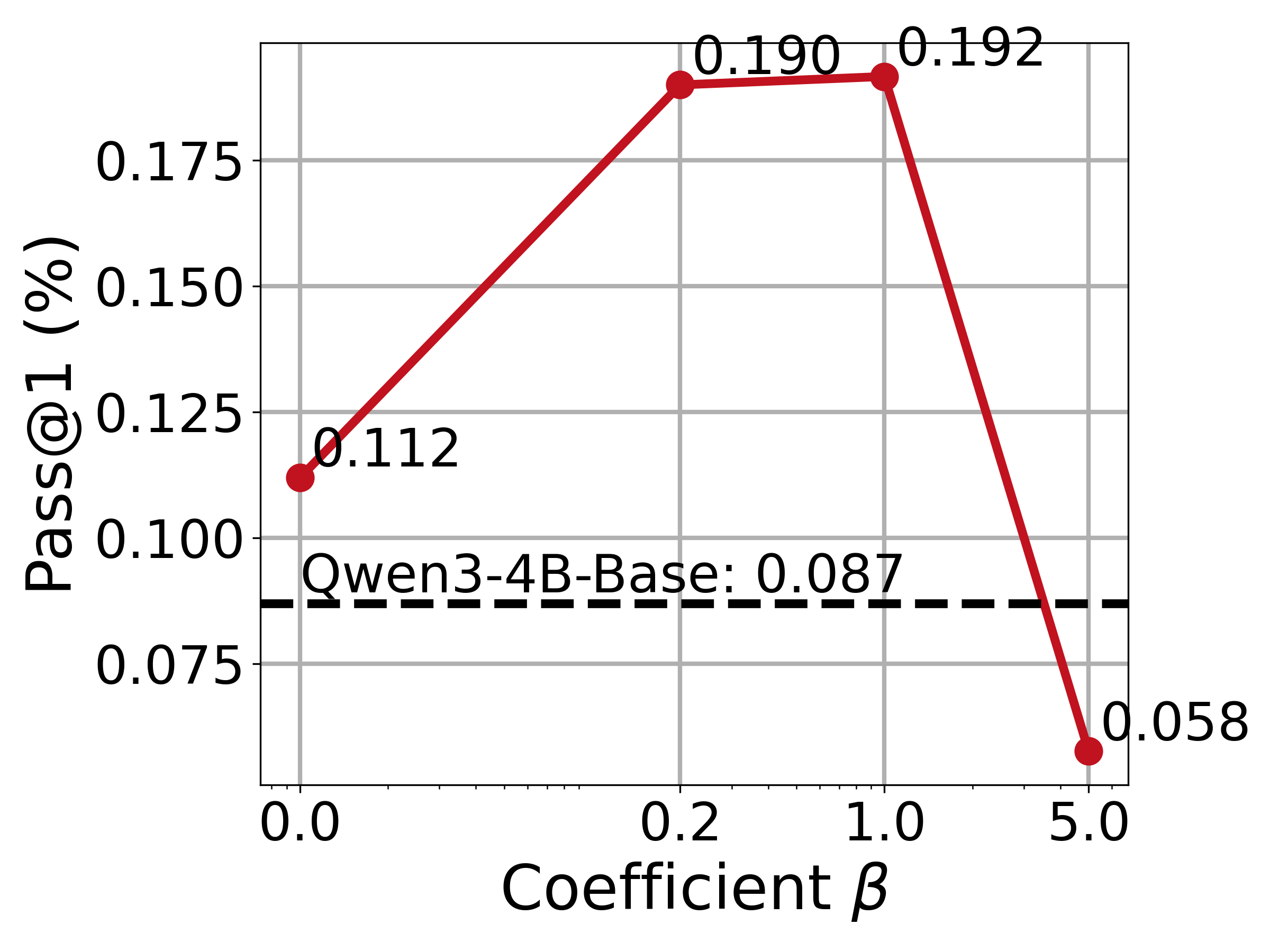}\label{fig:ab_sf_a}}
    \subfigure[]{\includegraphics[width=0.26\linewidth]{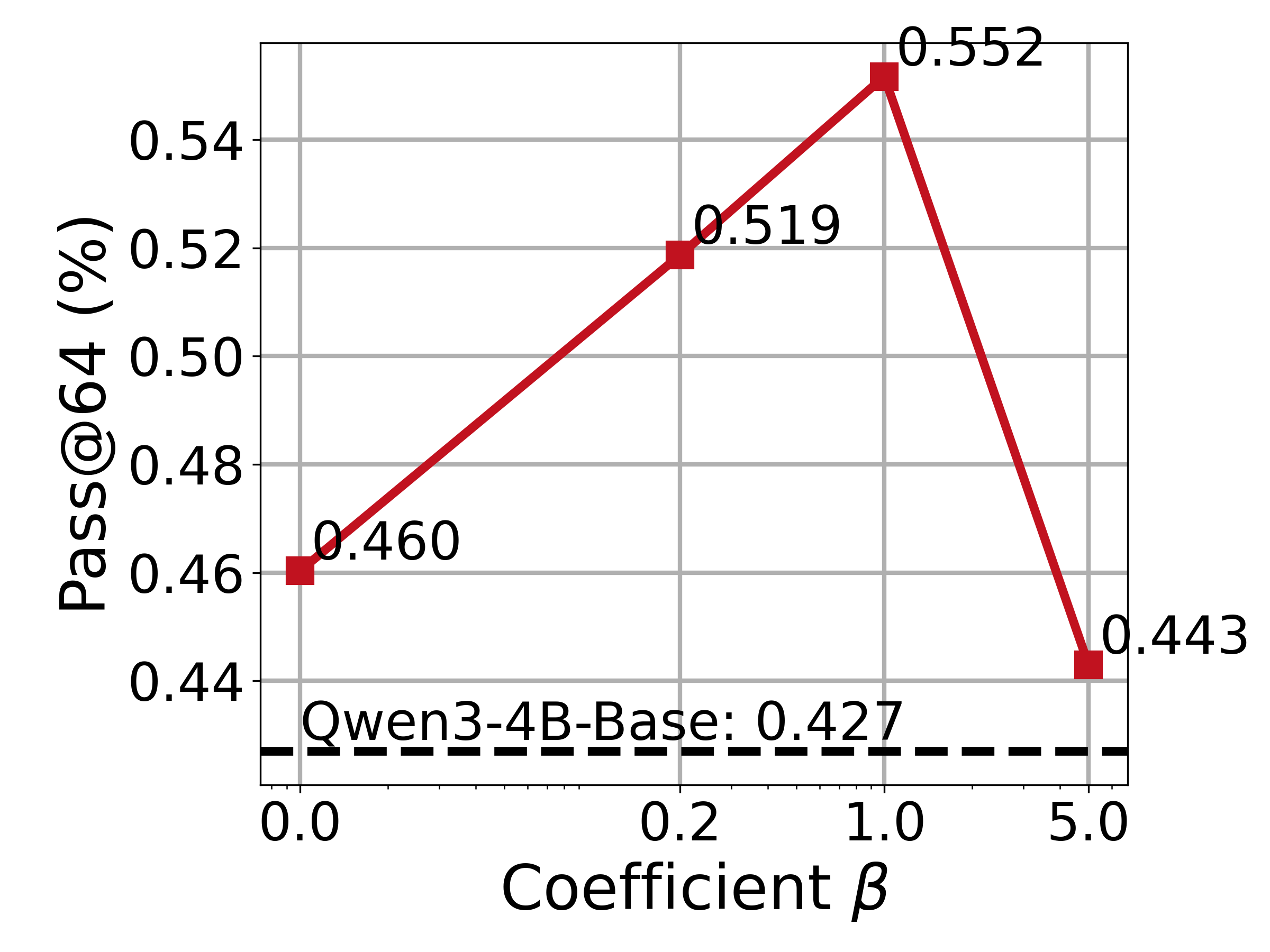}\label{fig:ab_sf_b}}
    \subfigure[]{\includegraphics[width=0.23\linewidth]{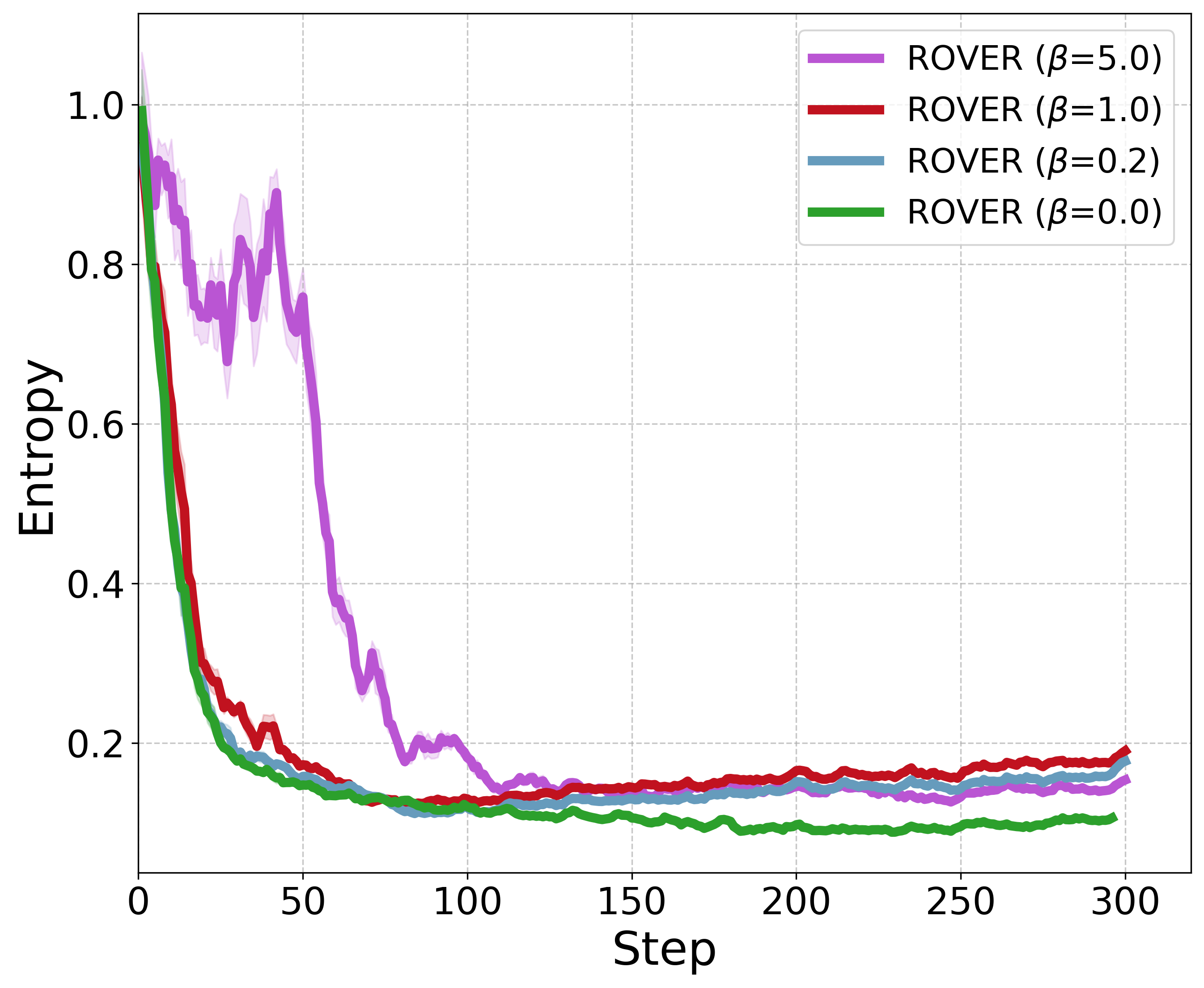}\label{fig:ab_sf_c}}
    \subfigure[]{\includegraphics[width=0.23\linewidth]{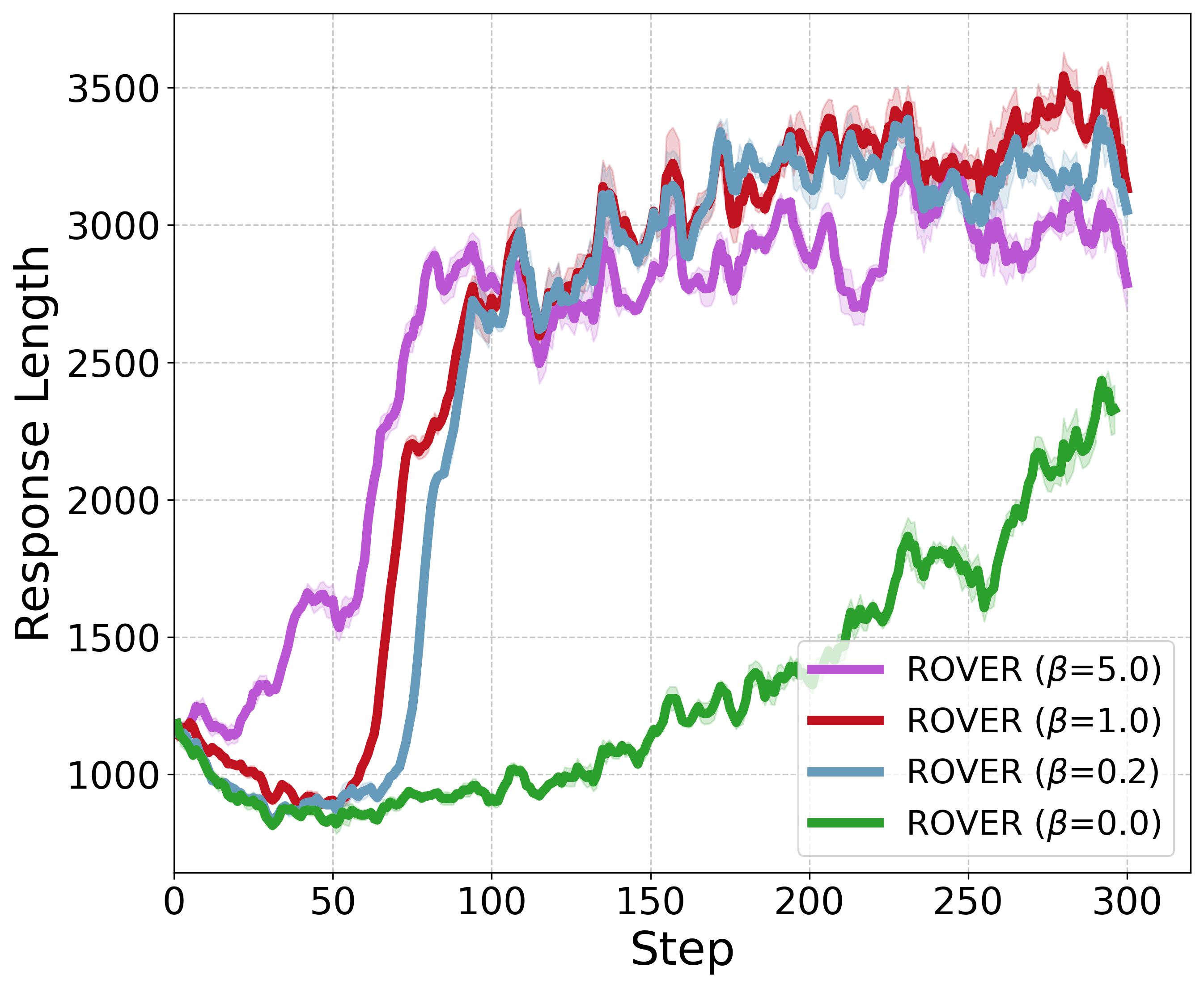}\label{fig:ab_sf_d}}
    \vspace{-.10in}
    \caption{(a)\&(b). Impact of coefficient $\beta$ in \ours on pass@1 \& pass@64, average performance on AIME24, AIME25, HMMT25 is reported. The X-axis is on a log scale. (c)\&(d). Entropy and response length curves throughout training. All experiments are conducted on Qwen3-4B-Base with LLM decoding temperature $1.0$, and trained for 300 steps.}
    \vspace{-.10in}
    \label{fig:ablation_sf}
\end{figure*}

The Bellman target used for Q-value updates is composed of two components: centered reward, $\tilde{r}$, and the expected Q-value of the successor state under a uniform policy, $Q'=\frac{1}{|\mathcal{V}|}\sum_{a_{t+1}\in\mathcal{V}}{Q(a_{t+1}|{s_{t+1})}}$ (see Appendix~\ref{app:gradient} for the impact $Q'$ in a theoretical perspective). We ablate the contribution of $Q'$ in the Bellman target by scaling it with a coefficient $\beta=[0.0, 0.2, 1.0, 5.0]$. The results show that this term is essential: removing it ($\beta=0$) causes a collapse in entropy and response length (see Fig.~\ref{fig:ab_sf_c} and \ref{fig:ab_sf_d}), leading to a sharp drop in pass@$k$ performance. Conversely, an overly dominant Q-term ($\beta=5.0$) diminishes the reward signal, which also degrades performance. Crucially, as shown in Fig.~\ref{fig:ab_sf_a} and \ref{fig:ab_sf_b}, our method is not sensitive to the precise scaling of this term, with performance remaining stable across a wide range ($\beta$ from 0.2 to 1.0). By default, we set $\beta = 1.0$ in other experiments. Detailed pass@$k$ performances under different $\beta$ values are shown in Fig.~\ref{fig:passk-ablation-flow-scale}. Fig.~\ref{fig:8b-adv-nextq} summarizes the statistics of $Q'$ throughout training.

\section{Related Work}\label{sec:related_work}
\noindent \textbf{RLVR.} RL with verifiable rewards (RLVR)~\citep{guo2025deepseek,qwq32b,yang2025qwen3,comanici2025gemini} has found great success in post-training LLMs on verifiable tasks. To bypass the need for the value
model of PPO~\citep{schulman2017proximal}, many actor-only variants have been proposed, such as GRPO~\citep{shao2024deepseekmath}, RLOO~\citep{RLOO}, ReMax~\citep{remax}, and REINFORCE++~\citep{hu2025reinforce++}. Nevertheless, leading algorithms like GRPO still exhibit unstable learning dynamics and are prone to model collapse~\citep{xu2025single}. Recent works propose to add various heuristics on advantage normalization~\citep{liu2025understanding,zheng2025group,zhao2025geometric}, clipping ratio~\citep{yu2025dapo}, KL regularization~\citep{liu2025prorl}, entropy loss~\citep{he2025skywork,zhang2025right}, reward shaping~\citep{yao2025diversity,cheng2025reasoning,chen2025seed,chen2025pass}, data augmentation~\citep{yang2025depth,liang2025beyond}, and others~\citep{cui2025entropy,wang2025beyond}. Crucially, these existing works are still constrained by the same surrogate, policy-gradient-based PPO objective, and often necessitate complex, case-specific tuning~\citep{liu2025part}. Our work departs from this paradigm, proposing a method grounded in random policy valuation that offers a minimalist yet theoretically guaranteed approach to fine-tuning LLMs.

\noindent \textbf{Diversity metrics and diversity-aware RLVR}. Diversity measurement can generally be categorized into token-level and response-level approaches. At the token level, entropy serves as the primary metric~\citep{song2025outcome}, where higher entropy applied to model sampling typically indicates greater potential diversity in model outputs. Entropy in RLVR has emerged as a prominent research topic~
\citep{cheng2025reasoning}, with the prevailing view suggesting that RLVR processes lead to entropy collapse, whereby RL-trained models achieve improved pass@1 performance at the expense of pass@k performance compared to base models~\citep{yue2025does}. Consequently, numerous recent studies have focused on entropy preservation (especially on those high-entropy forking tokens~\citep{wang2025beyond}) during RL post-training to maintain and enhance models' exploratory capabilities~\citep{liu2025uniform, wang2025stabilizing}. 
As for response-level diversity, several works have examined lexical and syntactic diversity metrics, including n-gram overlap~\citep{lanchantin2025diverse}, self-BLEU, and self-ROUGE~\citep{chen2025enhancing}. 
Nevertheless, syntactic diversity often lacks substantive significance, particularly in complex reasoning tasks where identical strategies can manifest through numerous different syntactic expressions. 
Therefore, in this work, we primarily adopt semantic-level diversity measures, encompassing both classical cosine distance metrics in embedding space and the "number of distinct strategies" and utility metrics proposed in NoveltyBench~\citep{zhang2025noveltybench}.
Several diversity-aware RLVR methods employ reward reweighting to downweight redundant responses and amplify rewards for diverse ones, in which diversity measurements are based on sequence-level probability~\citep{he2025rewarding}, submodular mutual information~\citep{chen2025dra}, or semantic classification~\citep{darling}, respectively.
Unlike prior works, \ours preserves generation diversity significantly without explicit regularization or heuristics, naturally preventing mode collapse during RL optimization.

\section{Conclusion}
We present ROVER, a minimalist approach to RLVR that achieves high-quality and diverse reasoning policies from uniformly random policy Q-values, which eliminates the need for complex evaluation-improvement loops with superior performance and diversity compared to SOTA methods.
Our experiments are limited to math reasoning tasks with models up to 8B parameters due to restricted computational resources. 

\noindent \textbf{Discussion.} ROVER provides strong foundations for simplifying RLVR in deterministic tree-structured MDPs with binary terminal rewards. While autoregressive LLM generation naturally aligns with these properties, it may not strictly hold in all extended RLVR applications (e.g., with tool calls or with intermediate feedback). 
The practical implementation of ROVER for scaling up to large action spaces and long horizons also introduces approximation. Although the empirical success suggests robustness in the underlying principles despite these approximations, an interesting future direction is to further bridge this gap.
We consider these as opportunities to reconsider RLVR from the first principles, develop more robust simplified approaches, and extend ROVER to other tasks. 
We believe that our approach establishes a valuable foundation for future research by demonstrating the power of a surprising simplification in this domain, and hope that it inspires future research to adapt and extend these insights to other structures while maintaining the core benefits of simplicity for high-quality performance and diversity preservation. 

\section*{Acknowledgements}

This work was supported by computing resources and infrastructure provided by StepFun. We are grateful to researchers from StepFun for their valuable feedback and contributions.

\bibliography{main}
\bibliographystyle{plainnat}

\clearpage
\appendix
\section{Proofs in \S~\ref{sec:theory}}
\label{app:theorem}

\subsection{Proof of Theorem~\ref{theo:theorem1}}
\label{app:theo_1}
\textbf{Theorem 1}
\emph{Consider an episodic finite-horizon episodic MDP with binary terminal rewards $\mathcal{R}(s) \in \{0, R\}$ where $R > 0$ ($R$ for a correct solution, $0$ otherwise). Let $\pi_u$ be a uniform policy, and let $Q^{\pi_u}$ denote its Q-function. Define the greedy policy with respect to $Q^{\pi_u}$ by $\pi_{\rm greedy}(s) = \arg\max_a Q^{\pi_u}(s,a)$. Then $\pi_{\rm greedy}$ is the optimal policy.}

\begin{proof}
As the underlying graph is a tree, starting from $s_0$ under policy $\pi_{\rm greedy}$ gives a unique chain $s_0 \to s_1 \to \cdots \to s_n$. By definition, for any state-action pair $(s,a)$, if the subtree below $(s,a)$ does not contain a correct terminal state, then $Q^{\pi_u}(s,a)=0$; conversely, if its subtree contains a correct terminal state, then $Q^{\pi_u}(s,a)>0$. Therefore, at $s_0$ we choose $a_0=\arg\max_a Q^{\pi_u}(s,a)$, the next state $s_1$ will necessarily lie on a path that reaches a correct terminal state. We keep proceeding until $s_{n-1}$, and $\pi_{\rm greedy}(a|s_{n-1})=\arg\max_a Q^{\pi_u}(s_{n-1},a)$ also selects the optimal action $a$ (as $Q^{\pi_u}(s_{n-1},a_{n-1})=R(s_{n-1}, a_{n-1})=R$).

\end{proof}

\subsection{Proof of Theorem~\ref{theo:pi_bound}}
\label{app:theo_2}
\textbf{Theorem 2}
\textit{Consider the same MDP $\mathcal{M}$, and let $Q^{\pi_u}(s,a)$ denote the Q-function under the uniform random policy $\pi_u$ from state-action pair $(s,a)$, $N(s)=|\{a: Q^{\pi_u}(s,a)=0\}|$ be the number of zero-valued actions at state $s$, $A(s)$ be the number of available actions at state $s$, and $P$ denotes the set of key states where both optimal and suboptimal actions exist, i.e., $P=\{s: 1\leq N(s)\leq A(s)-1\}$.
Given the softmax policy $\pi_s(a|s)=\frac{\exp( Q^{\pi_u}(s,a)/\rho)}{\sum_{a'}\exp(Q^{\pi_u}(s,a')/\rho)}$ with temperature $\rho>0$, and $Pr^{\pi_s}(s|s_0)$ is the probability of reaching $s$ from $s_0$ with the policy $\pi_s$,  the value function of the induced policy $\pi_s$ satisfies the following lower bound: $V^{\pi_s} (s_0)\geq R\left( 1-\sum_{s \in P} Pr^{\pi_s}(s|s_0) \frac{N(s)}{N(s) + \exp(\max_a Q^{\pi_u}(s,a)/\rho)} \right)$.} 
\begin{proof}
Let us sample trajectories from the initial state $s_0$ using policy $\pi_s$. 
For any incorrect trajectory $\tau$ that achieves a reward value of $0$ (one that fails to reach a correct terminal state with positive reward $R$), there must exist at least one key state along $\tau$. For each $\tau$, let $s_{\tau}$ denote the last key state along $\tau$.

The probability of trajectory $\tau$ can be factored as:
\begin{equation}
 Pr(\tau)=Pr^{\pi_s}(s_{\tau}|s_0)\prod_{t \geq t_{\tau}}\pi_s(a_t|s_t),
\end{equation}
where $t_{\tau}$ denotes the index of state $s_\tau$ in the trajectory sequence.

Let $\mathcal{T}_w$ denote the set of all incorrect trajectories, then we have that 
\begin{equation}
Pr(\mathcal{T}_w)=\sum_{\tau \in \mathcal{T}_w}Pr(\tau).
\end{equation}

For any key state $s \in P$, 
let $\mathcal{T}(s)$ denote the set of incorrect trajectories for which $s$ is the last key state. Since the underlying MDP $\mathcal{M}$ has a tree structure, the sets $\{\mathcal{T}(s)\}_{s \in P}$ form a partition of $\mathcal{T}_w$. Therefore, we have that
\begin{equation}
Pr(\mathcal{T}_w)=\sum_{s\in P}Pr^{\pi_s}(s|s_0)
\sum_{\tau 
\in \mathcal{T}(s)}\prod_{t \geq t_s}\pi_s(a_t|s_t).
\label{eq:1}
\end{equation}

As the state $s$ is the last key state on any trajectory in $\mathcal{T}(s)$, we have that
\begin{equation}
\sum_{\tau 
\in \mathcal{T}(s)}\prod_{t \geq t_s}\pi_s(a_t|s_t)=Pr(Q^{\pi_u}(s,a)=0|s,\pi_{s}),
\label{eq:2}
\end{equation}
where $Pr(Q^{\pi_u}(s,a)=0|s,\pi_{s})$ is the probability that policy $\pi_s$ selects an action with zero Q-value at state $s$.

By the definition of the softmax policy, we have that 
\begin{align}
Pr(Q^{\pi_u}(s,a)=0|s,\pi_{s})&=\sum_{a: Q^{\pi_u}(s,a)=0}\pi_s(a|s) \\
&= \sum_{a:Q^{\pi_u}(s,a)=0}\frac{\exp( Q^{\pi_u}(s,a)/\rho)}{\sum_{a'}\exp(Q^{\pi_u}(s,a')/\rho)} \\
&= \frac{N(s)}{N(s)+\sum_{a':Q^{\pi_u}(s,a')>0}\exp(Q^{\pi_u}(s,a')/\rho)} \\
& \leq \frac{N(s)}{N(s)+\exp(\max_{a'} Q^{\pi_u}(s,a')/\rho)}
\label{eq:3}
\end{align}

Combing Eq.~(\ref{eq:1}), Eq.~(\ref{eq:2}), and Eq.~(\ref{eq:3}), we have that
\begin{equation}
Pr(\mathcal{T}_w)\leq \sum_{s\in P}Pr^{\pi_s}(s|s_0)\frac{N(s)}{N(s)+\exp(\max_{a'} Q^{\pi_u}(s,a')/\rho)}.
\end{equation}

By definition, the value function of $\pi_{s}$ is related to the probability of correct trajectories: 
\begin{equation}
V^{\pi_s}(s_0)=(1-Pr(\mathcal{T}_w))R. 
\end{equation}

Substituting our upper bound on $Pr(\mathcal{T}_w)$, we have that 
\begin{equation}
V^{\pi_s}(s_0)\geq R \left(1-\sum_{s\in P}Pr^{\pi_s}(s|s_0)\frac{N(s)}{N(s)+\exp(\max_{a'} Q^{\pi_u}(s,a')/\rho)}\right).
\label{eq:4}
\end{equation}

For any key state $s\in P$, we have that $\max_{a'} Q^{\pi_u}(s,a')>0$ by definition. As $\rho \rightarrow 0$, the right-hand side in Eq.~(\ref{eq:4}) converges to $R$, which is the optimal value. 
\end{proof}

\section{Gradient Analysis}
\label{app:gradient}
In this section, we analyze the relationship between our method and existing policy optimization methods from the gradient perspective. 

\textbf{Proposition 1} \textit{Assume only \(\log \pi_\theta\) has parameters (i.e., LLM policy \(\pi\) depends on \(\theta\)). Define importance sampling ratio \({\rm IS} = \frac{\pi_\theta(a|s)}{\pi_{\theta_{\rm old}}(a|s)}\), where \(\pi_{\theta_{\rm old}}\) is the behavior policy. Denote $\tilde{r}$ as our mean-centered reward. Then the gradient of ROVER's objective takes the following form, which is similar to policy-gradient: }
\begin{equation}
    \nabla_\theta \mathcal{L}_{\text{ROVER}}=\mathbb{E}_{s,a,s'\sim P}\left[\left((\tilde{r}+Q')-\log {\rm IS} \right) \nabla_\theta \log \pi_\theta(a|s)\right],\ {\rm where}\ Q'=\frac{1}{|\mathcal{V}|}\sum_{a'\in\mathcal{V}}{Q(a'|{s')}}
\end{equation}
\begin{proof}
    Recall the details of our method provided in \S~\ref{sec:implementation}, ROVER has the following loss function:
    \begin{equation}
        \mathcal{L}_{\text{ROVER}} = \mathbb{E}_{s,a,s'\sim P}\left[\big(\tilde{r}+\frac{1}{|\mathcal{V}|}\sum_{a'\in\mathcal{V}}{Q(a'|{s')}}-Q(s,a)\big)^2\right].
    \end{equation}
    Let 
    \begin{equation}
    \label{eq:gradient_u}
    \begin{aligned}
        u &= \tilde{r}+\frac{1}{|\mathcal{V}|}\sum_{a'\in\mathcal{V}}{Q(a'|{s')}}-Q(s,a)\\
        &=\tilde{r}+\frac{1}{|\mathcal{V}|}\sum_{a'\in\mathcal{V}}{Q(a'|{s')}}-(\log \pi_\theta - \log \pi_{\theta_{\rm old}})\\
        &= \tilde{r}+\frac{1}{|\mathcal{V}|}\sum_{a'\in\mathcal{V}}{Q(a'|{s')}}-\log \left( \frac{\pi_\theta}{\pi_{\theta_{\rm old}}} \right) \\
        &= \tilde{r}+\frac{1}{|\mathcal{V}|}\sum_{a'\in\mathcal{V}}{Q(a'|{s')}}-\log {\rm IS}. 
    \end{aligned}
    \end{equation}
    Then the gradient is 
    \begin{equation}
    \label{eq:gradient_mse}
        \nabla_\theta \mathcal{L}_{\text{ROVER}} = \mathbb{E}_{s,a,s'\sim P}[u \cdot \nabla_\theta u]
    \end{equation}
    Given that $Q'=\frac{1}{|\mathcal{V}|}\sum_{a'\in\mathcal{V}}{Q(a'|{s')}}$, where the gradient of $Q'$ is stopped (see Alg.~\ref{alg:ours}), and $\pi_{\theta_{\rm old}}$ does not involve gradient backpropagation, by combining Eq.~\ref{eq:gradient_u} and Eq.~\ref{eq:gradient_mse} we have:
    \begin{equation}
    \begin{aligned}
    \nabla_\theta \mathcal{L}_{\text{ROVER}}&=\mathbb{E}_{s,a,s'\sim P}\left[\big(\tilde{r}+\frac{1}{|\mathcal{V}|}\sum_{a'\in\mathcal{V}}{Q(a'|{s')}}-\log {\rm IS}\big)\nabla_\theta \log {\rm IS}\right]\\
    &=\mathbb{E}_{s,a,s'\sim P}\left[\big(\tilde{r}+\frac{1}{|\mathcal{V}|}\sum_{a'\in\mathcal{V}}{Q(a'|{s')}}-\log {\rm IS}\big)\nabla_\theta \log \pi_\theta(a|s)\right].
    \end{aligned}
    \end{equation}
\end{proof}
Note the gradient of a typical policy optimization method, i.e., GRPO~\citep{shao2024deepseekmath}, is
\begin{equation}
    \nabla_\theta \mathcal{L}_{\text{GRPO}} = \mathbb{E}_{s,a} [A \cdot {\rm IS} \cdot\nabla_\theta {\rm IS}]=\mathbb{E}_{s,a} [A \cdot {\rm IS} \cdot \nabla_\theta \log\pi_\theta(a|s)].
\end{equation}
Therefore, we have the following key observation:
\begin{itemize}[leftmargin=*]
  \item Both gradients share the term \(\nabla_\theta \log \pi_\theta\) (core of policy gradient).
  \item When importance sampling ratio \({\rm IS} \to 1\) (small policy update), i.e., \(\log {\rm IS} \to 0\), so:
        \[
        \nabla_\theta \mathcal{L}_{\text{ROVER}} \approx \mathbb{E}\big[(\tilde{r}+Q') \nabla_\theta \log \pi_\theta\big], \quad \nabla_\theta \mathcal{L}_{\text{GRPO}} = \mathbb{E}\big[A\cdot \nabla_\theta \log \pi_\theta\big].
        \]
\end{itemize}

\begin{figure*}[htbp]
    \centering
    \subfigure[AIME 2024]{\includegraphics[width=0.32\linewidth]{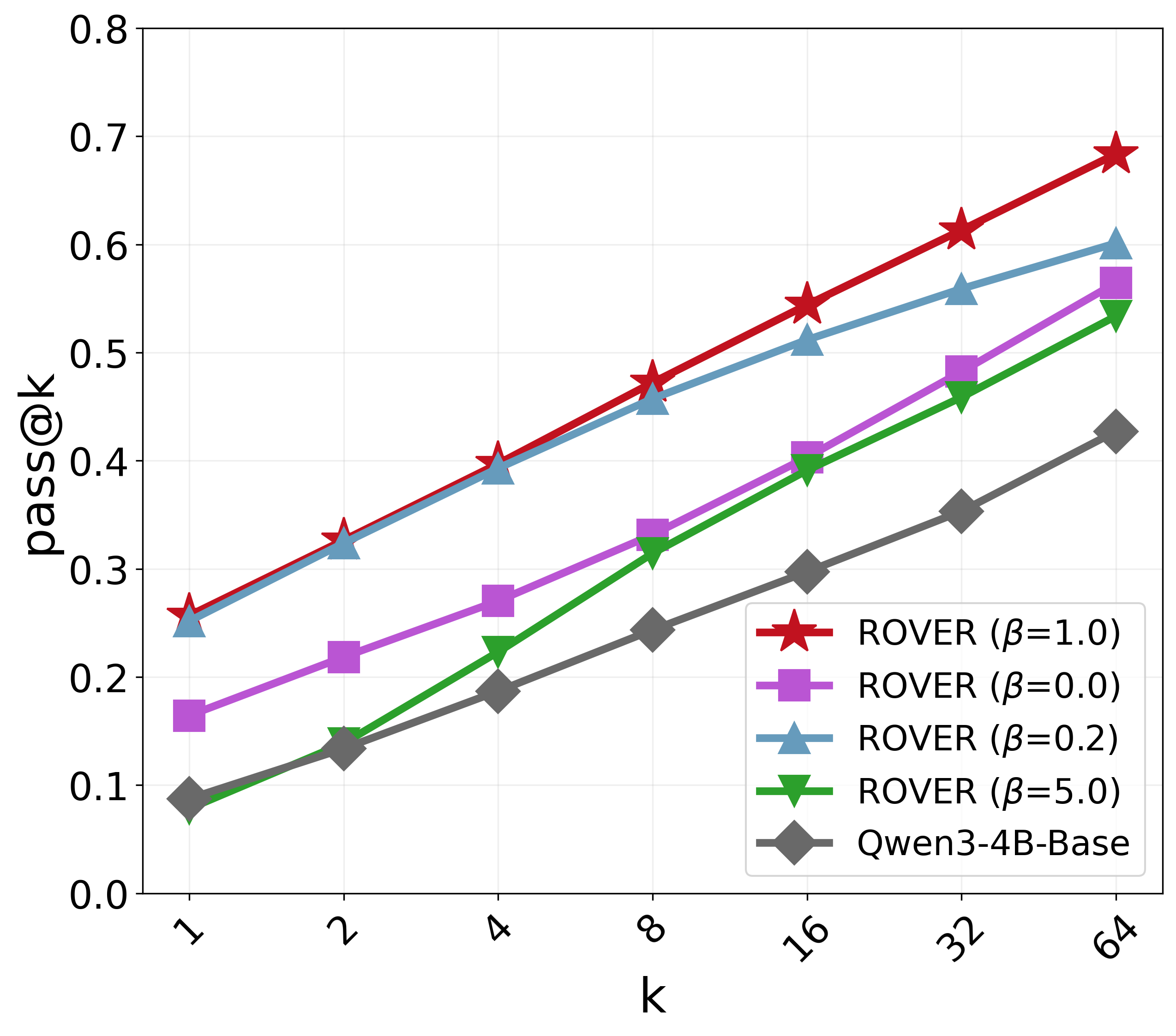}}
    \subfigure[AIME 2025]{\includegraphics[width=0.32\linewidth]{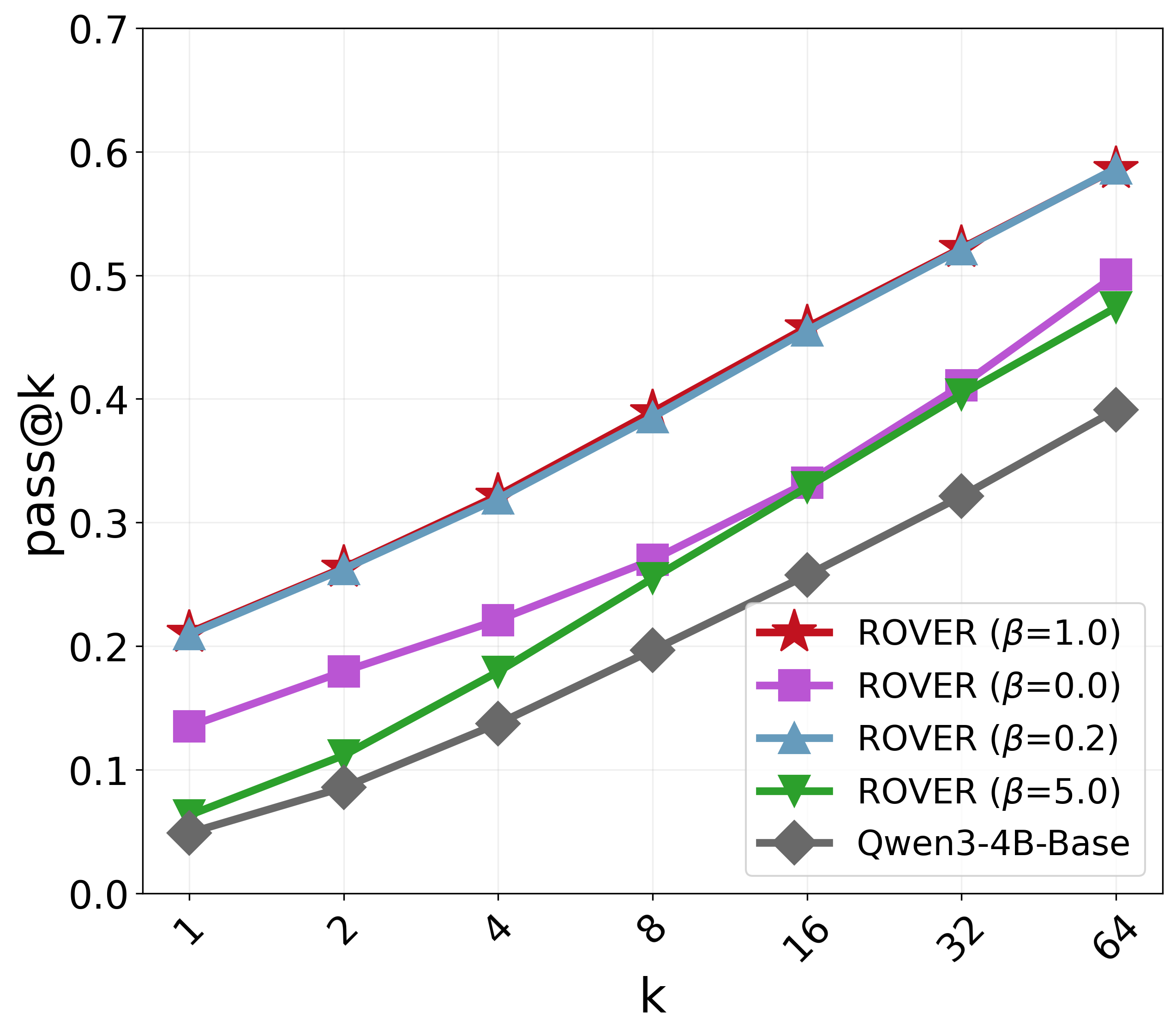}}
    \subfigure[HMMT 2025]{\includegraphics[width=0.32\linewidth]{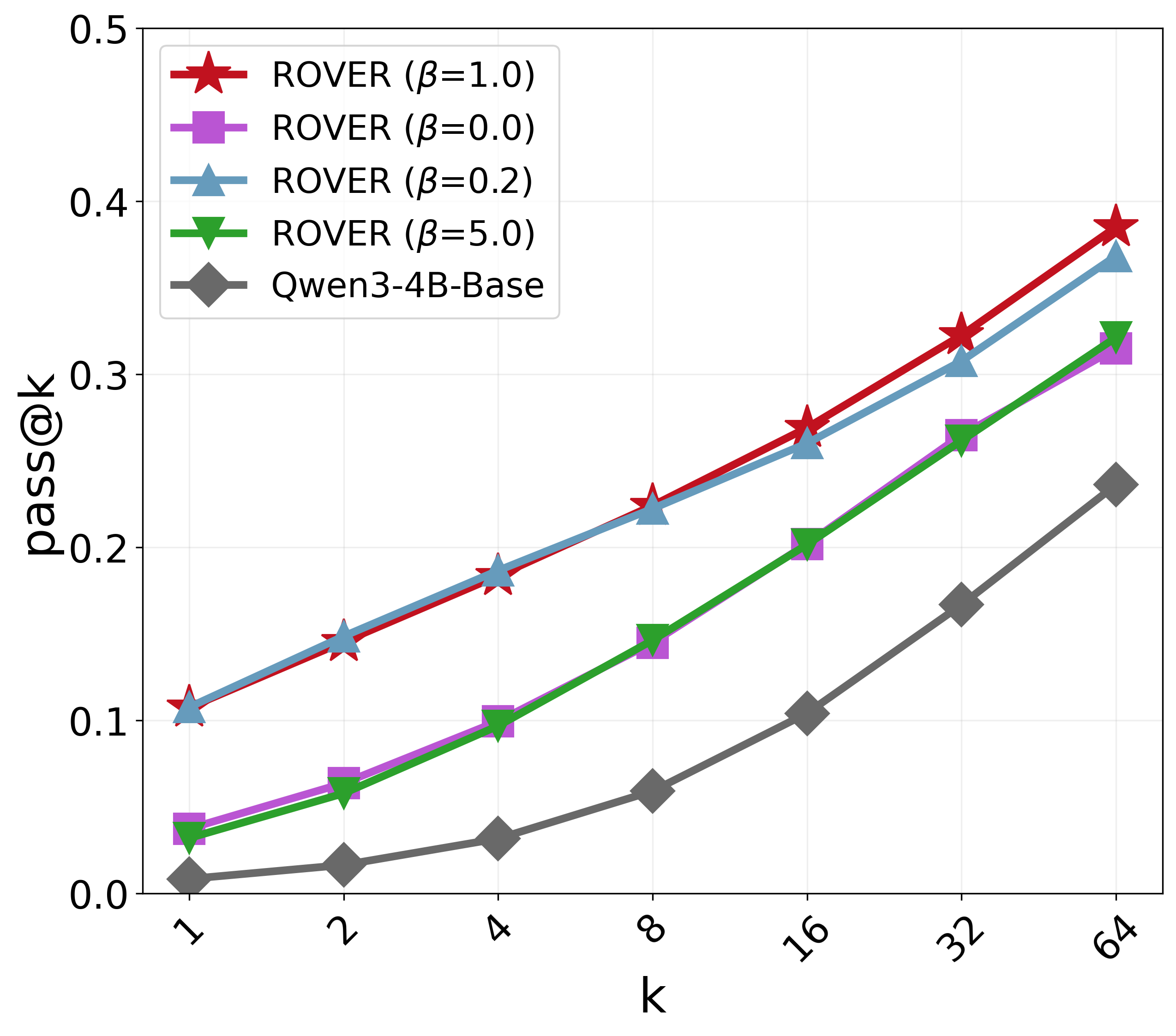}}
    \vspace{-.15in}
    \caption{pass@$k$ performances under different value of coefficient $\beta$ in \ours. All experiments are conducted on Qwen3-4B-Base and trained for 300 steps.}
    \vspace{-.10in}
    \label{fig:passk-ablation-flow-scale}
\end{figure*}

These two objectives can be approximately equal if we remove the term $Q'$ in \ours and the advantage $A$ in GRPO is normalized without the standard deviation term. See \S~\ref{sec:ablation} for a empirical justification where we analyze the importance of the term $Q'$ in \ours. Supplemental pass@$k$ results are provided in Fig.~\ref{fig:passk-ablation-flow-scale}.

\section{The Countdown Task}
\label{app:countdown}
\paragraph{Task Details.}
Countdown~\citep{gandhi2025cognitive} is a math reasoning task capable of evaluating the arithmetic capabilities of LLMs. Below illustrates the toy example of the countdown task:
\begin{equation*}
\begin{aligned}
    &\text{nums:}\ \ \ [19,36,55,7],\\
    &\text{target:}\ \ \ \ \ \ \ 65, \\
    &\text{answer:}\ \ \  55+36-7-19,
\end{aligned}
\end{equation*}
where the LLM should find the correct solution using the given numbers and basic arithmetic operations ($+,-,\times,\div$). The simplicity of the Countdown’s reasoning path, yet challenging for small LLMs to solve
effectively, makes it an accessible test bed for math reasoning.
\paragraph{Training Details.} We use the training and testing dataset provided by TinyZero~\citep{tinyzero}. The training dataset contains 327680 problems, and the testing dataset contains 1024 unseen problems. A reward of 1 is given if the LLM finds the correct equation; Otherwise, it receives a zero reward. We set the batch size to 128 and the mini batch size to 64 during training. Optimization is conducted by an AdamW~\citep{loshchilov2017decoupled} where learning rate is $1 \times 10^{-6}$. The response length is set to 1k for both training and evaluation. We rollout 5 responses per prompt to calculate the mean-centered reward. Other configurations follow the default setting of TinyZero~\citep{tinyzero}. For the baseline of GRPO with Clip\_higher technique, we set clip ratio $\epsilon_{\rm low}=0.2$ and $\epsilon_{\rm high}=0.4$.
Note that all the experimental settings across different methods remain the same for a fair comparison.

\section{Results of Math Tasks on DeepSeek-1.5B}
\label{app:deepseek_1.5b}
We provide the details of the training setup on {DeepSeek-R1-Distill-Qwen-1.5B} model~\citep{guo2025deepseek} as follows.
\begin{itemize}[leftmargin=*]
    \item We employ the datasets provided by DeepScaler~\citep{deepscaler2025}, which contains 40k verifiable math questions.
    \item Built upon the veRL infra~\citep{sheng2024hybridflow}, we set batch size to 128 and mini-batch size to 64. 
    \item we use the AdamW~\citep{loshchilov2017decoupled} optimizer with a constant learning rate of $1\times 10^{-6}$ for gradient backpropagation.
    \item We rollout 8 responses per prompt to calculate the mean-centered reward $\tilde{r}$.
    \item Following DeepScaler~\citep{deepscaler2025}, we first train {DeepSeek-R1-Distill-Qwen-1.5B} for 1k steps with a 8k response length. Then we scale the response to 16k for an additional 1k training steps. Experiments are conducted on 8 H200 GPUs for around 5 days.
\end{itemize} 
Following the evaluation scripts provided by veRL, we use a sampling temperature of 0.6, nucleus sampling~\citep{holtzman2019curious} with top\_p = 0.95, and a maximum response length of 24k for evaluation. We evaluate our ROVER-trained model and previous SOTA models such as {DeepScaler-1.5B}~\citep{deepscaler2025} and {ProRLv2-1.5B}~\citep{liu2025prorl} on AIME24, AIME25, AMC23, and MATH tasks. We rollout 128 responses per prompt for each task, and report both the pass@1 (avg@128) and pass@64 (calculated by an unbiased estimator~\citep{chen2021evaluating}) for comprehensive comparison. 
\begin{table}[tb]
\centering
\caption{Results of {DeepSeek-R1-Distill-Qwen-1.5B} on typical math competition tasks. The highest and the second-best scores are shown in bold and underlined, respectively.}
\resizebox{\linewidth}{!}{
\begin{tabular}{@{}c|cccc|cccc@{}}
\toprule
\multirow{2}{*}{Models}                                                     & \multicolumn{4}{c|}{Pass@1}& \multicolumn{4}{c}{Pass@64}\\
& AIME24& AIME25 & AMC23 & MATH  & AIME24  & AIME25& AMC23& MATH          \\ \midrule
\gray{DeepSeek-R1-Distill-Qwen-1.5B~\citep{guo2025deepseek}} & \gray{29.3}                          & \gray{24.3}                          & \gray{62.5}                          & \gray{82.9}                          & \gray{79.8}          & \gray{58.3}          & \gray{92.9}          & \gray{97.3}          \\
DeepScaleR-1.5B~\citep{deepscaler2025}                & 41.6                          & 30.8                          & 73.4                          & 87.7                          & 78.5          & 62.9          & 95.0          & 96.8          \\
ProRLv2-Qwen-1.5B~\citep{liu2025prorl}                & 52.6                          & 35.2                          & 81.5                          & 90.6                          & 79.2          & 59.7          & 94.3          & 96.1          \\
\textbf{ROVER (Ours)}                                                                 & \underline{42.2} & \underline{31.2} & \underline{74.3} & \underline{88.3} & \textbf{80.6} & \textbf{64.4} & \textbf{95.2} & \textbf{97.1} \\ \bottomrule
\end{tabular}
}
\label{tab:deepsekk-1.5b}
\end{table}

From the results summarized in Table~\ref{tab:deepsekk-1.5b}, we observe that ROVER achieves the best performance in terms of both pass@1 and pass@64 scores compared with DeepScaler, which is trained on the same dataset as ours. Note that the comparison with ProRLv2 is not fair since ROVER uses more than 3$\times$ smaller datasets (40k (ours) vs. 136k (ProRLv2)). Moreover, the training of ROVER only lasts for around 960 GPU hours, while ProRLv2 is trained for 16k GPU hours. However, thanks to the better reasoning diversity brought by our method, ROVER can achieve higher scores than ProRLv2 on pass@64.

\section{Results of Math Tasks on Qwen Models}
\label{app:math}
\subsection{Training and Evaluation Details}\label{app:train-eval-details}

\textbf{Training Details.} To ensure a fair comparison, both \ours and baselines are trained using the same learning rate, batch size, and training steps (see Table~\ref{tab:hyper-train}).
We fix 600 training steps for \ours and baselines. 
During each training step, $128 \times 8$ samples are involved to calculate gradients. The computational requirements are approximately 1,280 GPU hours for experiments initialized with {Qwen3-8B-Base} and 832 GPU hours for those with {Qwen3-4B-Base}. 

\textbf{Evaluation Details.} Default hyperparameters of evaluation are summarized in Table~\ref{tab:hyper-eval}. To compute average pass@1, we sample 256 independent runs for AIME24, AIME25, HMMT25, and AMC23 for comprehensive evaluation to reduce the variance introduced by the relatively small sizes of these benchmarks, while 10 runs are sufficient for the larger OlympiadBench, MATH500, and GPQA-diamond benchmark.

\begin{table}[htbp]
\centering
\begin{minipage}[c]{0.49\textwidth}
\centering
\caption{Default hyperparameters for RL training.}
\begin{tabular}{@{}ll@{}}
\toprule
Hyper-parameter  & Value           \\ \midrule
Temperature      & $0.6$           \\
Response length  & $8 \times 1024$ \\
Responses per prompt & $8$         \\
Train batch size & $128$           \\
Mini batch size  & $32$            \\
PPO\_epoch       & $1$             \\
Learning rate    & $1e-6$          \\ \bottomrule
\end{tabular}
\label{tab:hyper-train}
\end{minipage}
\hfill
\begin{minipage}[c]{0.49\textwidth}
\centering
\caption{Default hyperparameters for evaluation.}
\begin{tabular}{@{}ll@{}}
\toprule
Hyper-parameter & Value            \\ \midrule
Temperature     & $0.6$              \\
Response length & $24 \times 1024$ \\
top\_p          & $0.95$             \\ \bottomrule
\end{tabular}
\label{tab:hyper-eval}
\end{minipage}
\end{table}

\textbf{Baselines.} We compare \ours with the following baselines:
\begin{itemize}
    \item PPO~\citep{schulman2017proximal}: It uses a value network to estimate state values and compute advantages via GAE~\citep{schulman2015high}.
    \item GRPO~\citep{shao2024deepseekmath}: As a value-free method, it estimates advantages using group reward normalization, serving as a fundamental baseline for RLVR.
    \item DAPO~\citep{yu2025dapo}: It extends GRPO by introducing several techniques to enhance LLM training efficiency. These include clip-higher, dynamic sampling, and overlong reward shaping. We set $\epsilon_{low} = 0.2$, $\epsilon_{high} = 0.28$.
    \item REINFORCE++~\citep{hu2025reinforce++}: Different from GRPO, it incorporates global advantage normalization (across responses correspond to different prompts within a batch), resulting in an unbiased approach that significantly improves training stability. We implement the REINFORCE++-baseline version in this paper.
\end{itemize}

All baselines are rigorously implemented following the official veRL recipes~\citep{sheng2024hybridflow}.
\subsection{Case Studies}
\label{app:case_study}

\begin{figure}[htb]
    \centering
    \includegraphics[width=1\linewidth]{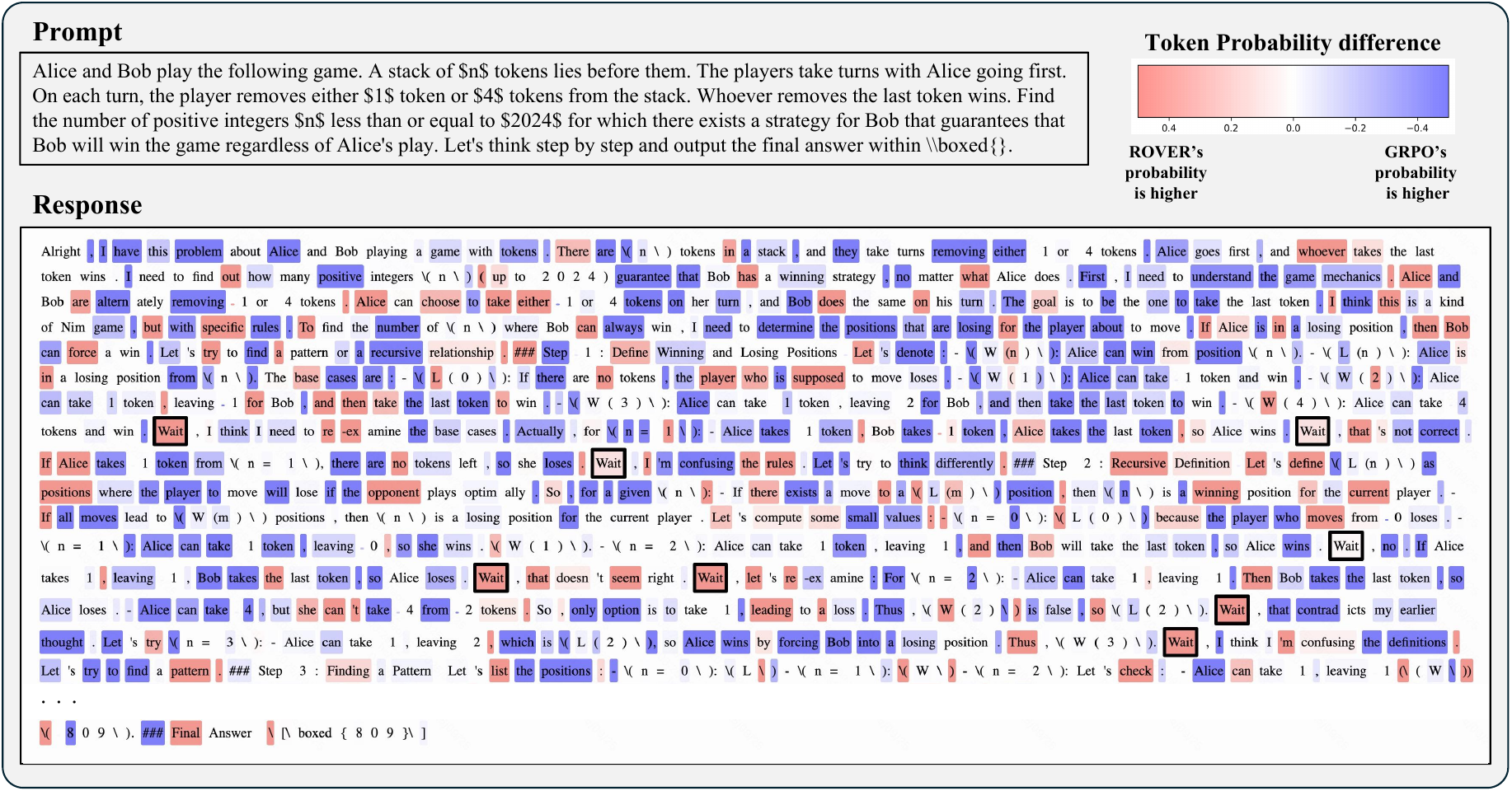}
    \vspace{-1em}
    \caption{Token probability differences between \ours and GRPO (visualized by the heatmap). \ours exhibits a significantly higher probabilities to tokens associated with \textit{reasoning contrasts or shifts}, exemplified by ``$\boxed{\rm Wait}$".}
    \vspace{-.10in}
    \label{fig:case_study_token_prob_diff}
    \vspace{-1em}
\end{figure}
We visualize a case study to show the token probability differences between \ours and GRPO in Fig.~\ref{fig:case_study_token_prob_diff} (a representative prompt from AIME24 is selected). \ours demonstrates higher probabilities for tokens indicating \textit{contrasts or shifts}, particularly ``wait", which facilitates the exploration of alternative reasoning paths, thereby contributing to increased strategic diversity. The specific \textit{forking tokens} mentioned in \S~\ref{sec:exp} are shown in Table~\ref{tab:forking_tokens}.

\begin{table}[htbp]
\centering
\caption{\textit{Forking token} categories and their corresponding tokens.}
\begin{tabular}{@{}ll@{}}
\toprule
Category & Tokens \\ \midrule
mathematical\_setup & suppose, assume, given, define \\
contrasts\_shifts & wait, however, unless \\
progression\_addition & thus, also \\ \bottomrule
\end{tabular}

\vspace{-.15in}
\label{tab:forking_tokens}
\end{table}

\subsection{Additional Experiment Results}\label{app:detailed-exp}

\begin{figure*}[htbp]
\begin{minipage}[t]{0.76\textwidth}
\vspace{6pt}
    \centering
    \subfigure[AIME 2024]{\includegraphics[width=0.32\linewidth]{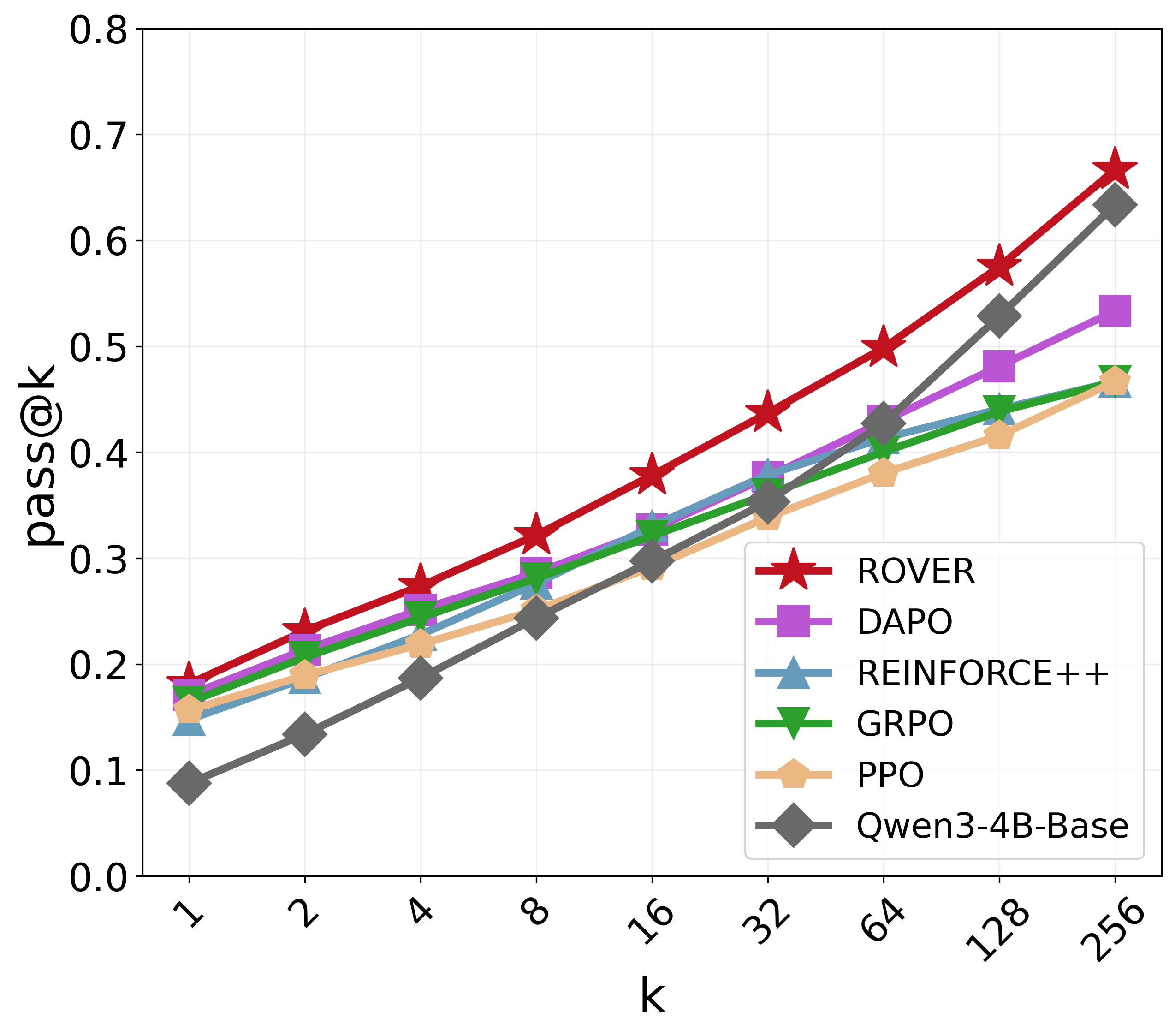}}
    \subfigure[AIME 2025]{\includegraphics[width=0.32\linewidth]{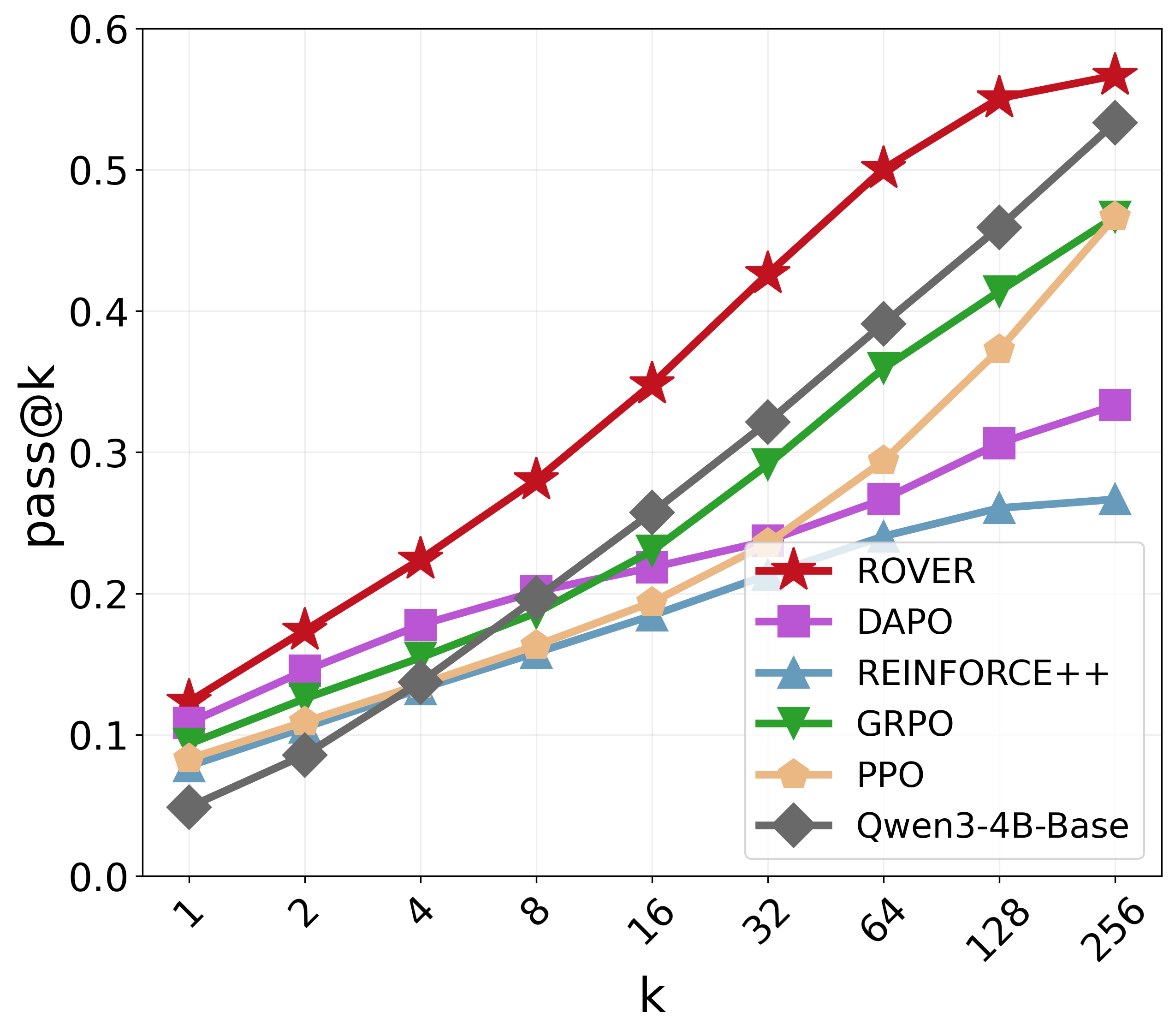}}
    \subfigure[HMMT 2025]{\includegraphics[width=0.32\linewidth]{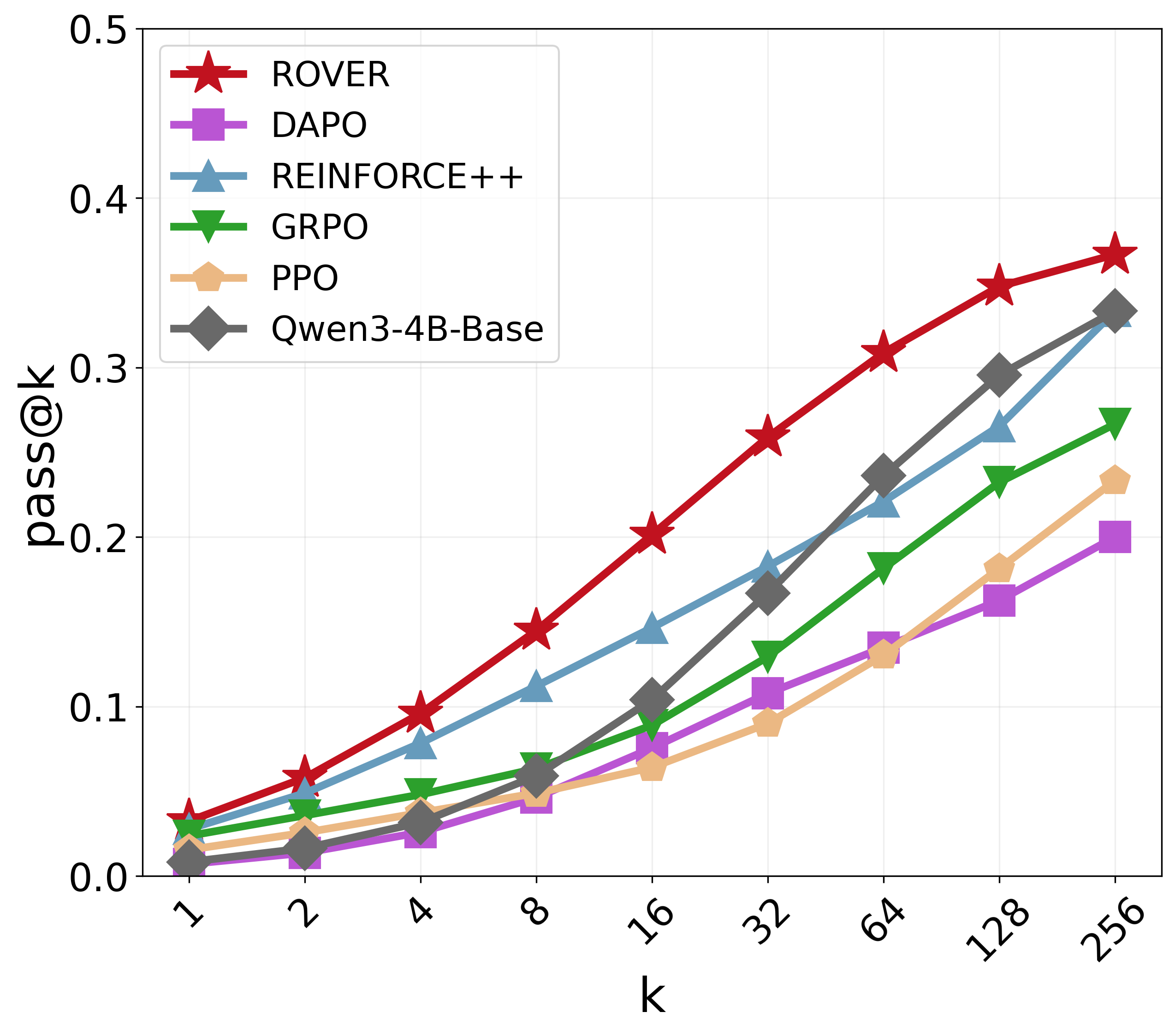}}
    \caption{pass@$k$ performances of \ours and baselines ({Qwen3-4B-Base}).}
    \label{fig:passk-4b}
\end{minipage}
\begin{minipage}[t]{0.23\textwidth}
\vspace{6pt}
    \centering
    {\includegraphics[width=1\textwidth]{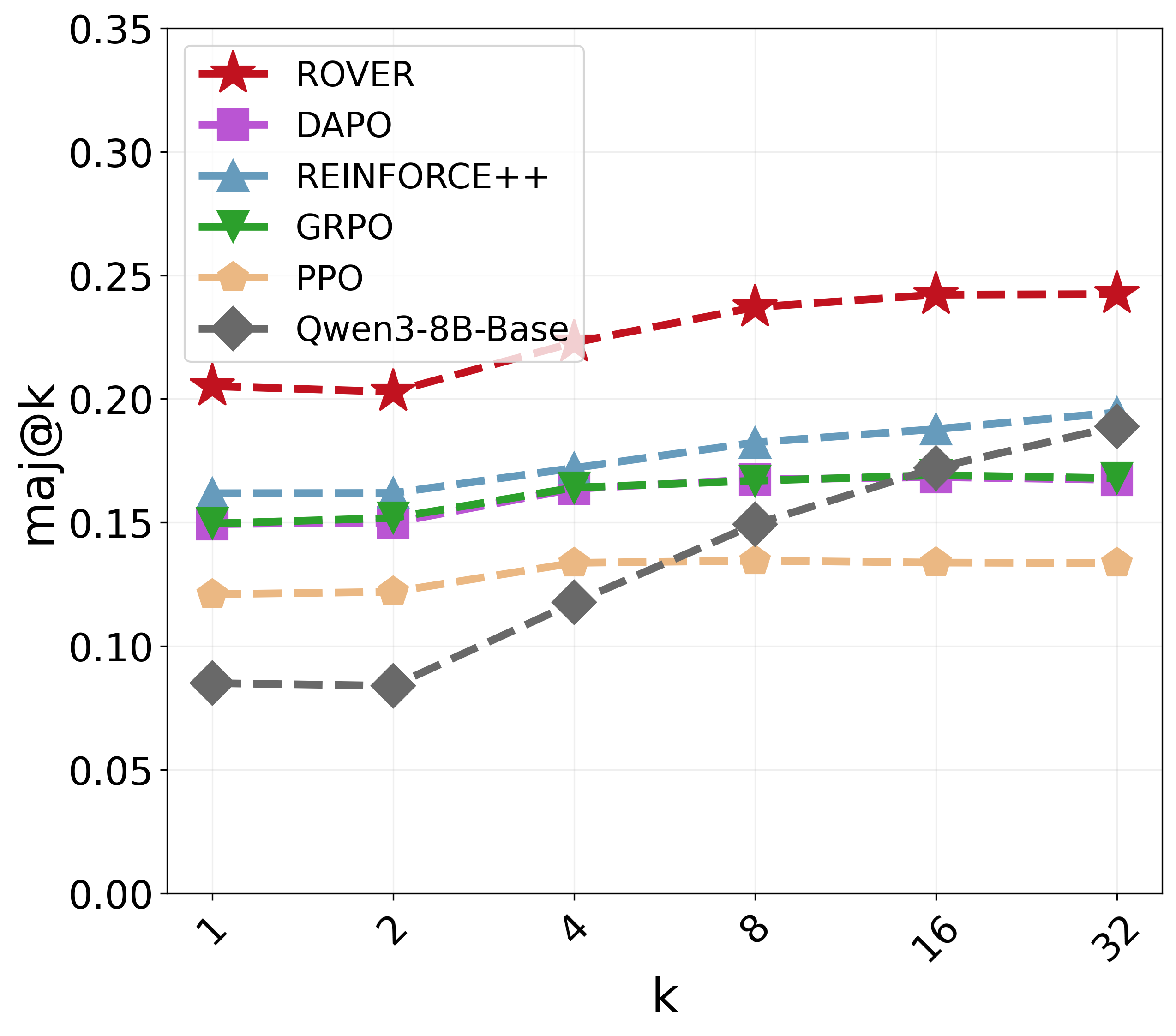}}
    \caption{Maj@$k$ performance of \ours and baselines on AIME25 for {Qwen3-8B-Base}.}
    \label{fig:majk-8b_app}
\end{minipage}
\end{figure*}

\textbf{Pass@$k$ results on {Qwen3-4B-Base}}. As shown in Fig.~\ref{fig:passk-4b}, similar to results on {Qwen3-8B-Base}, \ours demonstrates consistently superior pass@$k$ performance on {Qwen3-4B-Base} across all $k$ values, while other RL baselines drop when $k$ becomes higher.

\textbf{Maj@$k$ results.} The supplemental results of maj@$k$ performance on AIME25 for {Qwen3-8B-Base} is shown in Fig.~\ref{fig:majk-8b_app}. To mitigate random variations in evaluation results, we adopt a repeated sampling approach for computing maj@$k$: $k$ responses are randomly sampled from the response collection, and this sampling procedure is repeated 1000 times with the average value reported.

\subsubsection{Ablation of temperature $\rho$} 
\label{app:abaltion_rho_math}
\begin{figure*}[htbp]
    \centering
    \subfigure[]{\includegraphics[width=0.32\linewidth]{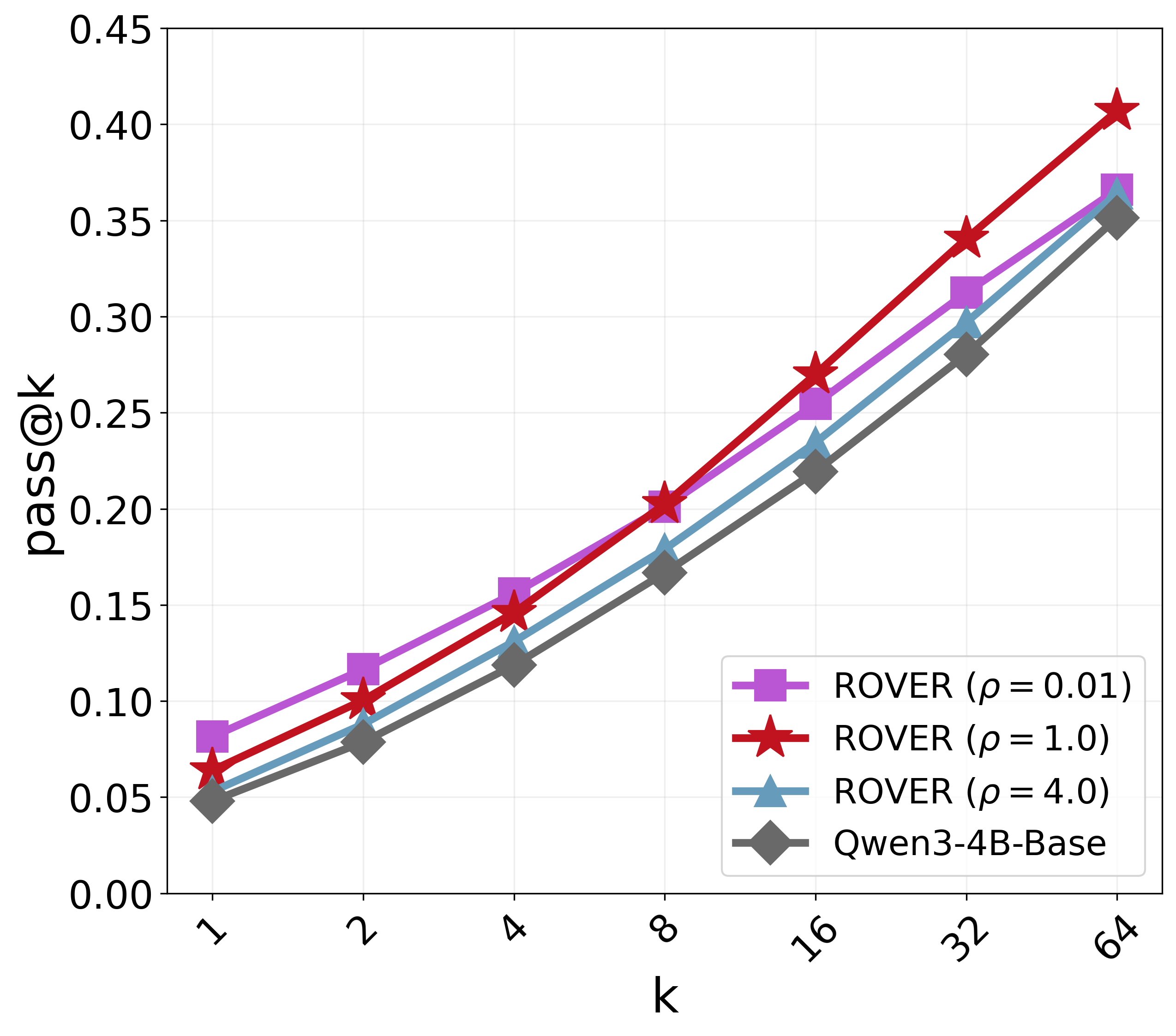}\label{fig:ab_rpe_a}}
    \subfigure[]{\includegraphics[width=0.35\linewidth]{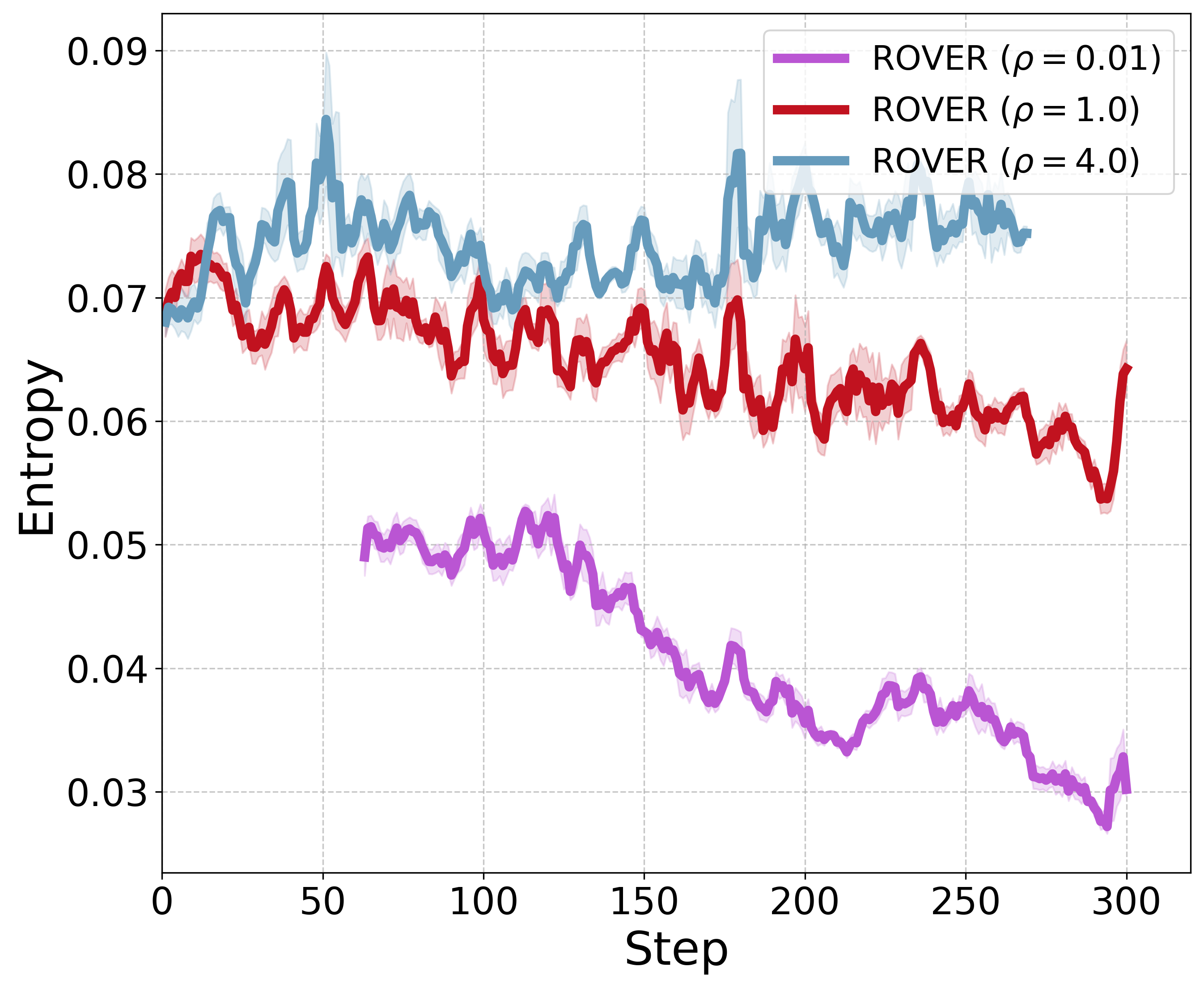}\label{fig:ab_rpe_b}}
    \vspace{-.10in}
    \caption{Impact of temperature $\rho$. All experiments are conducted on Qwen3-4B-Base and trained for 300 steps. (a): pass@$k$ results (average performance on AIME24, AIME25, HMMT25 are reported). (b): entropy curves throughout training.}
    \vspace{-.10in}
    \label{fig:ab_rpe}
\end{figure*}

Consistent with the findings on the countdown task in Fig.~\ref{fig:ablation_coundown_rho}, the training temperature $\rho$ serves as an exploration-exploitation trade-off. A large $\rho$ ($\rho=4$) results in more stochastic behavior and constant entropy throughout training, which affects the performance (see Fig.~\ref{fig:ab_rpe}). Conversely, a smaller $\rho$ ($\rho=0.01$) leads to a greedy and deterministic policy, which compromises diversity (e.g., reduced pass@$k$) for improved pass@1 performance. By default, we set $\rho = 1$ in other experiments.

\subsubsection{Training dynamics} 
\begin{figure*}[htbp]
    \centering
    \subfigure[]{\includegraphics[width=0.32\linewidth]{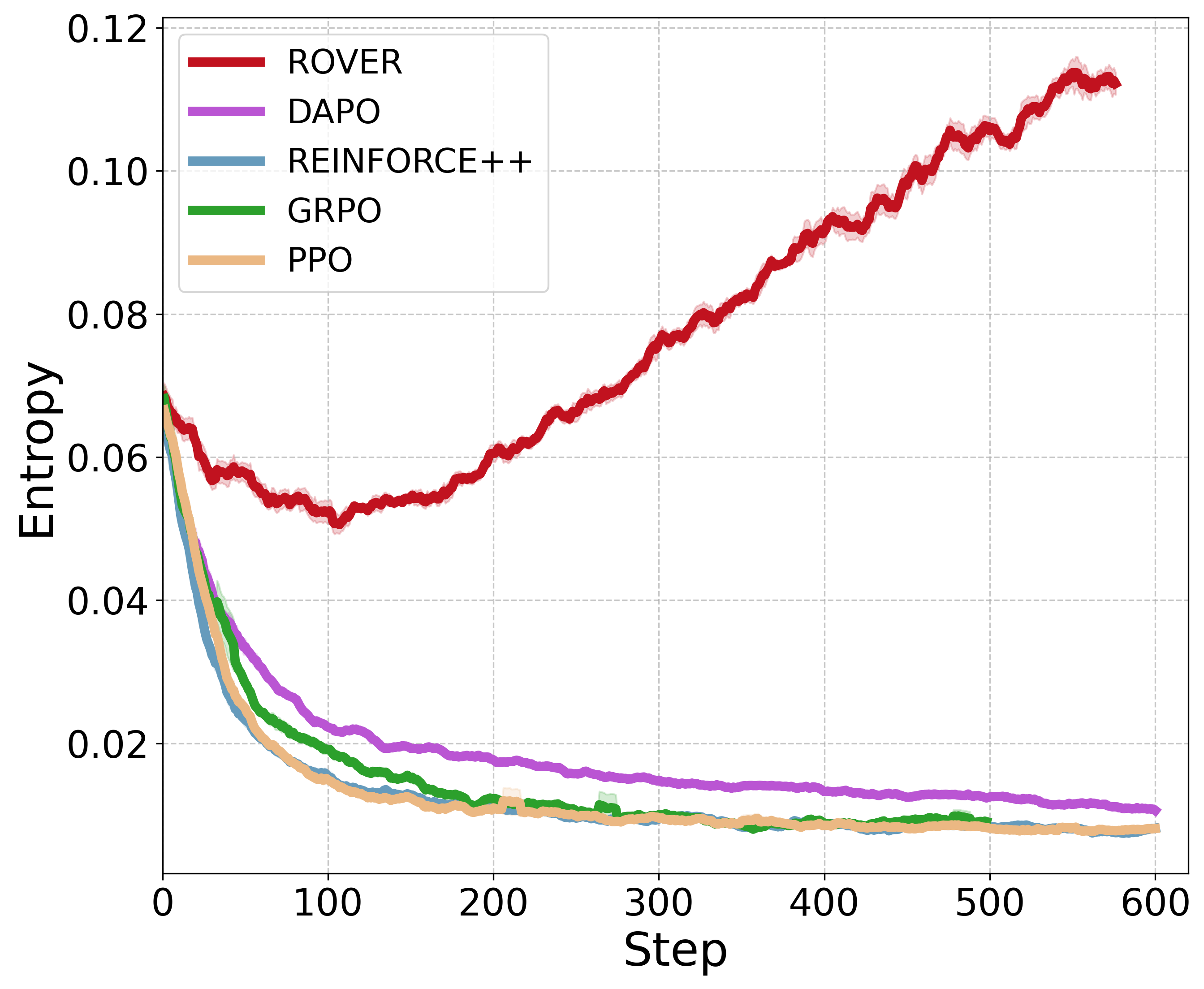}\label{fig:8b-entropy}}
    \subfigure[]{\includegraphics[width=0.32\linewidth]{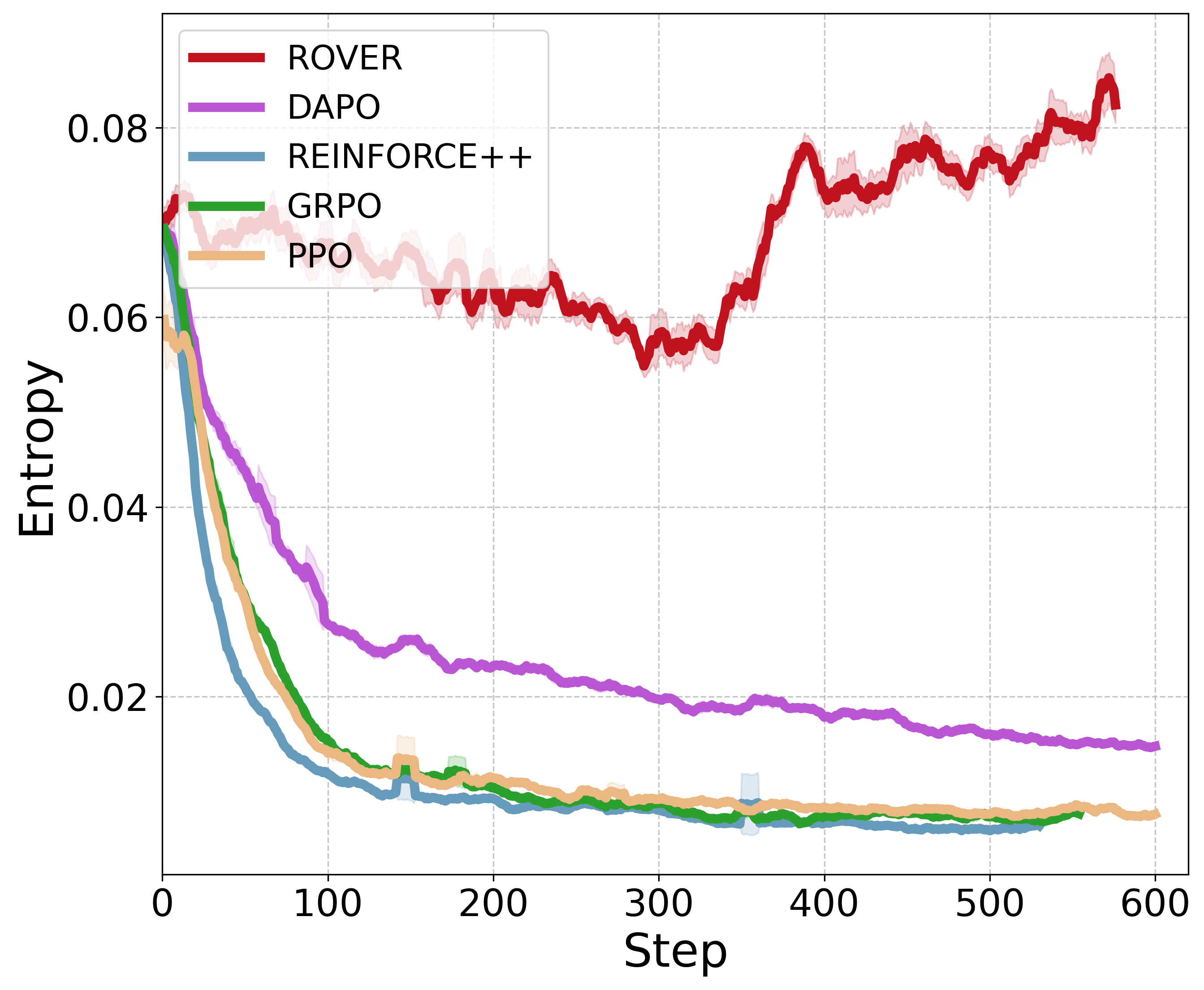}\label{fig:4b-entropy}}
    \vspace{-.15in}
    \caption{Training curves of entropy for \ours and baselines. (a) \& (b) are results on Qwen3-8B-Base and Qwen3-4B-Base, respectively. The entropy of ROVER is maintained at a relatively higher level, and can even increase stably at later training stages, indicating expanded exploration space. In contrast, the entropy of baselines inevitably decreases to a low level.}
    \vspace{-.15in}
    \label{fig:qwen-entropy}
    \vspace{-.15in}
\end{figure*}

\begin{figure*}[htbp]
    \centering
    
    \subfigure[$Q^\prime$ (Min in batch)]{\includegraphics[width=0.3\linewidth]{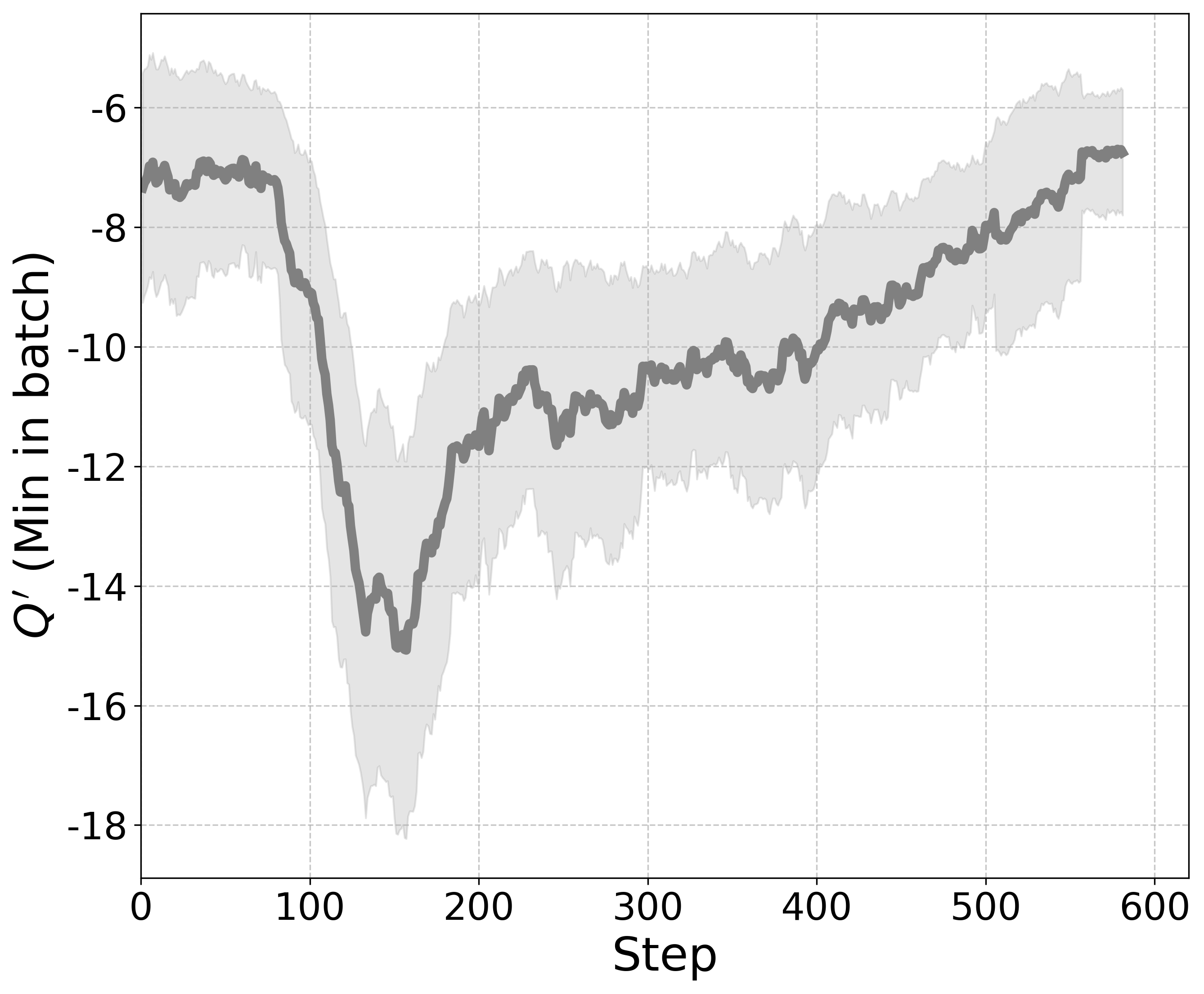}\label{fig:}}
    \subfigure[$Q^\prime$ (Mean in batch)]{\includegraphics[width=0.3\linewidth]{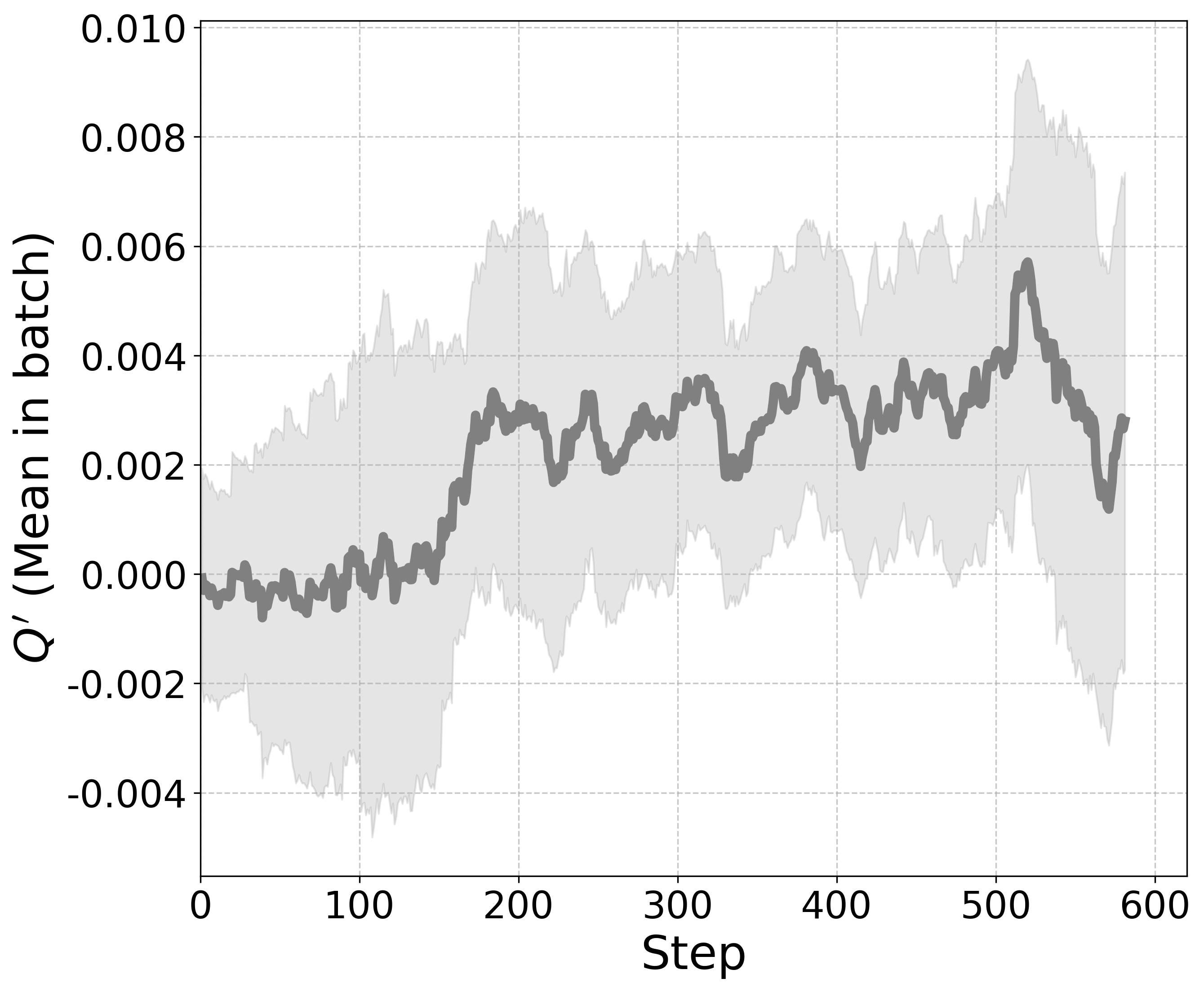}\label{fig:}}
    \subfigure[$Q^\prime$ (Max in batch)]{\includegraphics[width=0.3\linewidth]{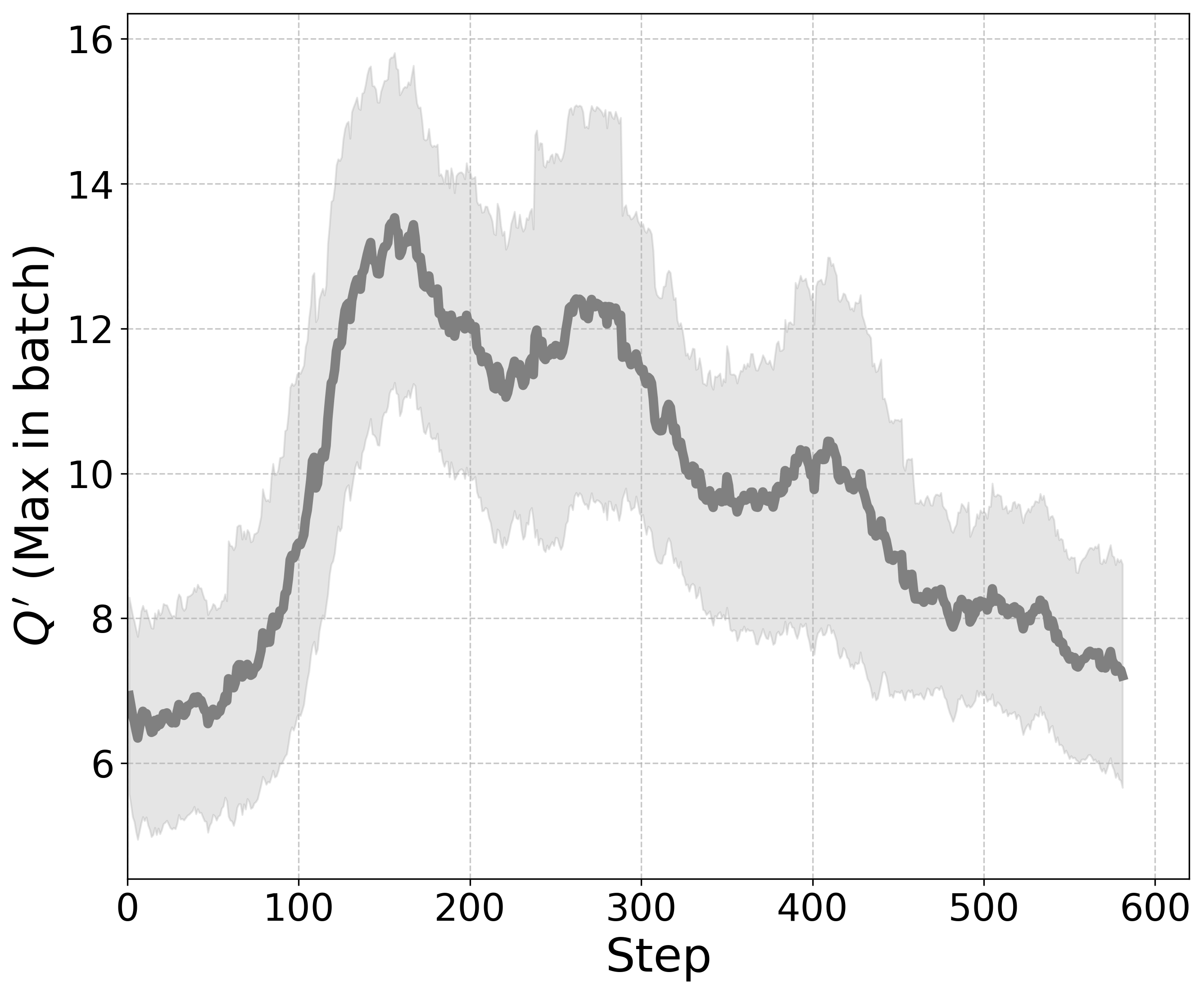}\label{fig:}}
    \vspace{-.15in}
    \caption{Absolute scales of $\tilde{r}$ and $Q^\prime$ throughout training for \ours (trained on Qwen-8B-Base).}
    \vspace{-.15in}
    \label{fig:8b-adv-nextq}
\end{figure*}

We present the training curves of entropy in Fig.~\ref{fig:qwen-entropy}. The min, mean, and max values of $Q^\prime$ within a training batch are visualized in Fig.~\ref{fig:8b-adv-nextq}.

\subsection{Measuring the diversity of LLM responses}\label{app:diversity}

In addition to the number of distinct strategies mentioned in \S~\ref{sec:exp}, we additionally incorporate two diversity metrics for a comprehensive evaluation. These diversity metrics are introduced as follows.

\textbf{No. of Distinct strategies}~\citep{zhang2025noveltybench}. It categorizes all generated responses into equivalent strategy classes and counts the total number of distinct classes.

\textbf{Utility}~\citep{zhang2025noveltybench}. It combines diversity and quality using a user patience model where users have a probability $p$ of requesting additional generations. It rewards novel responses while applying geometric decay to account for diminishing user attention over multiple generations. Models capable of generating multiple correct responses with distinct strategies will receive a higher utility score.

\textbf{Cosine Distance}. We embed all responses using Qwen3-8B-Embedding~\citep{qwen3embedding} and compute the average pairwise cosine distance between response vectors. Higher distances indicate greater semantic diversity among generated responses. Specifically, given a set of generated responses $\{y_1, y_2, \ldots, y_n\}$, let $E(y_i) \in \mathbb{R}^d$ denote the L2-normalized embedding vector of response $y_i$ obtained from Qwen3-8B-Embedding. The pairwise cosine similarity between responses $y_i$ and $y_j$ is:
$$S(y_i, y_j) = E(y_i) \cdot E(y_j).$$
The average pairwise cosine similarity is:
$$\bar{S} = \frac{1}{n(n-1)} \sum_{i=1}^{n} \sum_{j \neq i} S(y_i, y_j).$$
Finally, the cosine distance is defined as $1 - \bar{S}$.

\begin{figure*}[htb]
    \centering
    \subfigure[$t=0.3$]{\includegraphics[width=0.3\linewidth]{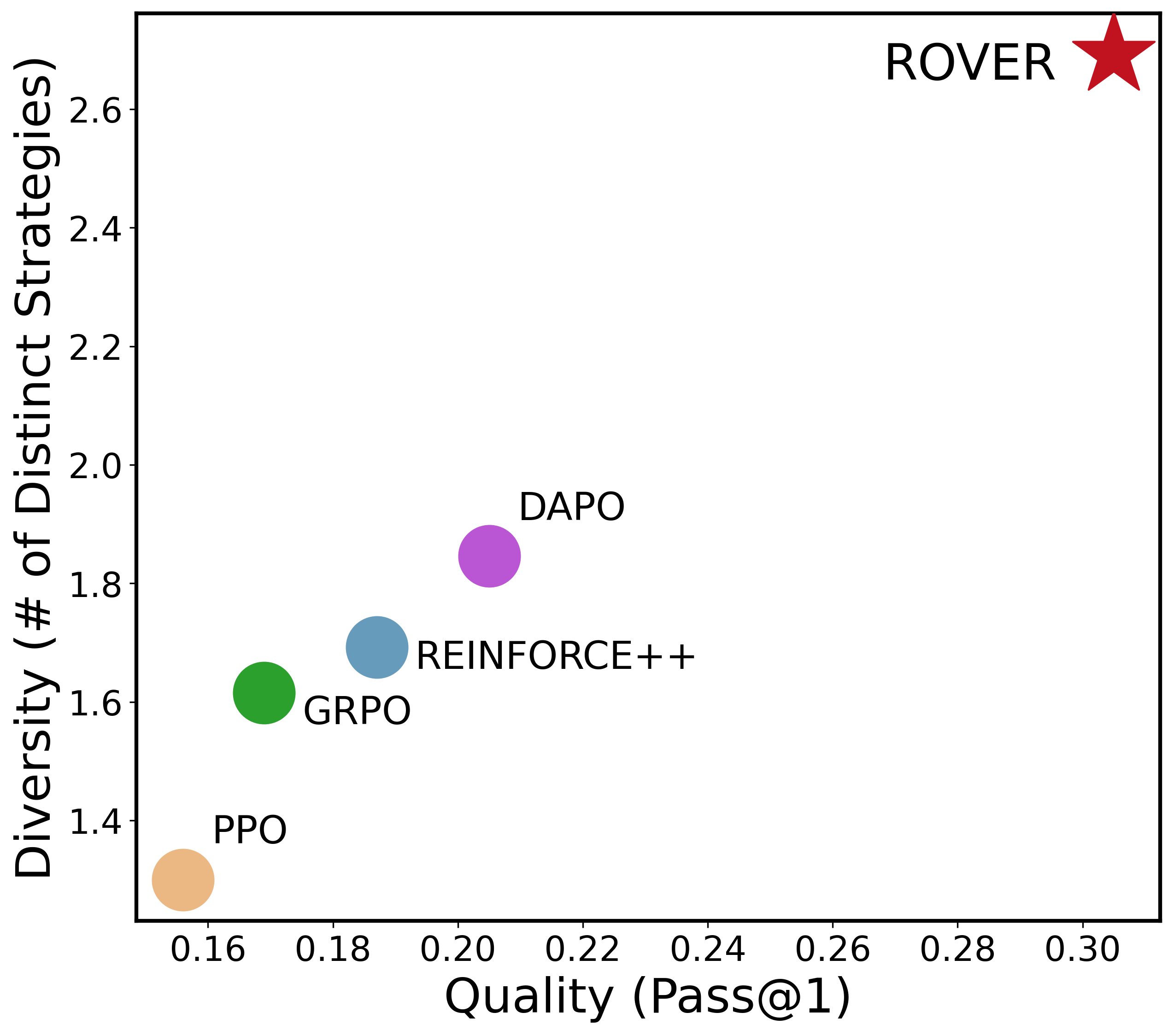}}
    \subfigure[$t=0.9$]{\includegraphics[width=0.3\linewidth]{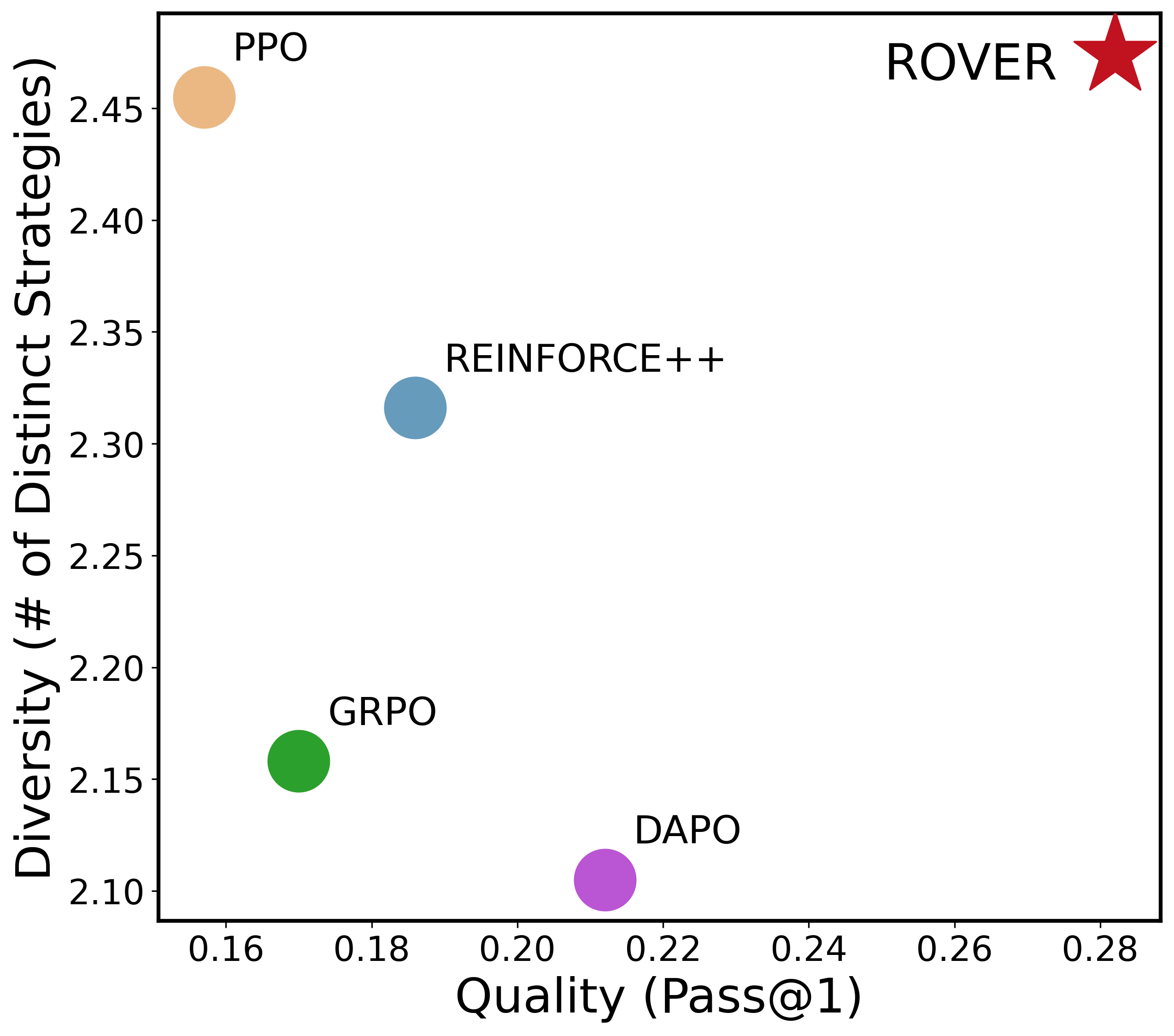}}
    \subfigure[$t=1.2$]{\includegraphics[width=0.3\linewidth]{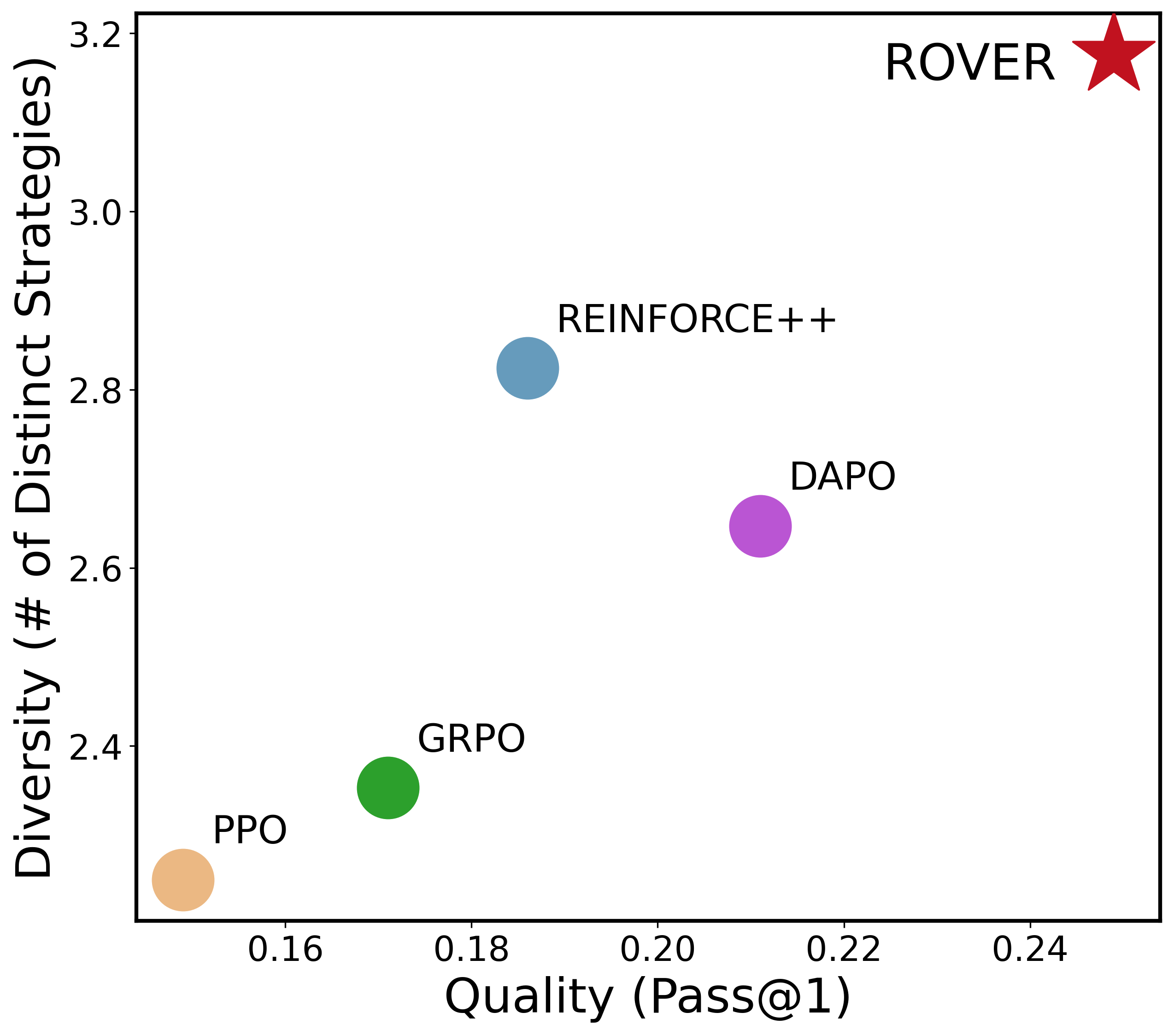}}
    \vspace{-.10in}
    \caption{Quality-diversity trade-off with different decoding temperature (AIME24). 
    }
    \label{fig:quality-diversity-trade-off-temperature}
\end{figure*}

As a supplement to Fig.~\ref{fig:quality-diversity-trade-off}, results of quality-diversity trade-off across $t \in [0.3, 0.9, 1.2]$ are shown in Fig.~\ref{fig:quality-diversity-trade-off-temperature}.

\begin{figure*}[htb]
    \centering
    \vspace{-1em}
    \subfigure[$t=0.3$]{\includegraphics[width=0.3\linewidth]{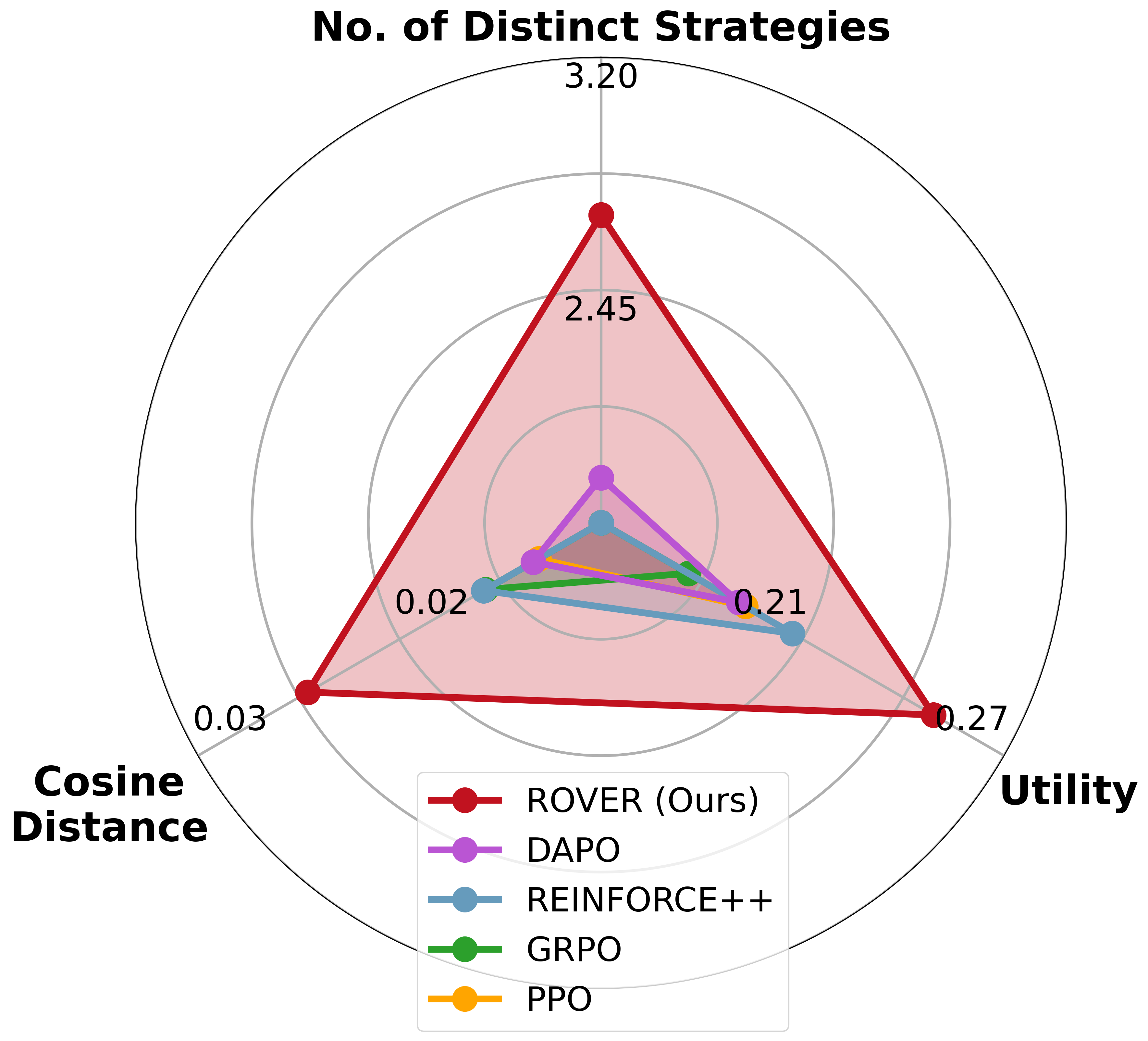}}
    \subfigure[$t=0.9$]{\includegraphics[width=0.3\linewidth]{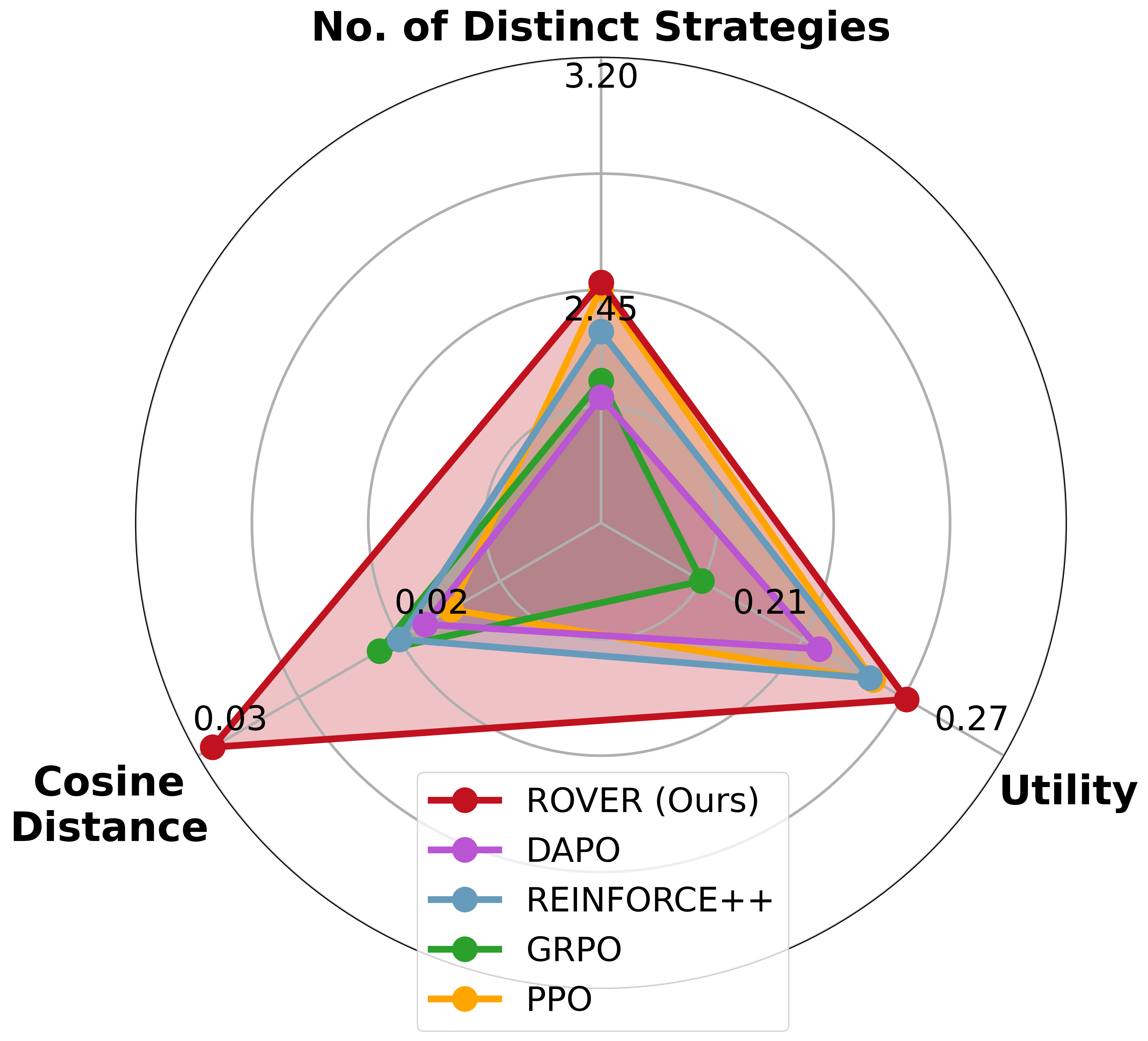}}
    \subfigure[$t=1.2$]{\includegraphics[width=0.3\linewidth]{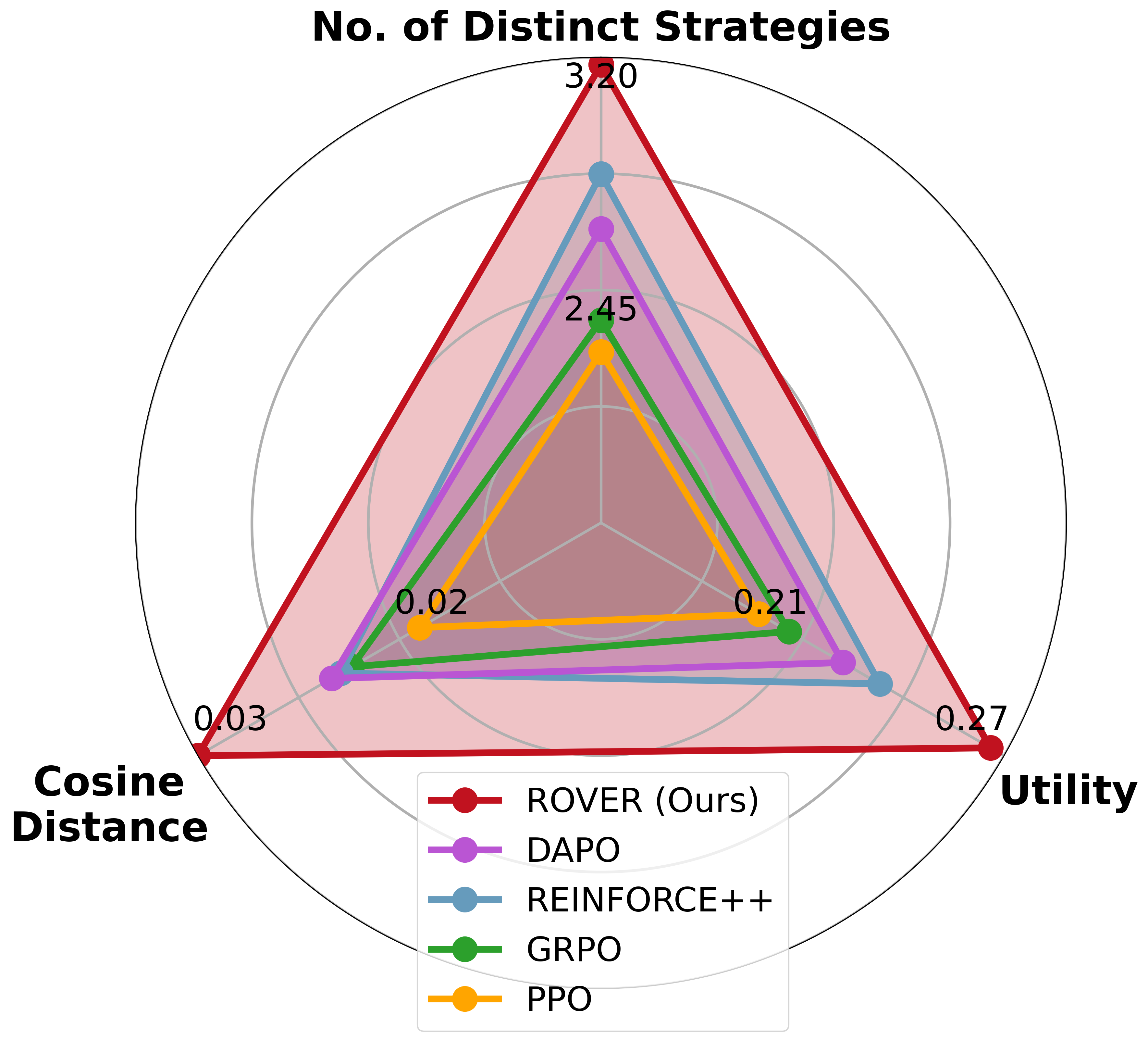}}
    \vspace{-.10in}
    \caption{Comparison of multiple diversity metrics with different decoding temperatures (AIME24). The superiority of \ours in terms of diversity is most pronounced when decoding temperature is relatively low (e.g., $t=0.3$). As the decoding temperature increases, the diversity of baseline RL methods improves, but remains lower than that of ROVER.
    }
    \label{fig:radar-all-temperature}
\end{figure*}

Furthermore, we demonstrate the comparison on all three diversity metrics under different decoding temperatures in Fig.~\ref{fig:radar-all-temperature}. \ours consistently exhibits greater diversity across all decoding temperatures.

\subsection{Prompts}

We present the prompt template for RL training and evaluation in Fig.~\ref{fig:prompt-template}, and the prompt for LLM judger in Fig.~\ref{fig:prompt-llm-judge}.

\begin{figure*}[ht]
\begin{tcolorbox}[colframe=black, colback=lightblue, coltitle=black, rounded corners, boxrule=0.3mm, width=\textwidth, halign=left]
\begin{verbatim}
<|im_start|>user
{question} 
Please reason step by step, and put your final answer within
\boxed{}.
<|im_end|>
<|im_start|>assistant
\end{verbatim}
\end{tcolorbox}
\caption{Prompt template for RL training and evaluation. The base model uses the same prompt template as the trained model during evaluation.}
\label{fig:prompt-template}
\end{figure*}

\begin{figure*}[ht]
\begin{tcolorbox}[colframe=black, colback=lightblue, coltitle=black, 
                  rounded corners, boxrule=0.3mm, width=\textwidth, halign=left]
\begin{verbatim}
You are given the original prompt and two model-generated 
responses. Determine whether the two responses use different
strategies to solve the problem.

Use the following guidelines to identify different strategies:

1. Mathematical Tools and Concepts:
- Using different mathematical tools (e.g., differentiation vs. 
integration, series expansion vs. direct computation)
- Applying different theorems or properties (e.g., mean value 
theorem vs. fundamental theorem of calculus)
- Different mathematical domains (e.g., algebraic vs. geometric, 
analytical vs. combinatorial)

2. Solution Structure Differences:
- Different variable substitutions or transformations
- Different equation setups for the same problem
- Different ways of breaking down the problem into subproblems

3. Specific Examples of Different Approaches:
- Direct computation vs. recursive method
- Forward solving vs. backward solving (working from the answer)
- Algebraic manipulation vs. numerical approximation
- Using contradiction vs. direct proof
- Using induction vs. direct formula
- Coordinate-based vs. coordinate-free methods

Even if two solutions arrive at the same answer, they should be 
considered different if they:
- Use different key mathematical tools or theorems
- Follow different logical sequences in critical steps
- Represent the problem using different mathematical frameworks
- Break down the problem in substantially different ways

Original prompt: {prompt}
Generation 0: {generation0}
Generation 1: {generation1}

Question: Do Generation 0 and Generation 1 use different 
strategies? First analyze the key mathematical tools and 
solution structure used in each solution, then respond 
with "[[yes]]" if the generations use different 
strategies or "[[no]]" if they do not.
\end{verbatim}
\end{tcolorbox}
\caption{Prompt for LLM judger to determine whether two responses use different strategies. We refined the prompt proposed in~\citep{darling} to enhance the LLM judge's capability for more nuanced strategy classification.}
\label{fig:prompt-llm-judge}
\end{figure*}

\end{document}